\documentclass[twoside]{article}
\usepackage[accepted]{aistats2018}
\usepackage[letterpaper, margin=1in]{geometry}
\usepackage[utf8]{inputenc} 
\usepackage[T1]{fontenc}    
\usepackage{hyperref}       
\usepackage{url}            
\usepackage{booktabs}       
\usepackage{amsfonts}       
\usepackage{nicefrac}       
\usepackage{microtype}      
\usepackage{amsmath,latexsym,amssymb,amsthm,bbm}
\usepackage{float}
\usepackage[]{algorithm2e}
\usepackage{graphicx}
\usepackage{lipsum}
\usepackage{caption}
\usepackage{subcaption}
\usepackage{natbib}
\usepackage{dsfont}

\newtheorem{thm}{Theorem}[section]
\newtheorem{cor}[thm]{Corollary}
\newtheorem{prop}[thm]{Proposition}
\newtheorem{lem}[thm]{Lemma}

\theoremstyle{definition}
\newtheorem{defn}[thm]{Definition}

\newtheorem{prog}[thm]{Program}

\theoremstyle{remark}

\begin{document}

%

%

\twocolumn[

\aistatstitle{An Optimization Approach to Learning Falling Rule Lists}

\aistatsauthor{Chaofan Chen \And Cynthia Rudin }

\aistatsaddress{ Duke University \And Duke University } ]



\begin{abstract}
  A falling rule list is a probabilistic decision list for binary classification, consisting of a series of {\bf if-then} rules with antecedents in the {\bf if} clauses and probabilities of the desired outcome (``1'') in the {\bf then} clauses. Just as in a regular decision list, the order of rules in a falling rule list is important -- each example is classified by the first rule whose antecedent it satisfies. Unlike a regular decision list, a falling rule list requires the probabilities of the desired outcome (``1'') to be monotonically decreasing down the list. We propose an optimization approach to learning falling rule lists and ``softly'' falling rule lists, along with Monte-Carlo search algorithms that use bounds on the optimal solution to prune the search space.
\end{abstract}


\section{INTRODUCTION}\label{section:introduction}


In many real-life scenarios, we want to learn a predictive model that allows us to easily identify the most significant conditions that are predictive of a certain outcome. For example, in health care, doctors often want to know the conditions that signify a high risk of stroke, so that patients with such conditions can be prioritized in receiving treatment. A falling rule list, whose form was first proposed by Wang and Rudin (\citeyear{Wang}), is a type of model that serves this purpose.


\begin{table*}[h]
\centering
\caption{Falling Rule List for bank-full Dataset} \label{table:frl_bank-full}
\centering
\begin{tabular}{llllll}
 & antecedent & & prob.        & $+$ & $-$ \\\hline
IF      & poutcome=success AND default=no     & THEN success prob. is & 0.65 & 978  & 531  \\
ELSE IF & 60 $\leq$ age $<$ 100 AND default=no& THEN success prob. is & 0.28 & 434  & 1113 \\
ELSE IF & 17 $\leq$ age $<$ 30 AND housing=no & THEN success prob. is & 0.25 & 504  & 1539 \\
ELSE IF & previous $\geq$ 2 AND housing=no    & THEN success prob. is & 0.23 & 242  & 794  \\
ELSE IF & campaign=1 AND housing=no           & THEN success prob. is & 0.14 & 658  & 4092 \\
ELSE IF & previous $\geq$ 2 AND education=tertiary & THEN success prob. is & 0.13 & 108  & 707  \\
ELSE    &                               & success prob. is      & 0.07 & 2365 & 31146
\end{tabular}
\end{table*}

Table \ref{table:frl_bank-full} shows a falling rule list we learned from the bank-full dataset, which was used by Moro et al. (\citeyear{Moro}) in their study of applying data mining techniques to direct marketing. As we can see, a falling rule list is a probabilistic decision list for binary classification, consisting of a series of {\bf if-then} rules with antecedents in the {\bf if} clauses and probabilities of the desired outcome (``1'') in the {\bf then} clauses, where the probabilities of the desired outcome (``1'') are monotonically decreasing down the list (hence the name ``falling'' rule list). The falling rule list in Table \ref{table:frl_bank-full} has identified clients for whom the previous marketing campaign was successful (``poutcome=success''), and who have no credit in default (``default=no''), as individuals who are most likely to subscribe to a term deposit in the current marketing campaign. Their probability of subscribing is $0.65$. Of the remaining clients, those who are next most likely to sign up for a term deposit are older people (aged between 60 and 100) with no credit in default. Their probability of subscribing is $0.28$. The two rightmost columns in Table \ref{table:frl_bank-full}, labeled $+$ and $-$, show the number of positive training examples (i.e. clients who subscribe to a term deposit in the current campaign) and of negative training examples, respectively, that satisfy the antecedent in each rule of the falling rule list.

Falling rule lists can provide valuable insight into data -- if we know how to construct them well. In this paper, we propose an optimization approach to learning falling rule lists and ``softly'' falling rule lists, along with Monte-Carlo search algorithms that use bounds on the optimal solution to prune the search space. The falling rule list shown in Table \ref{table:frl_bank-full} was produced using Algorithm FRL, which we shall introduce later.


Our work lives within several well-established fields, but is the first work we know of to use an optimization approach to handling monotonicity constraints in rule-based models. It relates closely to associative classification (e.g. the RIPPER$k$ algorithm \citep{Cohen} and the CBA algorithm \citep{Liu}; see \cite{Thabtah} for a comprehensive review) and inductive logic programming \citep{Muggleton}. The proposed algorithms are competitors for decision tree methods like CART \citep{Breiman}, ID3 \citep{Quinlan}, C4.5 \citep{Quinlan2}, and C5.0 \citep{Quinlan3}, and decision list learning \citep{Rivest}. Almost all methods from this class build decision trees from the top down using greedy splitting criteria. Greedy splitting criteria do not lend naturally to constrained models like falling rule lists. There are some works on decision trees with monotonicity constraints \citep[e.g.][]{Altendorf,Ben-David,Feelders}, but they focus mostly on enforcing the monotonic relationship between certain attributes and ordinal class labels. In addition, our work also relates to those that underline the importance of the interpretability of models \citep{Freitas,Huysmans,Kodratoff,Martens}.

Wang and Rudin (\citeyear{Wang}) proposed the form of a falling rule list, and a Bayesian approach to learning falling rule lists (extending the ideas of \cite{Letham} and \cite{Yang}). The Bayesian approach offers some advantages: e.g. a full posterior over rule lists allows model averaging. However, the optimization perspective has an important computational advantage: the search space is made substantially smaller by the tight bounds presented here. The concept of softly falling rule lists is novel to this paper and has not been done in the Bayesian setting.

\section{PROBLEM FORMULATION}


We first formalize the notion of an antecedent, of a rule list, of a falling rule list, and of a prefix.

\begin{defn}\label{defn:decision_rule}
An {\it antecedent} $a$ on an input domain $\mathcal{X}$ is a Boolean function that outputs true or false. Given an input $\mathbf{x} \in \mathcal{X}$, we say that $\mathbf{x}$ satisfies the antecedent $a$ if $a(\mathbf{x})$ evaluates to true. For example, (poutcome=success AND default=no) in Table \ref{table:frl_bank-full} is an antecedent.
\end{defn}

\begin{defn}\label{defn:rl}
A {\it rule list} $d: \mathcal{X} \rightarrow [0, 1]$ on an input domain $\mathcal{X}$ is a probabilistic decision list of the following form: ``if $\mathbf{x}$ satisfies $a_0^{(d)}$, then $\text{Pr}(y = 1|\mathbf{x}) = \hat{\alpha}_0^{(d)}$; else if $\mathbf{x}$ satisfies $a_1^{(d)}$, then $\text{Pr}(y = 1|\mathbf{x}) = \hat{\alpha}_1^{(d)}$; $...$; else if $\mathbf{x}$ satisfies $a_{|d|-1}^{(d)}$, then $\text{Pr}(y = 1|\mathbf{x}) = \hat{\alpha}_{|d|-1}^{(d)}$; else $\text{Pr}(y = 1|\mathbf{x}) = \hat{\alpha}_{|d|}^{(d)}$''
where $a_j^{(d)}$ is the $j$-th antecedent in $d$, $j \in \{0, 1, ..., |d| - 1\}$, and $|d|$ denotes the size of the rule list, which is defined as the number of rules, excluding the final else clause, in the rule list. We can denote the rule list $d$ as follows:
\begin{equation}\label{eq:rl}
\begin{split}
d = \{&(a_0^{(d)}, \hat{\alpha}_0^{(d)}), (a_1^{(d)}, \hat{\alpha}_1^{(d)}), ...,\\ &(a_{|d|-1}^{(d)}, \hat{\alpha}_{|d|-1}^{(d)}), \hat{\alpha}_{|d|}^{(d)}\}.
\end{split}
\end{equation}
The rule list $d$ of Equation (\ref{eq:rl}) is a {\it falling rule list} if the following inequalities hold:
\begin{equation}\label{eq:monotonicity}
\hat{\alpha}_0^{(d)} \geq \hat{\alpha}_1^{(d)} \geq ... \geq \hat{\alpha}_{|d|-1}^{(d)} \geq \hat{\alpha}_{|d|}^{(d)}.
\end{equation}
For convenience, we sometimes refer to the final else clause in $d$ as the $|d|$-th antecedent $a_{|d|}^{(d)}$ in $d$, which is satisfied by all $\mathbf{x} \in \mathcal{X}$. We denote the space of all possible rule lists on $\mathcal{X}$ by $\mathcal{D}(\mathcal{X})$.
\end{defn}

\begin{defn}\label{defn:prefix}
A {\it prefix} $e$ on an input domain $\mathcal{X}$ is a rule list without the final else clause. We can denote the prefix $e$ as follows:
\begin{equation}\label{eq:prefix}
e = \{(a_0^{(e)}, \hat{\alpha}_0^{(e)}), (a_1^{(e)}, \hat{\alpha}_1^{(e)}), ..., (a_{|e|-1}^{(e)}, \hat{\alpha}_{|e|-1}^{(e)})\}.
\end{equation}
where $a_j^{(e)}$ is the $j$-th antecedent in $e$, $j \in \{0, 1, ..., |e| - 1\}$, and $|e|$ denotes the size of the prefix, which is defined as the number of rules in the prefix. 

\end{defn}

\begin{defn}\label{defn:captured}
Given the rule list $d$ of Equation (\ref{eq:rl}) (or the prefix $e$ of Equation (\ref{eq:prefix})), we say that an input $\mathbf{x} \in \mathcal{X}$ is {\it captured} by the $j$-th antecedent in $d$ (or $e$) if $\mathbf{x}$ satisfies $a_j^{(d)}$ (or $a_j^{(e)}$, respectively), and for all $k \in \{0, 1, ..., |d|\}$ (or $k \in \{0, 1, ..., |e| - 1\}$, respectively) such that $\mathbf{x}$ satisfies $a_k^{(d)}$ (or $a_k^{(e)}$, respectively), $j \leq k$ holds -- in other words, $a_j^{(d)}$ (or $a_j^{(e)}$, respectively) is the first antecedent that $\mathbf{x}$ satisfies. We define the function $\text{capt}$ by $\text{capt}(\mathbf{x}, d) = j$ (or $\text{capt}(\mathbf{x}, e) = j$)
if $\mathbf{x}$ is captured by the $j$-th antecedent in $d$ (or $e$). Moreover, given the prefix $e$ of Equation (\ref{eq:prefix}), we say that an input $\mathbf{x} \in \mathcal{X}$ is captured by the prefix $e$ if $\mathbf{x}$ is captured by some antecedent in $e$, and we define $\text{capt}(\mathbf{x}, e) = |e|$ if $\mathbf{x}$ is not captured by the prefix $e$.

\end{defn}

Let $D = \{(\mathbf{x}_i, y_i)\}_{i=1}^n$ be the training data, with $\mathbf{x}_i \in \mathcal{X}$ and $y_i \in \{1, -1\}$ for each $i \in \{1, 2, ..., n\}$. We now define the empirical positive proportion of an antecedent, and introduce the notion of a rule list (or a prefix) that is compatible with $D$.

\begin{defn}\label{defn:emp_pos_prop}
Given the training data $D$ and the rule list $d$ of Equation (\ref{eq:rl}) (or the prefix $e$ of Equation (\ref{eq:prefix})), we denote by $n^+_{j, d, D}$, $n^-_{j, d, D}$, $n_{j, d, D}$ (or $n^+_{j, e, D}$, $n^-_{j, e, D}$, $n_{j, e, D}$), the number of positive, negative, and all training inputs captured by the $j$-th antecedent in $d$ (or $e$), respectively, and define the {\it empirical positive proportion} of the $j$-th antecedent in $d$ (or $e$), denoted by $\alpha_j^{(d, D)}$ (or $\alpha_j^{(e, D)}$), as:
\begin{equation*}
\alpha_j^{(d, D)} = n^+_{j, d, D}/n_{j, d, D}
\text{ (or } \alpha_j^{(e, D)} = n^+_{j, e, D}/n_{j, e, D} \text{).}
\end{equation*}
Moreover, given the training data $D$ and the prefix $e$ of Equation (\ref{eq:prefix}), we denote by $\tilde{n}^+_{e, D}$, $\tilde{n}^-_{e, D}$, $\tilde{n}_{e, D}$, the number of positive, negative, and all training inputs that are not captured by the prefix $e$, and define the {\it empirical positive proportion after the prefix} $e$, denoted by $\tilde{\alpha}_{e, D}$, as $\tilde{\alpha}_{e, D} = \tilde{n}^+_{e, D}/\tilde{n}_{e, D}$.

\end{defn}

\begin{defn}\label{defn:compatible}
Given the training data $D$ and the rule list $d$ of Equation (\ref{eq:rl}) (or the prefix $e$ of Equation (\ref{eq:prefix})), we say that the rule list $d$ (or the prefix $e$) is {\it compatible} with $D$ if for all $j \in \{0, 1, ..., |d|\}$ (or $j \in \{0, 1, ..., |e| - 1\}$, respectively), the equation $\hat{\alpha}_j^{(d)} = \alpha_j^{(d, D)}$ ($\hat{\alpha}_j^{(e)} = \alpha_j^{(e, D)}$, respectively) holds. We denote the space of all possible rule lists on $\mathcal{X}$ that are compatible with the training data $D$ by $\mathcal{D}(\mathcal{X}, D)$.
\end{defn}


To formulate the problem of learning falling rule lists from data as an optimization program, we first observe that, given a threshold $\tau$, the rule list $d$ of Equation (\ref{eq:rl}) can be viewed as a classifier $\tilde{d}_\tau: \mathcal{X} \rightarrow \{1, -1\}$ that predicts $1$ for an input $\mathbf{x} \in \mathcal{X}$ only if the inequality $\hat{\alpha}_{\text{capt}(\mathbf{x}, d)}^{(d)} > \tau$ holds. Hence, we can define the empirical risk of misclassification by the rule list $d$ on the training data $D$ as that by the classifier $\tilde{d}_\tau$. More formally, we have the following definition.


\begin{defn}\label{defn:emp_risk}
Given the training data $D$, the rule list $d$ of Equation (\ref{eq:rl}), a threshold $\tau$, and the weight $w$ for the positive class, the {\it empirical risk of misclassification by the rule list} $d$ on the training data $D$ with threshold $\tau$ and with weight $w$ for the positive class, denoted by $R(d, D, \tau, w)$, is:
\begin{equation}\label{eq:emp_risk}
\begin{split}
R(d, D, \tau, w) = \frac{1}{n}&\left(w\sum_{i: y_i = 1} \mathds{1}[\hat{\alpha}_{\text{capt}(\mathbf{x}_i, d)}^{(d)} \leq \tau] \right. \\
                   & \left. \vphantom{\sum_{i: y_i = 1} \mathds{1}[\hat{\alpha}_{\text{capt}(\mathbf{x}_i, d)}^{(d)} \leq \tau]} + \sum_{i: y_i = -1} \mathds{1}[\hat{\alpha}_{\text{capt}(\mathbf{x}_i, d)}^{(d)} > \tau]\right).
\end{split}
\end{equation}
If $d$ is compatible with $D$, we can replace $\hat{\alpha}_{\text{capt}(\mathbf{x}_i, d)}^{(d)}$ in Equation (\ref{eq:emp_risk}) with $\alpha_{\text{capt}(\mathbf{x}_i, d)}^{(d, D)}$. We define the {\it empirical risk of misclassification by the prefix} $e$ on the training data $D$ with threshold $\tau$ and with weight $w$ for the positive class, denoted by $R(e, D, \tau, w)$, analogously:
\begin{equation}\label{eq:emp_risk_prefix}
\begin{split}
R(e, D, \tau, w) = &\frac{1}{n}\left(w\sum_{\substack{i: y_i = 1 \wedge \\ \text{capt}(\mathbf{x}_i, e) \neq |e|}} \mathds{1}[\hat{\alpha}_{\text{capt}(\mathbf{x}_i, e)}^{(e)} \leq \tau] \right. \\
                   & \left. \vphantom{\sum_{\substack{i: y_i = 1 \wedge \\ \text{capt}(\mathbf{x}_i, e) \neq |e|}} \mathds{1}[\hat{\alpha}_{\text{capt}(\mathbf{x}_i, e)}^{(e)} \leq \tau]} + \sum_{\substack{i: y_i = -1 \wedge \\ \text{capt}(\mathbf{x}_i, e) \neq |e|}} \mathds{1}[\hat{\alpha}_{\text{capt}(\mathbf{x}_i, e)}^{(e)} > \tau]\right).
\end{split}
\end{equation}
If $e$ is compatible with $D$, we can replace $\hat{\alpha}_{\text{capt}(\mathbf{x}_i, e)}^{(e)}$ in Equation (\ref{eq:emp_risk_prefix}) with $\alpha_{\text{capt}(\mathbf{x}_i, e)}^{(e, D)}$. Note that for any rule list $d$ that begins with a given prefix $e$, $R(e, D, \tau, w)$ is the contribution by the prefix $e$ to $R(d, D, \tau, w)$.

\end{defn}

We can formulate the problem of learning falling rule lists as a minimization program of the empirical risk of misclassification, given by Equation (\ref{eq:emp_risk}), with a regularization term $C|d|$ that penalizes each rule in $d$ with a cost of $C$ to limit the number of rules, subject to the monotonicity constraint (\ref{eq:monotonicity}). For now, we focus on the problem of learning falling rule lists that are compatible with the training data $D$. 

Let $L(d, D, \tau, w, C) = R(d, D, \tau, w) + C|d|$ and $L(e, D, \tau, w, C) = R(e, D, \tau, w) + C|e|$ be the regularized empirical risk of misclassification by the rule list $d$ and by the prefix $e$, respectively, on the training data $D$. The former defines the objective of the minimization program, and the latter gives the contribution by the prefix $e$ to $L(d, D, \tau, w, C)$ for any rule list $d$ that begins with $e$. The following theorem provides a motivation for setting the threshold $\tau$ to $1/(1+w)$ in the minimization program -- the empirical risk of misclassification by a given rule list $d$ is minimized when $\tau$ is set in this way.

\begin{thm}
Given the training data $D$, a rule list $d$ that is compatible with $D$, and the weight $w$ for the positive class, we have $R(d, D, 1/(1+w), w) \leq R(d, D, \tau, w)$ for all $\tau \geq 0$.
\end{thm}

For reasons of computational tractability and model interpretability, we further restrict our attention to learning compatible falling rule lists whose antecedents must come from a pre-determined set of antecedents $A = \{A_l\}_{l=1}^m$. We now present the optimization program for learning falling rule lists, which forms the basis of the rest of this paper.

\begin{prog}[Learning compatible falling rule lists]\label{prog:frl}
\begin{equation*}
\min_{d \in \mathcal{D}(\mathcal{X}, D)} L(d, D, 1/(1+w), w, C) \text{ subject to }
\end{equation*}
\begin{equation}\label{eq:monotonicity_constr}
\alpha_0^{(d, D)} \geq \alpha_1^{(d, D)} \geq ... \geq \alpha_{|d|-1}^{(d, D)} \geq \alpha_{|d|}^{(d, D)},
\end{equation}
\begin{equation}\label{eq:antecedent_constr}
a_j^{(d)} \in A, \text{ for all } j \in \{0, 1, ..., |d| - 1\}.
\end{equation}
\end{prog}

The constraint (\ref{eq:monotonicity_constr}) is exactly the monotonicity constraint (\ref{eq:monotonicity}) for the falling rule lists that are compatible with $D$. The constraint (\ref{eq:antecedent_constr}) limits the choice of antecedents. An instance of Program \ref{prog:frl} is defined by the tuple $(D, A, w, C)$.

\section{ALGORITHM}


In this section, we outline a Monte-Carlo search algorithm, Algorithm FRL, based on Program \ref{prog:frl}, for learning compatible falling rule lists from data. Given an instance $(D, A, w, C)$ of Program \ref{prog:frl}, the algorithm constructs a compatible falling rule list $d$ in each iteration, while keeping track of the falling rule list $d^*$ that has the smallest objective value $L_{\text{best}} = L(d^*, D, \tau, w, C)$ among all the falling rule lists that the algorithm has constructed so far. At the end of $T$ iterations, the algorithm outputs the falling rule list that has the smallest objective value out of the $T$ lists it has constructed.

In the process of constructing a falling rule list $d$, the algorithm chooses the antecedents successively, and uses various properties of Program \ref{prog:frl}, presented in Section \ref{section:prefix_bound}, to prune the search space. In particular, when the algorithm is choosing the $p$-th antecedent in $d$, it considers only those antecedents $A_l \in A$ satisfying the following conditions: (1) the inclusion of $A_l$ as the $p$-th antecedent in $d$ gives rise to a rule $(a_p^{(d)}, \alpha_p^{(d, D)})$ that respects the monotonicity constraint $\alpha_p^{(d, D)} \leq \alpha_{p-1}^{(d, D)}$ and the necessary condition for optimality $\alpha_p^{(d, D)} > 1/(1+w)$ (Corollary \ref{cor:candidate}), and (2) the inclusion of $A_l$ as the $p$-th antecedent in $d$ gives rise to a prefix $e'$ such that $e'$ is feasible for Program \ref{prog:frl} under the training data $D$ (Proposition \ref{prop:feasible}), and the best possible objective value $L^*(e', D, w, C)$ achievable by any falling rule list that begins with $e'$ and is compatible with $D$ (Theorem \ref{thm:prefix_bound}) is less than the current best objective value $L_{\text{best}}$. The algorithm terminates the construction of $d$ if Inequality (\ref{eq:terminating_condition}) in Theorem \ref{thm:prefix_bound} holds. The details of the algorithm can be found in the supplementary material.

\section{PREFIX BOUND}\label{section:prefix_bound}


The goal of this section is to find a lower bound on the objective value of any compatible falling rule list that begins with a given compatible prefix, which we call a {\it prefix bound}, and to prove the various results used in the algorithm. To derive this prefix bound, we first introduce the concept of a feasible prefix, with which it is possible to construct a compatible falling rule list from data.

\begin{defn}\label{defn:feasible}
Given the training data $D$ and the set of antecedents $A$, a prefix $e$ is feasible for Program \ref{prog:frl} under the training data $D$ and the set of antecedents $A$ if $e$ is compatible with $D$, and there exists a falling rule list $d$ such that $d$ is compatible with $D$, the antecedents of $d$ come from $A$, and $d$ begins with $e$.
\end{defn}

The following proposition gives necessary and sufficient conditions for a prefix $e$ to be feasible.

\begin{prop}\label{prop:feasible}
Given the training data $D$, the set of antecedents $A$, and a prefix $e$ that is compatible with $D$ and satisfies $a_j^{(e)} \in A$ for all $j \in \{0, 1, ..., |e| - 1\}$ and $\alpha_{k-1}^{(e, D)} \geq \alpha_k^{(e, D)}$ for all $k \in \{1, 2, ..., |e|-1\}$, the following statements are equivalent: (1) $e$ is feasible for Program \ref{prog:frl} under $D$ and $A$; (2) $\tilde{\alpha}_{e, D} \leq \alpha_{|e|-1}^{(e, D)}$ holds; (3) $\tilde{n}^-_{e, D} \geq ((1/\alpha_{|e|-1}^{(e, D)})-1)\tilde{n}^+_{e, D}$ holds.
\end{prop}

We now introduce the concept of a hypothetical rule list, whose antecedents do not need to come from the pre-determined set of antecedents $A$. 

\begin{defn}\label{defn:hypothetical_rl}
Given a pre-determined set of antecedents $A$, a hypothetical rule list with respect to $A$ is a rule list that contains an antecedent that is not in $A$.
\end{defn}


We need the following lemma to prove the necessary condition for optimality (Corollary \ref{cor:candidate}), and to derive a prefix bound (Theorem \ref{thm:prefix_bound}).

\begin{lem}\label{lem:1_more_rule}
Suppose that we are given an instance $(D, A, w, C)$ of Program \ref{prog:frl}, a prefix $e$ that is feasible for Program \ref{prog:frl} under $D$ and $A$, and a (possibly hypothetical) falling rule list $d$ that begins with $e$ and is compatible with $D$. Then there exists a falling rule list $d'$, possibly hypothetical with respect to $A$, such that $d'$ begins with $e$, has at most one more rule (excluding the final else clause) following $e$, is compatible with $D$, and satisfies
\begin{equation*}
L(d', D, 1/(1+w), w, C) \leq L(d, D, 1/(1+w), w, C).
\end{equation*}
As a special case, if either $\alpha_j^{(d, D)} > 1/(1+w)$ holds for all $j \in \{|e|, |e|+1, ..., |d|\}$, or $\alpha_j^{(d, D)} \leq 1/(1+w)$ holds for all $j \in \{|e|, |e|+1, ..., |d|\}$, then the falling rule list $\bar{e} = \{e, \tilde{\alpha}_{e, D}\}$ (i.e. the falling rule list in which the final else clause follows immediately the prefix $e$, and the probability estimate of the final else clause is $\tilde{\alpha}_{e, D}$) is compatible with $D$ and satisfies $L(\bar{e}, D, 1/(1+w), w, C) \leq L(d, D, 1/(1+w), w, C)$.
\end{lem}

A consequence of the above lemma is that an optimal solution for a given instance $(D, A, w, C)$ of Program \ref{prog:frl} should not have any antecedent whose empirical positive proportion falls below $1/(1+w)$.

\begin{cor}\label{cor:candidate}
If $d^*$ is an optimal solution for a given instance $(D, A, w, C)$ of Program \ref{prog:frl}, then we must have $\alpha_j^{(d^*, D)} > 1/(1+w)$ for all $j \in \{0, 1, ..., |d^*| - 1\}$.
\end{cor}

Another implication of Lemma \ref{lem:1_more_rule} is that the objective value of any compatible falling rule list that begins with a given prefix $e$ cannot be less than a lower bound on the objective value of any compatible falling rule list that begins with the same prefix $e$, and has at most one more rule (excluding the final else clause) following $e$. This leads to the following theorem.

\begin{thm}\label{thm:prefix_bound}
Suppose that we are given an instance $(D, A, w, C)$ of Program \ref{prog:frl} and a prefix $e$ that is feasible for Program \ref{prog:frl} under $D$ and $A$. Then any falling rule list $d$ that begins with $e$ and is compatible with $D$ satisfies
\begin{equation*}
L(d, D, 1/(1+w), w, C) \geq L^*(e, D, w, C),
\end{equation*}
where
\begin{equation}\label{eq:prefix_bound}
\begin{split}
&L^*(e, D, w, C) = L(e, D, 1/(1+w), w, C) + \\&\min\left(\frac{1}{n}\left(\frac{1}{\alpha_{|e|-1}^{(e, D)}} - 1\right)\tilde{n}^+_{e, D} + C, \frac{w}{n}\tilde{n}^+_{e, D}, \frac{1}{n}\tilde{n}^-_{e, D}\right)
\end{split}
\end{equation}
is a lower bound on the objective value of any compatible falling rule list that begins with $e$, under the instance $(D, A, w, C)$ of Program \ref{prog:frl}. We call $L^*(e, D, w, C)$ the prefix bound for $e$. Further, if
\begin{equation}\label{eq:terminating_condition}
C \geq \min\left(\frac{w}{n}\tilde{n}^+_{e, D}, \frac{1}{n}\tilde{n}^-_{e, D}\right) - \frac{1}{n}\left(\frac{1}{\alpha_{|e|-1}^{(e, D)}} - 1\right)\tilde{n}^+_{e, D}
\end{equation}
holds, then the compatible falling rule list $\bar{e} = \{e, \tilde{\alpha}_{e, D}\}$, where the prefix $e$ is followed directly by the final else clause, satisfies $L(\bar{e}, D, 1/(1+w), w, C) = L^*(e, D, w, C)$.
\end{thm}

The results presented in this section are used in Algorithm FRL to prune the search space. The proofs can be found in the supplementary material.

\section{SOFTLY FALLING RULE LISTS}


Program \ref{prog:frl} and Algorithm FRL have some limitations. Let us consider a toy example, where we have a training set $D$ of $19$ instances, with $14$ positive and $5$ negative instances. Suppose that we have an antecedent $A_1$ that is satisfied by $8$ positive and $3$ negative training instances. If $A_1$ were to be the first rule of a falling rule list $d$ that is compatible with $D$, we would obtain a prefix $e = \{(A_1, 8/11)\}$. However, the empirical positive proportion after the prefix $e$ is $\tilde{\alpha}_{e, D} = 6/8 > 8/11$. This violates (2) in Proposition \ref{prop:feasible}, so $e$ is not a feasible prefix for Program \ref{prog:frl} under the training data $D$. In fact, if every antecedent in $A$ is satisfied by $8$ positive and $3$ negative instances in the training set $D$, then the only possible compatible falling rule list we can learn using Algorithm FRL is the trivial falling rule list, which has only the final else clause. At the same time, if we consider the rule list $d = \{(A_1, 8/11), 6/8\}$, which is compatible with the given toy dataset $D$ but is not a falling rule list, we may notice that the two probability estimates in $d$ are quite close to each other -- it is very likely that the difference between them is due to sampling variability in the dataset itself.

The two limitations of Program \ref{prog:frl} and Algorithm FRL -- the potential non-existence of a feasible non-trivial solution and the rigidness of using empirical positive proportions as probability estimates -- motivate us to formulate a new optimization program for learning ``softly'' falling rule lists, where we remove the monotonicity constraint and instead introduce a penalty term in the objective function that penalizes violations of the monotonicity constraint (\ref{eq:monotonicity_constr}) in Program \ref{prog:frl}. More formally, define a softly falling rule list as a rule list of Equation (\ref{eq:rl}) with $\hat{\alpha}_j^{(d)} = \min_{k \leq j} \alpha_k^{(d, D)}$. Note that any rule list $d$ that is compatible with the given training data $D$ can be turned into a softly falling rule list by setting $\hat{\alpha}_j^{(d)} = \min_{k \leq j} \alpha_k^{(d, D)}$. Hence, we can learn a softly falling rule list by first learning a compatible rule list with the ``softly falling objective'' (denoted by $\tilde{L}$ below), and then transforming the rule list into a softly falling rule list. Let
\begin{equation*}
\begin{split}
&\quad\tilde{L}(d, D, \tau, w, C, C_1) \\
&= L(d, D, \tau, w, C) +C_1\sum_{j=0}^{|d|}\lfloor\alpha_j^{(d, D)} - \min_{k<j} \alpha_k^{(d, D)}\rfloor_+
\end{split}
\end{equation*}
\begin{equation*}
\begin{split}
&\text{and } \tilde{L}(e, D, \tau, w, C, C_1) \\
&= L(e, D, \tau, w, C) + C_1\sum_{j=0}^{|e|-1}\lfloor\alpha_j^{(e, D)} - \min_{k<j} \alpha_k^{(e, D)}\rfloor_+
\end{split}
\end{equation*}
be the regularized empirical risk of misclassification by a rule list $d$ and by a prefix $e$, respectively, with a penalty term that penalizes violations of monotonicity in the empirical positive proportions of the antecedents in $d$ and in $e$, respectively. We call $\tilde{L}$ the softly falling objective function, set the threshold $\tau = 1/(1+w)$ as before, and obtain the following optimization program:

\begin{prog}[Learning compatible rule lists with the softly falling objective]\label{prog:soft_frl}
\begin{equation*}
\begin{split}
\min_{d \in \mathcal{D}(\mathcal{X}, D)}& \tilde{L}(d, D, 1/(1+w), w, C, C_1) \\
\text{ subject to } \quad a_j^{(d)} &\in A, \text{ for all } j \in \{0, 1, ..., |d| - 1\}.
\end{split}
\end{equation*}
\end{prog}

An instance of Program \ref{prog:soft_frl} is defined by the tuple $(D, A, w, C, C_1)$. Similarly, we have a Monte-Carlo search algorithm, Algorithm softFRL, based on Program \ref{prog:soft_frl}, for learning softly falling rule lists from data. Given an instance $(D, A, w, C, C_1)$ of Program \ref{prog:soft_frl}, this algorithm searches through the space of rule lists that are compatible with $D$ and finds a compatible rule list whose antecedents come from $A$, and whose objective value is the smallest among all the rule lists that the algorithm explores. It then turns this compatible rule list into a softly falling rule list. In the search phase, the algorithm uses the following prefix bound (Theorem \ref{thm:prefix_bound_soft_frl}) to prune the search space of compatible rule lists. The details of Algorithm softFRL and the proof of Theorem \ref{thm:prefix_bound_soft_frl} can be found in the supplementary material.

\begin{thm}\label{thm:prefix_bound_soft_frl}
Suppose that we are given an instance $(D, A, w, C, C_1)$ of Program 5.1 and a prefix $e$ that is compatible with $D$. Then any rule list $d$ that begins with $e$ and is compatible with $D$ satisfies
\begin{equation*}
\tilde{L}(d, D, 1/(1+w), w, C, C_1) \geq \tilde{L}^*(e, D, w, C, C_1),
\end{equation*}
where
\begin{equation}\label{eq:prefix_bound_soft_frl}
\begin{split}
\tilde{L}^*&(e, D, w, C, C_1) = \tilde{L}(e, D, 1/(1+w), w, C, C_1) \\
+ &\min\left(\frac{1}{n}\left(\frac{1}{\alpha_{\min}^{(e, D)}} - 1\right)\tilde{n}^+_{e, D} + C \right.\\& \left. + C_1\lfloor\tilde{\alpha}_{e, D} - \alpha_{\min}^{(e, D)}\rfloor_+ + \frac{w}{n}\tilde{n}^+_{e, D}\mathds{1}[\tilde{\alpha}_{e, D} \geq \alpha_{\min}^{(e, D)}], \right.\\
    & \left. \inf_{\beta: \zeta < \beta \leq 1} g(\beta), \frac{w}{n}\tilde{n}^+_{e, D} + C_1\lfloor\tilde{\alpha}_{e, D} - \alpha_{\min}^{(e, D)}\rfloor_+, \right. \\
                   & \left. \vphantom{\frac{1}{\alpha_{|e|-1}^{(e, D)}} - 1} \frac{1}{n}\tilde{n}^-_{e, D} + C_1\lfloor\tilde{\alpha}_{e, D} - \alpha_{\min}^{(e, D)}\rfloor_+\right)
\end{split}
\end{equation}
is a lower bound on the objective value of any compatible rule list that begins with $e$, under the instance $(D, A, w, C, C_1)$ of Program 5.1. In Equation (\ref{eq:prefix_bound_soft_frl}), $\alpha_{\min}^{(e, D)}$, $\zeta$, and $g$ are defined by
\begin{equation*}
\alpha_{\min}^{(e, D)} = \min_{k < |e|} \alpha_k^{(e, D)},
\end{equation*}
\begin{equation*}
\zeta = \max(\alpha_{\min}^{(e, D)}, \tilde{\alpha}_{e, D}, 1/(1+w)),
\end{equation*}
\begin{equation*}
g(\beta) = \frac{1}{n}\left(\frac{1}{\beta} - 1\right)\tilde{n}^+_{e, D} + C + C_1(\beta - \alpha_{\min}^{(e, D)}).
\end{equation*}
Note that $\inf_{\beta: \zeta < \beta \leq 1} g(\beta)$ can be computed analytically: $\inf_{\beta: \zeta < \beta \leq 1} g(\beta) = g(\beta^*)$ if $\beta^* = \sqrt{\tilde{n}^+_{e, D}/(C_1 n)}$ satisfies $\zeta < \beta^* \leq 1$, and $\inf_{\beta: \zeta < \beta \leq 1} g(\beta) = \min(g(\zeta), g(1))$ otherwise.
\end{thm}

\section{EXPERIMENTS}


\begin{figure*}[h!]
    \centering
    \begin{subfigure}[b]{0.36\textwidth}
        \includegraphics[clip, trim={0.5cm 0cm 1.5cm 1cm}, width=\textwidth]{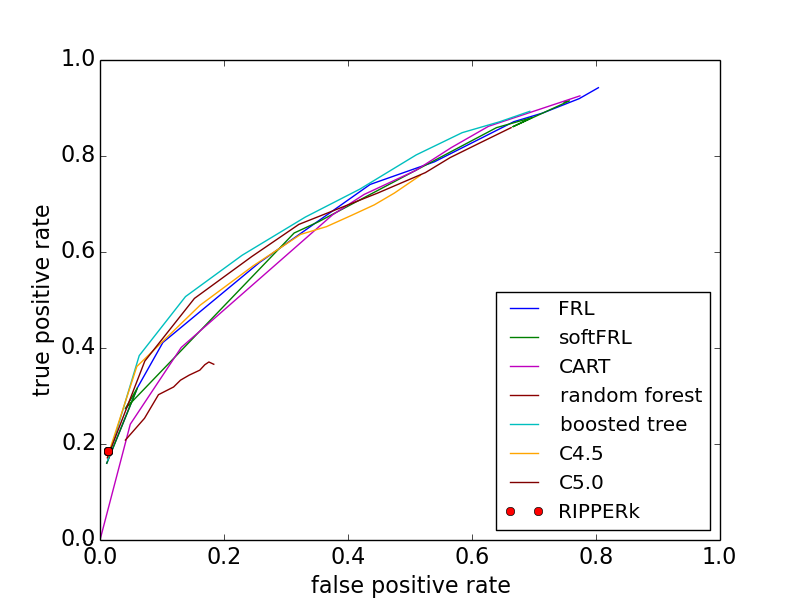}
        \caption{ROC curves on the test set using different $w$ values for one training-test split}
        \label{fig:roc}
    \end{subfigure}
    ~
    \centering
    \begin{subfigure}[b]{0.3\textwidth}
        \includegraphics[clip, trim={0.75cm 0cm 2cm 1cm}, width=\textwidth, height=3.5cm]{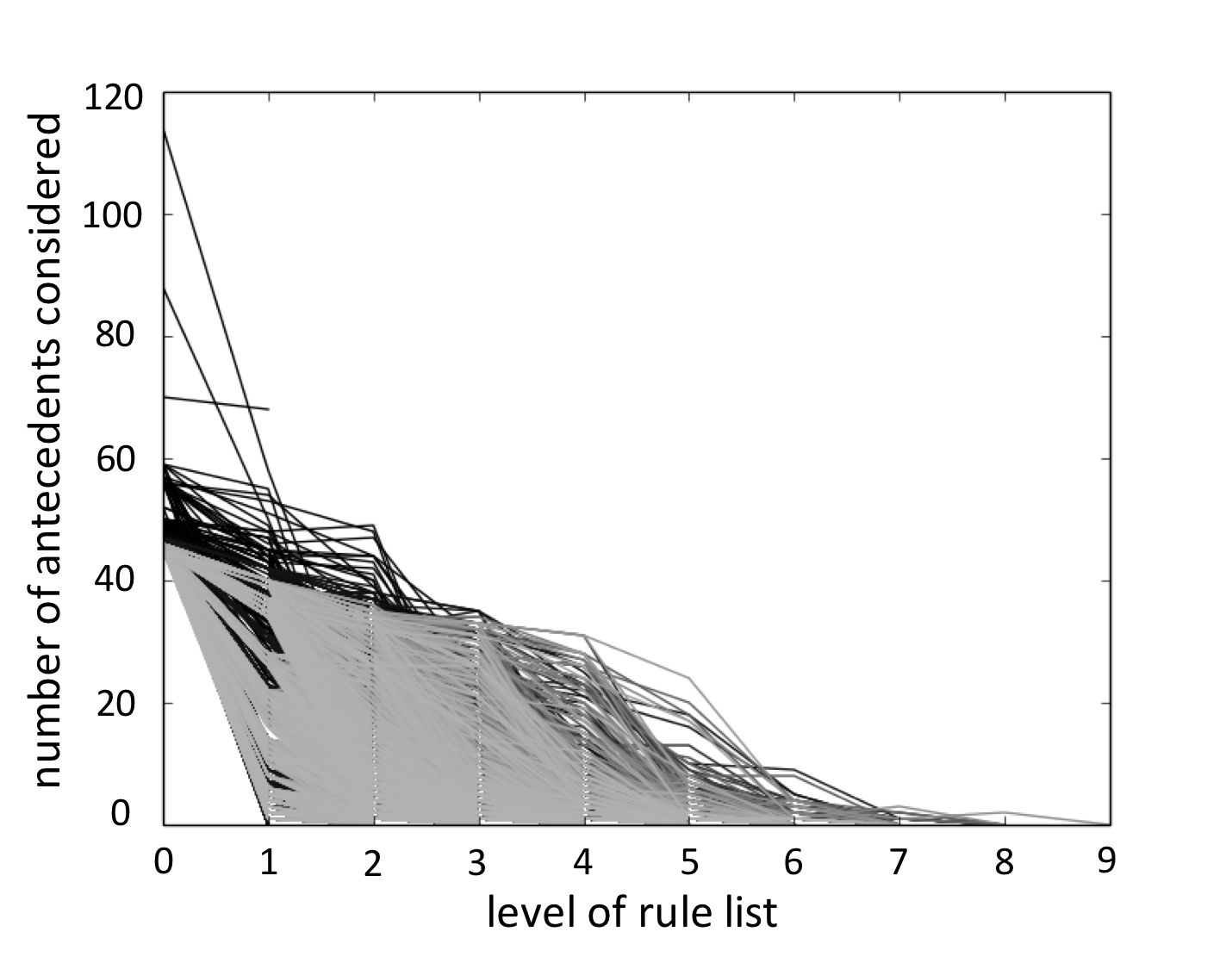}
        \caption{Number of antecedents considered by Algorithm FRL}
        \label{fig:num_antecedents_frl}
    \end{subfigure}
    ~
    \centering
    \begin{subfigure}[b]{0.3\textwidth}
        \includegraphics[clip, trim={0.75cm 0cm 2cm 1cm}, width=\textwidth, height=3.5cm]{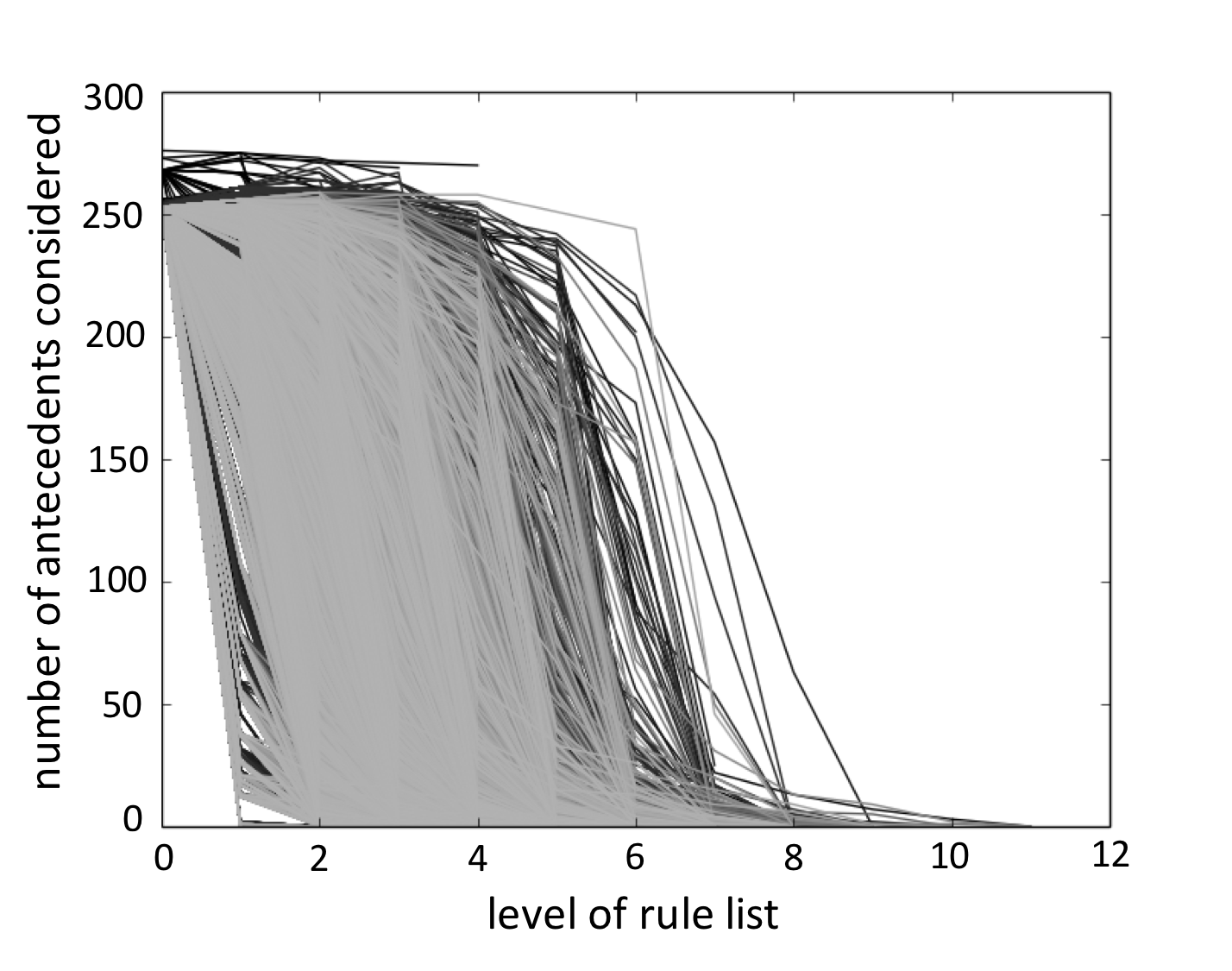}
        \caption{Number of antecedents considered by Algorithm softFRL}
        \label{fig:num_antecedents_soft_frl}
    \end{subfigure}
    \caption{Experimental Results}\label{fig:exp_results}
\end{figure*}

\begin{figure*}[h!]
    \centering
    \begin{subfigure}[b]{0.23\textwidth}
        \includegraphics[clip, trim={3cm 0cm 2cm 0cm}, width=\textwidth, height=2.75cm]{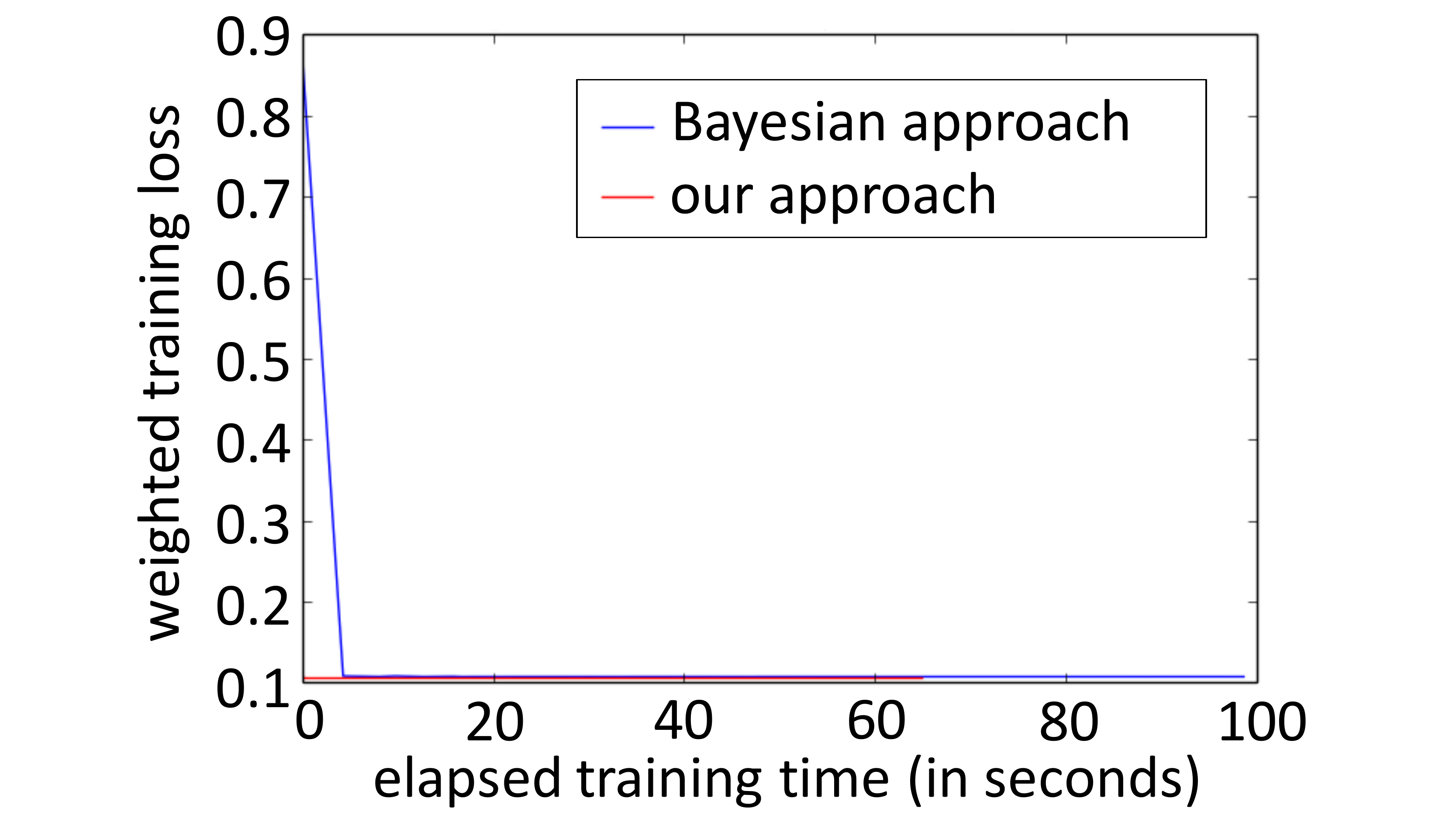}
        \caption{positive weight $w = 1$}
        \label{fig:comp_w=1}
    \end{subfigure}
    ~
    \centering
    \begin{subfigure}[b]{0.23\textwidth}
        \includegraphics[clip, trim={3cm 0cm 2cm 0cm}, width=\textwidth, height=2.75cm]{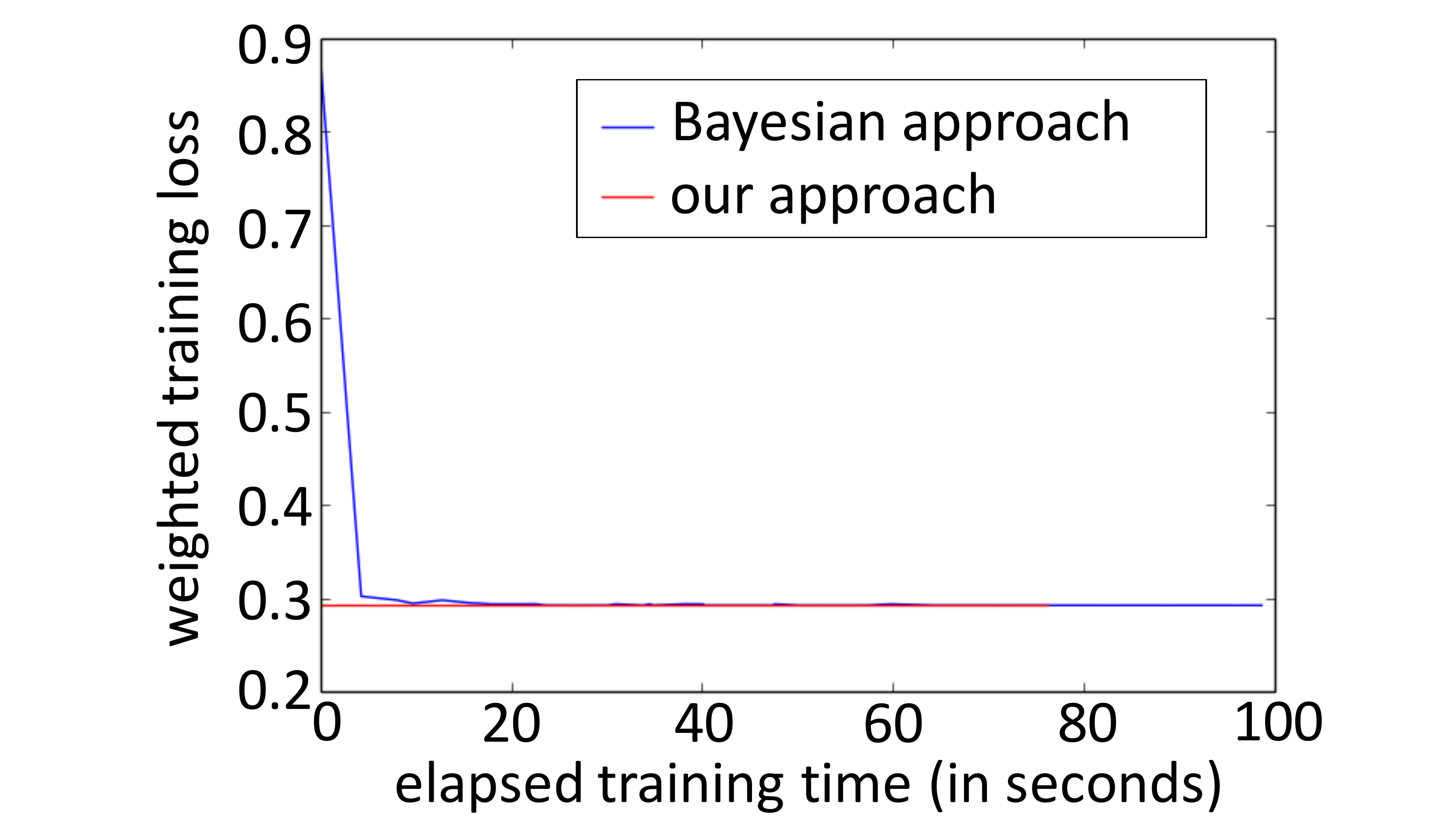}
        \caption{positive weight $w = 3$}
        \label{fig:comp_w=3}
    \end{subfigure}
    ~
    \centering
    \begin{subfigure}[b]{0.23\textwidth}
        \includegraphics[clip, trim={3cm 0cm 2cm 0cm}, width=\textwidth, height=2.75cm]{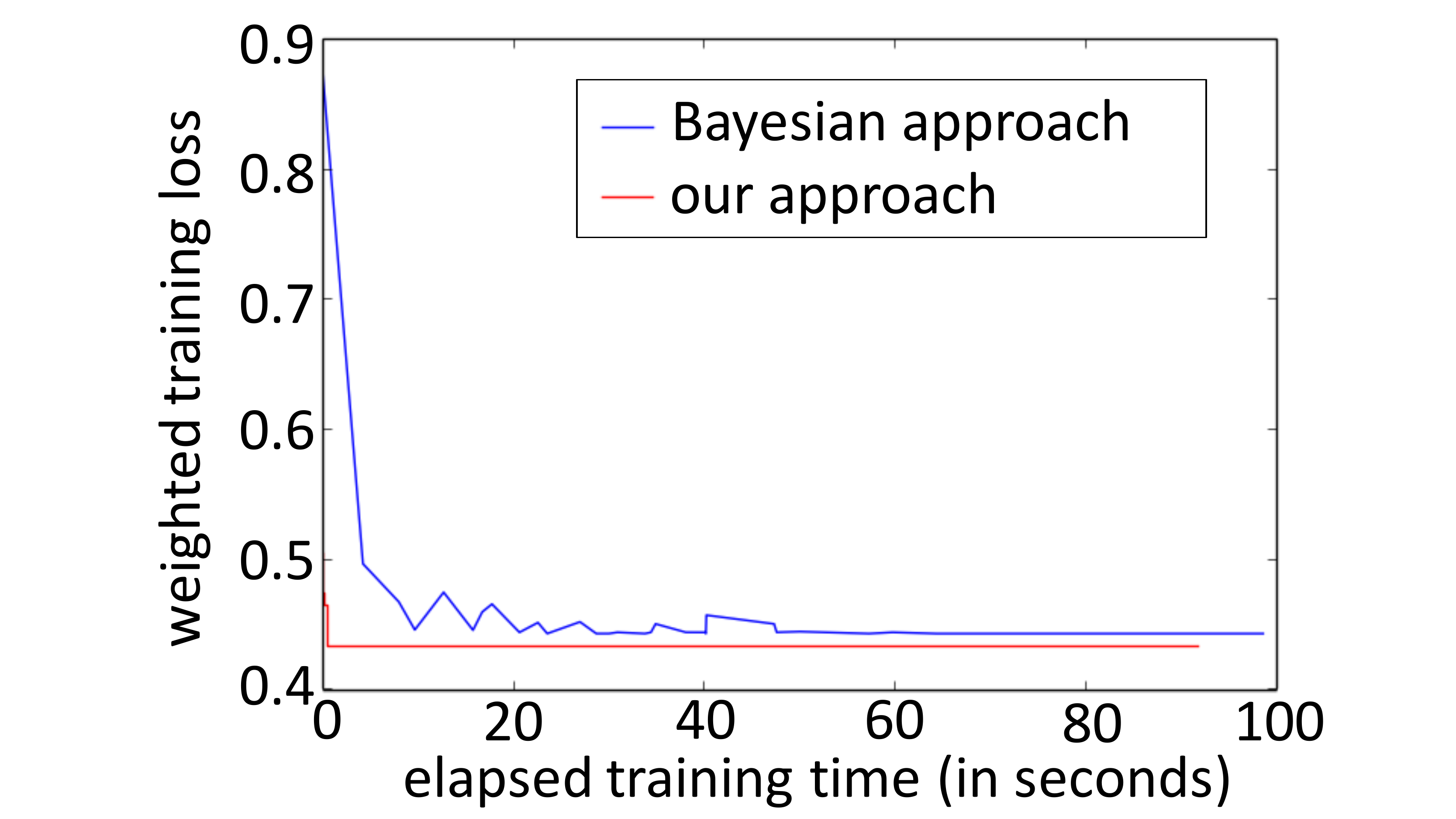}
        \caption{positive weight $w = 5$}
        \label{fig:comp_w=5}
    \end{subfigure}
    ~
    \centering
    \begin{subfigure}[b]{0.23\textwidth}
        \includegraphics[clip, trim={3cm 0cm 2cm 0cm}, width=\textwidth, height=2.75cm]{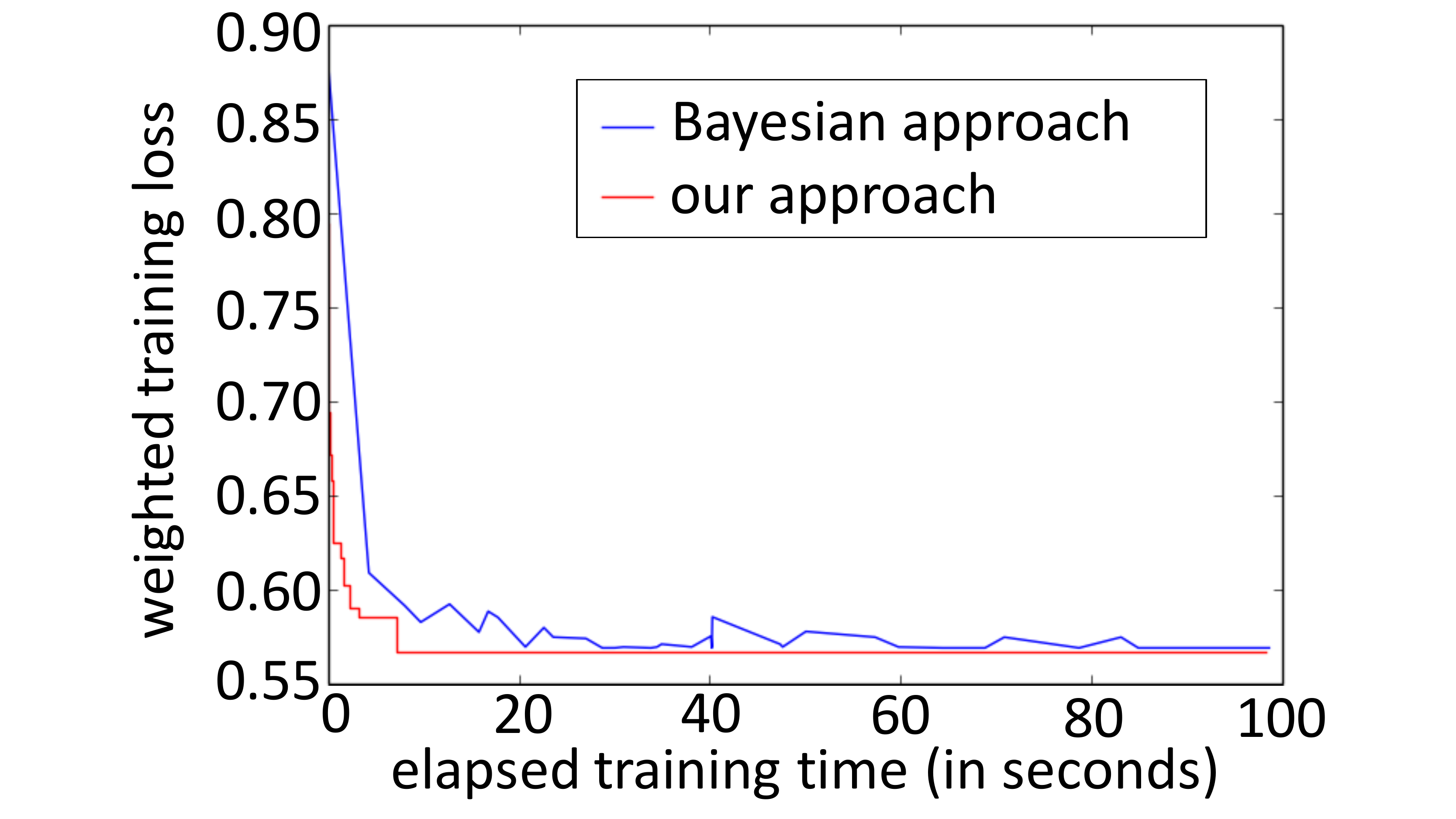}
        \caption{positive weight $w = 7$}
        \label{fig:comp_w=7}
    \end{subfigure}
    \caption{Plots of Weighted Training Loss over Real Runtime for Bayesian Approach and Algorithm FRL}\label{fig:compare_with_BayesianFRL}
\end{figure*}

In this section, we demonstrate our algorithms for learning falling rule lists using a real-world application -- learning the conditions that are predictive of the success of a bank marketing effort, from previous bank marketing campaign data. We used the public bank-full dataset \citep{Moro}, which contains $45211$ observations, with $12$ predictor variables that were discretized. We used the frequent pattern growth (FP-growth) algorithm \citep{Han} to generate the set of antecedents $A$ from the dataset. For reasons of model interpretability and generalizability, we included in $A$ the antecedents that have at most $2$ predicates, and have at least $10\%$ support within the data that are labeled positive or within the data that are labeled negative. Besides the FP-growth algorithm, there is a vast literature on rule mining algorithms \citep[e.g.][]{Agrawal,Han2,Landwehr}, and any of these can be used to produce antecedents for our algorithms.

The bank-full dataset is imbalanced -- there are only $5289$ positive instances out of $45211$ observations. A trivial model that always predicts the negative outcome for a bank marketing campaign will achieve close to $90\%$ accuracy on this dataset, but it will not be useful for the bank to understand what makes a marketing campaign successful. Moreover, when predicting if a future campaign will be successful in finding a client, the bank cares more about ``getting the positive right'' than about ``getting the negative right'' -- a false negative means a substantial loss in revenue, while a false positive incurs little more than some phone calls.


We compared our algorithms with other classification algorithms in a cost-sensitive setting, where a false negative and a false positive have different costs of misclassification. We generated five random splits into a training and a test set, where $80\%$ of the observations in the original bank-full dataset were placed into the training set. For each training-test split, and for each positive class weight $w \in \{1, 3, 5, 7, 9, 11, 13, 15, 17, 19\}$, we learned from the training set: (1) a falling rule list $d$, which is treated as a classifier $\tilde{d}_{1/(1+w)}$, using Algorithm FRL with $C = 0.000001$ (which is small enough so that no training accuracy will be sacrificed for sparsity), (2) a softly falling rule list $d'$, which is treated as a classifier $\tilde{d}'_{1/(1+w)}$, using Algorithm softFRL with $C = 0.000001$ and $C_1 = 0.5$, (3) three decision trees using cost-sensitive CART \citep{Breiman}, cost-sensitive C4.5 \citep{Quinlan2}, and cost-sensitive C5.0 \citep{Quinlan3}, respectively, (6) a random forest \citep{Breiman2} of decision trees trained with cost-sensitive CART, (7) a boosted tree classifier using AdaBoost \citep{Freund} on trees trained with cost-sensitive CART, and (8) a decision list using RIPPER$k$ \citep{Cohen}, and we computed the true positive rate and the false positive rate on the test set for each classifier. For each split and for each algorithm, we plotted a receiver operating characteristic (ROC) curve on the test set using different values of $w$. Figure \ref{fig:roc} shows the ROC curves for one of the training-test splits. The ROC curves for the other training-test splits can be found in the supplementary material. Note that since RIPPER$k$ is not a cost-sensitive algorithm, its ROC curve based on different $w$ values has only a single point. As we can see, the curves in Figure \ref{fig:roc} lie close to each other. This demonstrates the effectiveness of our algorithms in producing falling rule lists that, when used as classifiers, are comparable with classifiers produced by other widely used classification algorithms, in a cost-sensitive setting. This is possibly surprising since our models are much more constrained than other classification methods.

We also plotted the number of antecedents considered by Algorithm FRL and Algorithm softFRL in the process of constructing a rule list at each iteration (Figures \ref{fig:num_antecedents_frl} and \ref{fig:num_antecedents_soft_frl}), when we applied the two algorithms to the entire dataset. Each curve in either plot corresponds to a rule list constructed in an iteration of the appropriate algorithm. The intensity of the curve is inversely proportional to the iteration number -- the larger the iteration number, the lighter the curve is. The number of antecedents considered by Algorithm FRL stays below $60$ in all but a few early iterations (despite a choice of $276$ antecedents available), and the number considered by either algorithm generally decreases drastically in each iteration after three or four antecedents have been chosen. The curves generally become lighter as we move vertically down the plots, indicating that as we find better rule lists, there are less antecedents to consider at each level. Algorithm softFRL needs to consider more antecedents in general since the search space is less constrained. All of these demonstrate that the prefix bounds we have derived for our algorithms are effective in excluding a large portion of the search space of rule lists. The supplementary material contains more rule lists created using our algorithms with different parameter values.





Since this paper was directly inspired by Wang and Rudin (\citeyear{Wang}), who proposed a Bayesian approach to learning falling rule lists, we conducted a set of experiments comparing their work to ours. We trained falling rule lists on the entire bank-full dataset using both the Bayesian approach and our optimization approach, and plotted the weighted training loss over real runtime for each positive class weight $w \in \{1, 3, 5, 7\}$ with the threshold set to $1/(1+w)$ (By Theorem 2.8, this is the threshold with the least weighted training loss for any given rule list). Since we want to focus our experiments on the efficiency of searching the model space, the runtimes recorded do not include the time for mining the antecedents. Note that the Bayesian approach is not cost-sensitive, and does not optimize the weighted training loss directly. However, in many real-life applications such as predicting the success of a future marketing campaign, it is desirable to minimize the expected weighted loss. Therefore, it is reasonable to compare the two approaches using the weighted training loss to demonstrate the advantages of our optimization approach. We compared the Bayesian approach only with Algorithm FRL, because both methods strictly enforce the monotonicity constraint on the positive proportions of the training data that are classified into each rule. Softly falling rule lists do not strictly enforce the monotonicity constraint, and are therefore not used for comparison. Figure \ref{fig:compare_with_BayesianFRL} shows the plots of the weighted training loss over real runtime. Due to the random nature of both approaches, the experiments were repeated several times -- more plots of the weighted training loss over real runtime for different trials of the same experiment, along with falling rule lists created using both approaches, can be found in the supplementary material. As shown in Figure \ref{fig:compare_with_BayesianFRL}, our optimization approach tends to find a falling rule list with a smaller weighted training loss faster than the Bayesian approach. This is not too surprising because in our approach, the search space is made substantially smaller by the tight bounds presented here, whereas in the original Bayesian approach, there are no tight bounds on optimal solutions to restrict the search space -- even if we constructed bounds for the original Bayesian approach, they would involve loose approximations to gamma functions.

\section{CONCLUSION}


We have proposed an optimization approach to learning falling rule lists and softly falling rule lists, along with Monte-Carlo search algorithms that use bounds on the optimal solution to prune the search space. A recent work by \cite{Angelino} on (non-falling) rule lists showed that it is possible to exhaustively optimize an objective over rule lists, indicating that the space of lists is not as large as one might think. Our search space is a dramatically constrained version of their search space, allowing us to reasonably believe that it can be searched exhaustively. Unfortunately, almost none of the logic of \cite{Angelino} can be used here. 
Indeed, introducing the falling constraint or the monotonicity penalty changes the nature of the problem, and the bounds in our work are entirely different. The algorithm of \cite{Angelino} is not cost-sensitive, which led in this work to another level of complexity for the bounds.

Falling rule lists are optimized for ease-of-use -- users only need to check a small number of conditions to determine whether an observation is in a high risk or high probability subgroup. As pointed out by \cite{Wang}, the monotonicity in probabilities in falling rule lists allows doctors to identify the most at-risk patients easily. Typical decision tree methods (CART, C4.5, C5.0) do not have the added interpretability that comes from the falling constraint in falling rule lists: one may have to check many conditions in a decision tree to determine whether an observation is in a high risk or high probability subgroup -- even if the decision tree has a small depth, it is possible that high risk subgroups are in different parts of the tree, so that one still has to check many conditions in order to find high risk subgroups. In this sense, falling rule lists and softly falling rule lists are as sparse as we need them to be, and they can provide valuable insight into data.

\noindent\textbf{Supplementary Material and Code:} The supplementary material and code are available at \url{https://github.com/cfchen-duke/FRLOptimization}.

\subsubsection*{Acknowledgments}

This work was sponsored in part by MIT Lincoln Laboratory.

\subsubsection*{References}
\begingroup
\renewcommand{\section}[2]{}%

\bibliography{frl_opt}{}
\bibliographystyle{apalike}

\endgroup





\onecolumn

\section*{Supplementary Material}

\section{Algorithm FRL}


In this section, we present Algorithm FRL in detail. Given an instance $(D, A, w, C)$ of Program 2.9, the algorithm searches through the space of falling rule lists that are compatible with $D$ and outputs a compatible falling rule list that respects the constraints of Program 2.9, and whose objective value is the smallest among all the falling rule lists that the algorithm explores. It does so by iterating over $T$ steps, in each of which the algorithm constructs a compatible falling rule list $d$, while keeping track of the falling rule list $d^*$ that has the smallest objective value $L_{\text{best}} = L(d^*, D, 1/(1+w), w, C)$ among all the falling rule lists that the algorithm has constructed so far. At the end of $T$ iterations, the algorithm outputs the falling rule list that has the smallest objective value out of the $T$ lists it has constructed.

In the process of constructing a falling rule list $d$, the algorithm chooses the antecedents successively: first for the antecedent $a_0^{(d)}$ in the top rule, then for the antecedent $a_1^{(d)}$ in the next rule, and so forth. For each antecedent $a_j^{(d)}$ chosen, the algorithm also computes its empirical positive proportion $\alpha_j^{(d, D)}$. After $p$ rules have been constructed so that $d$ currently holds the prefix $e = \{(a_0^{(d)}, \alpha_0^{(d, D)}),$ $(a_1^{(d)}, \alpha_1^{(d, D)}), ..., (a_{p-1}^{(d)}, \alpha_{p-1}^{(d, D)})\}$, the algorithm either: (1) terminates the construction of $d$ by computing the empirical positive proportion after $e$, $\tilde{\alpha}_{e, D}$, and then adding to $d$ the final else clause with probability estimate $\tilde{\alpha}_{e, D}$, or (2) randomly picks an antecedent from a candidate set $S$ of possible next antecedents, computes its empirical positive proportion, and uses these as the next rule $(a_p^{(d)}, \alpha_p^{(d, D)})$ for $d$.

The algorithm uses various properties of Program 2.9, which are presented in Section 4, to prune the search space. More specifically, the algorithm terminates the construction of $d$ if Inequality (9) in Theorem 4.6 holds. Otherwise it either terminates the construction of $d$ with some probability, or proceeds to construct a candidate set $S$ of possible next antecedents, as follows. For every antecedent $A_l \in A$ that has not been chosen before, it constructs a candidate next rule $(a_p^{(d)}, \alpha_p^{(d, D)})$ by setting $a_p^{(d)} = A_l$ and computing $\alpha_p^{(d, D)}$ using Definition 2.5. The algorithm then checks if the monotonicity constraint $\alpha_p^{(d, D)} \leq \alpha_{p-1}^{(d, D)}$ and the necessary condition for optimality $\alpha_p^{(d, D)} > 1/(1+w)$ (Corollary 4.5) are satisfied, if the prefix $e' = \{e, (a_p^{(d)}, \alpha_p^{(d, D)})\}$ is feasible under Program 2.9 (i.e. whether there exists a compatible falling rule list that begins with the prefix $e'$) using Proposition 4.2, and if the best possible objective value $L^*(e', D, w, C)$ achievable by any falling rule list that begins with $e'$ and is compatible with $D$ (Theorem 4.6) is less than the current best objective value $L_{\text{best}} = L(d^*, D, 1/(1+w), w, C)$. If all of the above conditions are satisfied, the algorithm adds $A_l$ to $S$. Once the construction of $S$ is complete, the algorithm randomly chooses an antecedent $A_l \in S$ with probability $P(A_l|S, e, D)$ and uses this antecedent, together with its empirical positive proportion, as the next rule $(a_p^{(d)}, \alpha_p^{(d, D)})$ for $d$. If $S$ is empty, the algorithm terminates the construction of $d$. 

In practice, we define the probability $P(A_l|S, e, D)$ for $A_l \in S$ by first defining a curiosity function $f_{S, e, D}: S \rightarrow \mathbb{R}_{\geq 0}$ and then normalizing it:
\begin{equation*}
P(A_l|S, e, D) = \frac{f_{S, e, D}(A_l)}{\sum_{A_{l'}} f_{S, e, D}(A_{l'})}.
\end{equation*}
A possible choice of the curiosity function $f_{S, e, D}$ for use in Algorithm FRL is given by
\begin{equation}\label{eq:curiosity_algoFRL}
f_{S, e, D}(A_l) = \lambda\alpha(A_l, e, D) + (1 - \lambda)\frac{n^+(A_l, e, D)}{\tilde{n}^+_{e, D}},
\end{equation}
where $\alpha(A_l, e, D)$ is the empirical positive proportion of $A_l$, and $n^+(A_l, e, D)$ is the number of positive training inputs captured by $A_l$, should $A_l$ be chosen as the next antecedent after the prefix $e$. The curiosity function $f_{S, e, D}$ given by (\ref{eq:curiosity_algoFRL}) is a weighted sum of $\alpha(A_l, e, D)$ and $n^+(A_l, e, D)/\tilde{n}^+_{e, D}$ for each $A_l \in S$: the former encourages the algorithm to choose antecedents that have large empirical positive proportions, and the latter encourages the algorithm to choose antecedents that have large positive supports in the training data not captured by $e$. We used this curiosity function for Algorithm FRL in our experiments.

The pseudocode of Algorithm FRL is shown in Algorithm \ref{algo:frl}.

\begin{figure}[htbp]
\centering
\fbox{\parbox{0.95\linewidth}{
\begin{algorithm}[H]\label{algo:frl}
 \KwIn{an instance $(D, A, w, C)$ of Program 2.9}
 \KwResult{a falling rule list $d^*$ that are compatible with $D$ and whose antecedents come from $A$}
 initialize $d^* = \emptyset$, $L_{\text{best}} = \infty$\;
 \For{$t = 1, ..., T$}{
  set $p = -1$, $\alpha_p = 1$,
  $d = e = \emptyset$\;
 
  \While{Inequality (9) in Theorem 4.6 does not hold}{
   go to Terminate with some probability\;
   set $p = p + 1$, $S = \emptyset$\;
   \For{every antecedent $A_l \in A$ that is not in $d$}{
    set $a_p^{(d)} = A_l$, compute $\alpha_p^{(d, D)}$, and let $e' = \{e, (a_p^{(d)}, \alpha_p^{(d, D)})\}$\;
    \If{$\alpha_p^{(d, D)} \leq \alpha_{p-1}^{(d, D)}$, $\alpha_p^{(d, D)} > 1/(1+w)$, and $e'$ is feasible under Program 2.9}{
     compute $L^*(e', D, w, C)$ using Theorem 4.6\;
     \If{$L^*(e', D, w, C) < L(d^*, D, 1/(1+w), w, C)$}{
      add $A_l$ to $S$\;
     }
    }
   }
   
   \eIf{$S \neq \emptyset$}{
    choose an antecedent $A_l \in S$ with probability $P(A_l|S, e, D)$ according to a discrete probability distribution over $S$\;
    set $a_p^{(d)} = A_l$ and add $(a_p^{(d)}, \alpha_p^{(d, D)})$ to $d$\;
    set $e = d$\; // save the partially constructed list $d$ as the prefix $e$
    }{
    go to Terminate
   }
  }
  
  Terminate:
  terminate the construction of $d$, and compute $L(d, D, 1/(1+w), w, C)$\;
  \If{$L(d, D, 1/(1+w), w, C) < L_{\text{best}}$}{set $d^* = d$, $L_{\text{best}} = L(d, D, 1/(1+w), w, C)$\;}
  
 }
 \caption{Algorithm FRL}
\end{algorithm}
}}
\end{figure}

\section{Algorithm softFRL}


In this section, we present Algorithm softFRL in detail. Given an instance $(D, A, w, C, C_1)$ of Program 5.1, the algorithm searches through the space of rule lists that are compatible with $D$ and finds a compatible rule list whose antecedents come from $A$, and whose objective value is the smallest among all the rule lists that the algorithm explores. It does so by iterating over $T$ steps, in each of which the algorithm constructs a compatible rule list $d$, while keeping track of the rule list $d^*$ that has the smallest objective value $\tilde{L}_{\text{best}} = \tilde{L}(d^*, D, 1/(1+w), w, C, C_1)$ among all the rule lists that the algorithm has constructed so far. At the end of $T$ iterations, the algorithm transforms the rule list $d^*$ that has the smallest objective value out of the $T$ lists it has constructed, into a falling rule list by setting $\hat{\alpha}_j^{(d^*)} = \min_{k \leq j} \alpha_k^{(d^*, D)}$.

In the process of constructing a rule list $d$, the algorithm chooses the antecedents successively: first for the antecedent $a_0^{(d)}$ in the top rule, then for the antecedent $a_1^{(d)}$ in the next rule, and so forth. For each antecedent $a_j^{(d)}$ chosen, the algorithm also computes its empirical positive proportion $\alpha_j^{(d, D)}$. After $p$ rules have been constructed so that $d$ currently holds the prefix $e = \{(a_0^{(d)}, \alpha_0^{(d, D)}),$ $(a_1^{(d)}, \alpha_1^{(d, D)}), ..., (a_{p-1}^{(d)}, \alpha_{p-1}^{(d, D)})\}$, the algorithm either: (1) terminates the construction of $d$ by computing the empirical positive proportion after $e$, $\tilde{\alpha}_{e, D}$, and then adding to $d$ the final else clause with probability estimate $\tilde{\alpha}_{e, D}$, or (2) randomly picks an antecedent from a candidate set $S$ of possible next antecedents, computes its empirical positive proportion, and use these as the next rule $(a_p^{(d)}, \alpha_p^{(d, D)})$ for $d$.

The algorithm uses Theorem 5.2 to prune the search space. More specifically, the algorithm terminates the construction of $d$ if $\tilde{L}^*(e, D, w, C, C_1)$ defined by Equation (10) in Theorem 5.2 is equal to $\tilde{L}(\bar{e}, D, 1/(1+w), w, C, C_1)$, where $\bar{e} = \{e, \tilde{\alpha}_{e, D}\}$ is the compatible rule list in which the prefix $e$ is followed directly by the final else clause. The condition $\tilde{L}^*(e, D, w, C, C_1) = \tilde{L}(\bar{e}, D, 1/(1+w), w, C, C_1)$ implies that $\bar{e}$ is an optimal compatible rule list that begins with $e$. If we have $\tilde{L}^*(e, D, w, C, C_1) < \tilde{L}(\bar{e}, D, 1/(1+w), w, C, C_1)$ instead, the algorithm either terminates the construction of $d$ with some probability, or it proceeds to construct a candidate set $S$ of possible next antecedents, as follows. For every antecedent $A_l \in A$ that has not been chosen before, it constructs a candidate next rule $(a_p^{(d)}, \alpha_p^{(d, D)})$ by setting $a_p^{(d)} = A_l$ and computing $\alpha_p^{(d, D)}$ using Definition 2.5. The algorithm then checks if the best possible objective value $\tilde{L}^*(e', D, w, C, C_1)$ achievable by any rule list that begins with $e' = \{e, (a_p^{(d)}, \alpha_p^{(d, D)})\}$ and is compatible with $D$ (Theorem 5.2) is less than the current best objective value $\tilde{L}_{\text{best}} = \tilde{L}(d^*, D, 1/(1+w), w, C, C_1)$. If so, the algorithm adds $A_l$ to $S$. Once the construction of $S$ is complete, the algorithm randomly chooses an antecedent $A_l \in S$ with probability $P(A_l|S, e, D)$ and uses this antecedent, together with its empirical positive proportion, as the next rule $(a_p^{(d)}, \alpha_p^{(d, D)})$ for $d$. If $S$ is empty, the algorithm terminates the construction of $d$. 

In practice, we define the probability $P(A_l|S, e, D)$ for $A_l \in S$ by first defining a curiosity function $f_{S, e, D}: S \rightarrow \mathbb{R}_{\geq 0}$ and then normalizing it:
\begin{equation*}
P(A_l|S, e, D) = \frac{f_{S, e, D}(A_l)}{\sum_{A_{l'}} f_{S, e, D}(A_{l'})}.
\end{equation*}
A possible choice of the curiosity function $f_{S, e, D}$ for use in Algorithm softFRL is given by
\begin{equation}\label{eq:curiosity_algosoftFRL}
f_{S, e, D}(A_l) = \lambda\lfloor\min(\alpha(A_l, e, D), \frac{1.01}{0.01}\alpha_{\min}^{(e, D)} - \frac{1}{0.01}\alpha(A_l, e, D))\rfloor_+ + (1 - \lambda)\frac{n^+(A_l, e, D)}{\tilde{n}^+_{e, D}},
\end{equation}
where $\alpha_{\min}^{(e, D)} = \min_{k < |e|} \alpha_k^{(e, D)}$ is the minimum empirical positive proportion of the antecedents in the prefix $e$, $\alpha(A_l, e, D)$ is the empirical positive proportion of $A_l$, and $n^+(A_l, e, D)$ is the number of positive training inputs captured by $A_l$, should $A_l$ be chosen as the next antecedent after the prefix $e$. The curiosity function $f_{S, e, D}$ given by (\ref{eq:curiosity_algosoftFRL}) is a weighted sum of $\lfloor\min(\alpha(A_l, e, D), (1.01/0.01)\alpha_{|e|-1}^{(e, D)} - (1/0.01)\alpha(A_l, e, D))\rfloor_+$ and $n^+(A_l, e, D)/\tilde{n}^+_{e, D}$ for each $A_l \in S$: the former encourages the algorithm to choose antecedents that have large empirical positive proportions but do not violate the monotonicity constraint $\alpha(A_l, e, D) \leq \alpha_{\min}^{(e, D)}$ by more than $1\%$, and the latter encourages the algorithm to choose antecedents that have large positive supports in the training data not captured by $e$. We used this curiosity function for Algorithm softFRL in our experiments.

The pseudocode of Algorithm softFRL is shown in Algorithm \ref{algo:soft_frl}.

\begin{figure}[htbp]
\centering
\fbox{\parbox{0.95\linewidth}{
\begin{algorithm}[H]\label{algo:soft_frl}
 \KwIn{an instance $(D, A, w, C, C_1)$ of Program 5.1}
 \KwResult{a falling rule list $d^*$ whose antecedents come from $A$}
 initialize $d^* = \emptyset$, $\tilde{L}_{\text{best}} = \infty$\;
 \For{$t = 1, ..., T$}{
  set $p = -1$, $\alpha_p = 1$,
  $d = e = \emptyset$\;
 
  \While{$\tilde{L}^*(e, D, w, C, C_1) < \tilde{L}(\bar{e}, D, 1/(1+w), w, C, C_1)$}{
   go to Terminate with some probability\;
   set $p = p + 1$, $S = \emptyset$\;
   \For{every antecedent $A_l \in A$ that is not in $d$}{
    set $a_p^{(d)} = A_l$, compute $\alpha_p^{(d, D)}$, and let $e' = \{e, (a_p^{(d)}, \alpha_p^{(d, D)})\}$\;
    compute $\tilde{L}^*(e', D, w, C, C_1)$ using Theorem 5.2\;
    \If{$\tilde{L}^*(e', D, w, C, C_1) < \tilde{L}(d^*, D, 1/(1+w), w, C, C_1)$}{
     add $A_l$ to $S$\; 
    }
   }
   
   \eIf{$S \neq \emptyset$}{
    choose an antecedent $A_l \in S$ with probability $P(A_l|S, e, D)$ according to a discrete probability distribution over $S$\;
    set $a_p^{(d)} = A_l$ and add $(a_p^{(d)}, \alpha_p^{(d, D)})$ to $d$\;
    set $e = d$\; // save the partially constructed list $d$ as the prefix $e$
    }{
    go to Terminate
   }
  }
  
  Terminate:
  terminate the construction of $d$, and compute $\tilde{L}(d, D, 1/(1+w), w, C, C_1)$\;
  \If{$\tilde{L}(d, D, 1/(1+w), w, C, C_1) < \tilde{L}_{\text{best}}$}{set $d^* = d$, $\tilde{L}_{\text{best}} = \tilde{L}(d, D, 1/(1+w), w, C, C_1)$\;}
  
 }
 
 transform $d^*$ into a falling rule list by setting $\hat{\alpha}_j^{(d^*)} = \min_{k \leq j} \alpha_k^{(d^*, D)}$\;
 
 \caption{Algorithm softFRL}
\end{algorithm}
}}
\end{figure}

\section{Proofs of Theorem 2.8, Proposition 4.2, Lemma 4.4, Corollary 4.5, and Theorem 4.6}


{\bf Theorem 2.8.} Given the training data $D$, a rule list $d$ that is compatible with $D$, and the weight $w$ for the positive class, we have
\begin{equation*}
R(d, D, 1/(1+w), w) \leq R(d, D, \tau, w)
\end{equation*}
for all $\tau \geq 0$.

\begin{proof}
Suppose $\tau > 1/(1+w)$. Consider the $j$-th rule $(a_j^{(d)}, \alpha_j^{(d, D)})$ in $d$, whose antecedent captures $\alpha_j^{(d, D)}n_{j, d, D}$ positive training inputs and $(1-\alpha_j^{(d, D)})n_{j, d, D}$ negative training inputs. Let $R_j(d, D, \tau, w)$ denote the contribution by the $j$-th rule to $R(d, D, \tau, w)$, i.e.
\begin{equation}\label{eq:jth_contrib}
R_j(d, D, \tau, w) = \frac{1}{n}\left(w\sum_{\substack{i: y_i = 1 \wedge \\ \text{capt}(\mathbf{x}_i, d) = j}} \mathds{1}[\alpha_j^{(d, D)} \leq \tau] + \sum_{\substack{i: y_i = -1 \wedge \\ \text{capt}(\mathbf{x}_i, d) = j}} \mathds{1}[\alpha_j^{(d, D)} > \tau]\right) = \begin{cases}
\frac{1}{n}n^-_{j, d, D} &\text{ if } \alpha_j^{(d, D)} > \tau \\
\frac{w}{n}n^+_{j, d, D} &\text{ otherwise.}
\end{cases}
\end{equation}

{\bf Case 1.} $1/(1+w) < \alpha_j^{(d, D)} \leq \tau$. In this case, we have
\begin{align*}
R_j(d, D, 1/(1+w), w) &= \frac{1}{n}n^-_{j, d, D} \quad\text{(by the definition of } R_j \text{ in Equation (\ref{eq:jth_contrib}))} \\
&= \frac{1}{n}(n_{j, d, D} - n^+_{j, d, D}) \quad\text{(by the definition of } n^+_{j, d, D}, n^-_{j, d, D}, n_{j, d, D} \text{ in Definition 2.5)} \\
&= \frac{1}{n}(n_{j, d, D} - \alpha_j^{(d, D)}n_{j, d, D}) \quad\text{(by the definition of } \alpha_j^{(d, D)} \text{ in Definition 2.5)} \\
&= \frac{1}{n}(1-\alpha_j^{(d, D)})n_{j, d, D} \\
&< \frac{1}{n}\left(1-\frac{1}{1+w}\right)n_{j, d, D} \\
&= \frac{w}{n}\frac{1}{1+w}n_{j, d, D} \\
&< \frac{w}{n}\alpha_j^{(d, D)}n_{j, d, D} \\
&= \frac{w}{n}n^+_{j, d, D} \quad\text{(by the definition of } \alpha_j^{(d, D)} \text{ in Definition 2.5)} \\
&= R_j(d, D, \tau, w). \quad\text{(by the definition of } R_j \text{ in Equation (\ref{eq:jth_contrib}))}
\end{align*}

{\bf Case 2.} $\alpha_j^{(d, D)} > \tau$. In this case, both $R_j(d, D, 1/(1+w), w)$ and $R_j(d, D, \tau, w)$ are equal to $\frac{1}{n}n^-_{j, d, D}$.

{\bf Case 3.} $\alpha_j^{(d, D)} \leq 1/(1+w)$. In this case, both $R_j(d, D, 1/(1+w), w)$ and $R_j(d, D, \tau, w)$ are equal to $\frac{w}{n}n^+_{j, d, D}$.

Hence, given $\tau > 1/(1+w)$, we have
\begin{equation*}
R(d, D, 1/(1+w), w) = \sum_{j = 0}^{|d|} R_j(d, D, 1/(1+w), w) \leq \sum_{j = 0}^{|d|} R_j(d, D, \tau, w) = R(d, D, \tau, w).
\end{equation*}

The proof for $R(d, D, 1/(1+w), w) \leq R(d, D, \tau, w)$ given $\tau < 1/(1+w)$ is similar.
\end{proof}

{\bf Proposition 4.2.} Given the training data $D$, the set of antecedents $A$, and a prefix $e$ that is compatible with $D$ and satisfies $a_j^{(e)} \in A$ for all $j \in \{0, 1, ..., |e|-1\}$ and $\alpha_{k-1}^{(e, D)} \geq \alpha_k^{(e, D)}$ for all $k \in \{1, 2, ..., |e|-1\}$, the following statements are equivalent: (1) $e$ is feasible for Program 2.9 under $D$ and $A$; (2) $\tilde{\alpha}_{e, D} \leq \alpha_{|e|-1}^{(e, D)}$ holds; (3) $\tilde{n}^-_{e, D} \geq ((1/\alpha_{|e|-1}^{(e, D)})-1)\tilde{n}^+_{e, D}$ holds.

\begin{proof}
(1) $\Rightarrow$ (3): Suppose that Statement (1) holds. Then there exists a falling rule list
\begin{equation*}
d = \{e, (a_{|e|}^{(d)}, \alpha_{|e|}^{(d, D)}), ..., (a_{|d|-1}^{(d)}, \alpha_{|d|-1}^{(d, D)}), \alpha_{|d|}^{(d, D)}\}
\end{equation*}
that is compatible with $D$, and we have
\begin{align*}
\tilde{n}^-_{e, D} &= \tilde{n}_{e, D} - \tilde{n}^+_{e, D} \\
&= n_{|e|, d, D} + ... + n_{|d|, d, D} - \tilde{n}^+_{e, D} \\
&= \frac{1}{\alpha_{|e|}^{(d, D)}}n^+_{|e|, d, D} + ... + \frac{1}{\alpha_{|d|}^{(d, D)}}n^+_{|d|, d, D} - \tilde{n}^+_{e, D} \quad \text{ (by Definition 2.5)}\\
&\geq \frac{1}{\alpha_{|e|-1}^{(d, D)}}n^+_{|e|, d, D} + ... + \frac{1}{\alpha_{|e|-1}^{(d, D)}}n^+_{|d|, d, D} - \tilde{n}^+_{e, D} \quad \text{ (by the monotonicity constraint)}\\
&= \frac{1}{\alpha_{|e|-1}^{(d, D)}}(n^+_{|e|, d, D} + ... + n^+_{|d|, d, D}) - \tilde{n}^+_{e, D} \\
&= \frac{1}{\alpha_{|e|-1}^{(d, D)}}\tilde{n}^+_{e, D}  - \tilde{n}^+_{e, D} \\
&= ((1/\alpha_{|e|-1}^{(d, D)})-1)\tilde{n}^+_{e, D} \\
&= ((1/\alpha_{|e|-1}^{(e, D)})-1)\tilde{n}^+_{e, D}.
\end{align*}

(3) $\Rightarrow$ (2): Suppose that Statement (3) holds. Then we have
\begin{align*}
\tilde{\alpha}_{e, D} &= \frac{\tilde{n}^+_{e, D}}{\tilde{n}_{e, D}} \quad \text{ (by Definition 2.5)}\\
&= \frac{\tilde{n}^+_{e, D}}{\tilde{n}^+_{e, D} + \tilde{n}^-_{e, D}} \\
&\leq \frac{\tilde{n}^+_{e, D}}{\tilde{n}^+_{e, D} + ((1/\alpha_{|e|-1}^{(d, D)})-1)\tilde{n}^+_{e, D}} \quad \text{ (by Statement (3))} \\
&= \frac{\tilde{n}^+_{e, D}}{(1 + (1/\alpha_{|e|-1}^{(d, D)}) - 1)\tilde{n}^+_{e, D}} = \alpha_{|e|-1}^{(d, D)}.
\end{align*}

(2) $\Rightarrow$ (1): Suppose that Statement (2) holds. Then the falling rule list $d = \{e, \tilde{\alpha}_{e, D}\}$ begins with $e$ and is compatible with $D$. By Definition 4.1, $e$ is feasible for Program 2.9 under the training data $D$.
\end{proof}

Before we proceed with proving Lemma 4.4, we make the following observation.

{\bf Observation 10.1}
For any rule list
\begin{equation*}
d' = \{e, (a_{|e|}^{(d')}, \hat{\alpha}_{|e|}^{(d')}), ..., (a_{|d'|-1}^{(d')}, \hat{\alpha}_{|d'|-1}^{(d')}), \hat{\alpha}_{|d'|}^{(d')}\} 
\end{equation*}
that begins with a given prefix $e$, we have
\begin{equation}\label{eq:tilde_n+_obs_10.1}
\tilde{n}^+_{e, D} = n^+_{|e|, d', D} + ... n^+_{|d'|, d', D},
\end{equation}
\begin{equation}\label{eq:tilde_n-_obs_10.1}
\tilde{n}^-_{e, D} = n^-_{|e|, d', D} + ... n^-_{|d'|, d', D},
\end{equation}
and
\begin{equation}\label{eq:tilde_n_obs_10.1}
\tilde{n}_{e, D} = n_{|e|, d', D} + ... n_{|d'|, d', D}.
\end{equation}

\begin{proof}
Any positive training input $\mathbf{x}_i$ that is not captured by the prefix $e$ must be captured by some antecedent $a_j^{(d')}$ with $|e| \leq j < |d'|$ in $d'$, or the final else clause in $d'$. Conversely, any positive training input $\mathbf{x}_i$ that is captured by some antecedent $a_j^{(d')}$ with $|e| \leq j < |d'|$ in $d'$, or the final else clause in $d'$, must not satisfy any antecedent in the prefix $e$ and is consequently not captured by the prefix $e$. This means that the set of positive training inputs that are not captured by $e$ is exactly the set of positive training inputs that are captured by some antecedent $a_j^{(d')}$ with $|e| \leq j < |d'|$ in $d'$, or the final else clause in $d'$. It then follows that these two sets have the same number of elements. The former set has $\tilde{n}^+_{e, D}$ number of elements, and the latter has $n^+_{|e|, d', D} + ... + n^+_{|d'|, d', D}$ number of elements. This establishes Equation (\ref{eq:tilde_n+_obs_10.1}).

We can establish Equations (\ref{eq:tilde_n-_obs_10.1}) and (\ref{eq:tilde_n_obs_10.1}) using essentially the same argument.
\end{proof}

We now prove Lemma 4.4.

{\bf Lemma 4.4.} Suppose that we are given an instance $(D, A, w, C)$ of Program 2.9, a prefix $e$ that is feasible for Program 2.9 under the training data $D$ and the set of antecedents $A$, and a (possibly hypothetical) falling rule list $d$ that begins with $e$ and is compatible with $D$. Then there exists a falling rule list $d'$, possibly hypothetical with respect to $A$, such that $d'$ begins with $e$, has at most one more rule (excluding the final else clause) following $e$, is compatible with $D$, and satisfies
\begin{equation*}
L(d', D, 1/(1+w), w, C) \leq L(d, D, 1/(1+w), w, C).
\end{equation*}
Moreover, if either $\alpha_j^{(d, D)} > 1/(1+w)$ holds for all $j \in \{|e|, |e|+1, ..., |d|\}$, or $\alpha_j^{(d, D)} \leq 1/(1+w)$ holds for all $j \in \{|e|, |e|+1, ..., |d|\}$, then the falling rule list $\bar{e} = \{e, \tilde{\alpha}_{e, D}\}$ (i.e. the falling rule list in which the final else clause immediately follows the prefix $e$, and the probability estimate of the final else clause is $\tilde{\alpha}_{e, D}$) is compatible with $D$ and satisfies $L(\bar{e}, D, 1/(1+w), w, C) \leq L(d, D, 1/(1+w), w, C)$.

\begin{proof}
{\bf Case 1.} There exists some $k \in \{|e|+1, ..., |d|\}$ that satisfies $\alpha_{k-1}^{(d, D)} > 1/(1+w)$ but $\alpha_k^{(d, D)} \leq 1/(1+w)$. For any $j \in \{|e|, ..., k-1\}$, we have $\alpha_j^{(d, D)} > 1/(1+w)$, and the contribution $R_j(d, D, 1/(1+w), w)$ by the $j$-th rule to $R(d, D, 1/(1+w), w)$, defined by Equation (\ref{eq:jth_contrib}) with $\tau = 1/(1+w)$, is given by
\begin{equation}\label{eq:jth_contrib_gt_thresh}
R_j(d, D, 1/(1+w), w) = \frac{1}{n}n^-_{j, d, D}.
\end{equation}
For any $j \in \{k, ..., |d|\}$, we have $\alpha_j^{(d, D)} \leq 1/(1+w)$, and the contribution $R_j(d, D, 1/(1+w), w)$ by the $j$-th rule to $R(d, D, 1/(1+w), w)$ is given by
\begin{equation}\label{eq:jth_contrib_leq_thresh}
R_j(d, D, 1/(1+w), w) = \frac{w}{n}n^+_{j, d, D}.
\end{equation}

The rest of the proof for this case proceeds in three steps.

{\bf Step 1.} Construct a hypothetical falling rule list $d'$ that begins with $e$, has exactly one more rule (excluding the final else clause) following $e$, and is compatible with $D$. In later steps, we shall show that the falling rule list $d'$ constructed in this step satisfies $L(d', D, 1/(1+w), w, C) \leq  L(d, D, 1/(1+w), w, C)$.

Let $d' = \{e, (a_{|e|}^{(d')}, \hat{\alpha}_{|e|}^{(d')}), \hat{\alpha}_{|e|+1}^{(d')}\}$ be the falling rule list of size $|d'| = |e|+1$ that is compatible with $D$, such that
\begin{equation*}
a_{|e|}^{(d')} = a_{|e|}^{(d)} \vee ... \vee a_{k-1}^{(d)}
\end{equation*}
is the antecedent given by the logical {\bf or}'s of the antecedents $a_{|e|}^{(d)}$ through $a_{k-1}^{(d)}$ in $d$.

{\bf Step 2.} Show that the empirical risk of misclassification by the falling rule list $d'$ is the same as that by the falling rule list $d$.

To see this, we observe that the training instances captured by $a_{|e|}^{(d')}$ in $d'$ are exactly those captured by the antecedents $a_{|e|}^{(d)}$ through $a_{k-1}^{(d)}$ in $d$, and the training instances captured by $a_{|e|+1}^{(d')}$ (i.e. the final else clause) in $d'$ are exactly those captured by the antecedents $a_k^{(d)}$ through $a_{|d|}^{(d)}$ in $d$. This observation implies
\begin{equation}\label{eq:n+_e_lemma_4.4}
n^+_{|e|, d', D} = n^+_{|e|, d, D} + ... + n^+_{k-1, d, D},
\end{equation}
\begin{equation}\label{eq:n-_e_lemma_4.4}
n^-_{|e|, d', D} = n^-_{|e|, d, D} + ... + n^-_{k-1, d, D},
\end{equation}
\begin{equation}\label{eq:n_e_lemma_4.4}
n_{|e|, d', D} = n_{|e|, d, D} + ... + n_{k-1, d, D},
\end{equation}
\begin{equation}\label{eq:n+_e+1_lemma_4.4}
n^+_{|e|+1, d', D} = n^+_{k, d, D} + ... + n^+_{|d|, d, D},
\end{equation}
and
\begin{equation}\label{eq:n_e+1_lemma_4.4}
n_{|e|+1, d', D} = n_{k, d, D} + ... + n_{|d|, d, D}.
\end{equation}

Since $d'$ is compatible with $D$, using the definition of a compatible rule list in Definition 2.6 and the definition of the empirical positive proportion in Definition 2.5, together with (\ref{eq:n+_e_lemma_4.4}), (\ref{eq:n_e_lemma_4.4}), (\ref{eq:n+_e+1_lemma_4.4}), and (\ref{eq:n_e+1_lemma_4.4}), we must have
\begin{align*}
\hat{\alpha}_{|e|}^{(d')} = \alpha_{|e|}^{(d', D)} = \frac{n^+_{|e|, d', D}}{n_{|e|, d', D}} &= \frac{n^+_{|e|, d, D} + ... + n^+_{k-1, d, D}}{n_{|e|, d, D} + ... + n_{k-1, d, D}} \\&= \frac{\alpha_{|e|}^{(d, D)}n_{|e|, d, D} + ... + \alpha_{k-1}^{(d, D)}n_{k-1, d, D}}{n_{|e|, d, D} + ... + n_{k-1, d, D}} > \frac{1}{1+w},
\end{align*}
and
\begin{align*}
\hat{\alpha}_{|e|+1}^{(d')} = \alpha_{|e|+1}^{(d', D)} = \frac{n^+_{|e|+1, d', D}}{n_{|e|+1, d', D}} &= \frac{n^+_{k, d, D} + ... + n^+_{|d|, d, D}}{n_{k, d, D} + ... + n_{|d|, d, D}} \\&= \frac{\alpha_k^{(d, D)}n_{k, d, D} + ... + \alpha_{|d|}^{(d, D)}n_{|d|, d, D}}{n_{k, d, D} + ... + n_{|d|, d, D}} \leq \frac{1}{1+w}.
\end{align*}
This means that the contribution $R_{|e|}(d', D, 1/(1+w), w)$ by the $|e|$-th rule to $R(d', D, 1/(1+w), w)$ is given by
\begin{equation*}
R_{|e|}(d', D, 1/(1+w), w) = \frac{1}{n}n^-_{|e|, d', D} = \frac{1}{n}(n^-_{|e|, d, D} + ... + n^-_{k-1, d, D}),
\end{equation*}
where we have used (\ref{eq:n-_e_lemma_4.4}), and the contribution $R_{|e|+1}(d', D, 1/(1+w), w)$ by the $(|e|+1)$-st ``rule'' (i.e. the final else clause) to $R(d', D, 1/(1+w), w)$ is given by
\begin{equation*}
R_{|e|+1}(d', D, 1/(1+w), w) = \frac{w}{n}n^+_{|e|+1, d', D} = \frac{w}{n}(n^+_{k, d, D} + ... + n^+_{|d|, d, D}),
\end{equation*}
where we have used (\ref{eq:n+_e+1_lemma_4.4}).

It then follows that the empirical risk of misclassification by the rule list $d'$ is the same as that by the rule list $d$:
\begin{align}
&\quad R(d', D, 1/(1+w), w) \nonumber\\
&= R(e, D, 1/(1+w), w) + R_{|e|}(d', D, 1/(1+w), w) + R_{|e|+1}(d', D, 1/(1+w), w) \nonumber\\
&= R(e, D, 1/(1+w), w) + \frac{1}{n}(n^-_{|e|, d, D} + ... + n^-_{k-1, d, D}) + \frac{w}{n}(n^+_{k, d, D} + ... + n^+_{|d|, d, D}) \nonumber\\
&= R(e, D, 1/(1+w), w) + \sum_{j=|e|}^{|d|} R_j(d, D, 1/(1+w), w) \nonumber\\
&= R(d, D, 1/(1+w), w). \label{eq:same_emp_risk_lemma_4.4}
\end{align}

{\bf Step 3.} Put everything together.

Using (\ref{eq:same_emp_risk_lemma_4.4}), together with the observation $|d'| = |e| + 1 \leq |d|$, we must also have
\begin{align*}
L(d', D, 1/(1+w), w, C) &= R(d', D, 1/(1+w), w) + C|d'| \\
&\leq R(d, D, 1/(1+w), w) + C|d| = L(d, D, 1/(1+w), w, C),
\end{align*}
as desired.

{\bf Case 2.} $\alpha_j^{(d, D)} > 1/(1+w)$ holds for all $j \in \{|e|, |e|+1, ..., |d|\}$. Then the contribution $R_j(d, D, 1/(1+w), w)$ by the $j$-th rule to $R(d, D, 1/(1+w), w)$, for all $j \in \{|e|, |e|+1, ..., |d|\}$, is given by Equation (\ref{eq:jth_contrib_gt_thresh}). Let $d' = \{e, \hat{\alpha}_{|e|}^{(d')}\}$ be the falling rule list of size $|d'| = |e|$ that is compatible with $D$. Then the instances captured by $a_{|e|}^{(d')}$ (i.e. the final else clause) in $d'$ are exactly those that are not captured by $e$, or equivalently, those that are captured by $a_{|e|}^{(d)}$ through $a_{|d|}^{(d)}$. This implies
\begin{equation}\label{eq:n+_e_case2_lemma_4.4}
n^+_{|e|, d', D} = n^+_{|e|, d, D} + ... + n^+_{|d|, d, D},
\end{equation}
\begin{equation}\label{eq:n-_e_case2_lemma_4.4}
n^-_{|e|, d', D} = n^-_{|e|, d, D} + ... + n^-_{|d|, d, D},
\end{equation}
and
\begin{equation}\label{eq:n_e_case2_lemma_4.4}
n_{|e|, d', D} = n_{|e|, d, D} + ... + n_{|d|, d, D}.
\end{equation}

Since $d'$ is compatible with $D$, using the definition of a compatible rule list in Definition 2.6 and the definition of the empirical positive proportion in Definition 2.5, together with (\ref{eq:n+_e_case2_lemma_4.4}) and (\ref{eq:n_e_case2_lemma_4.4}), we must have
\begin{align}
\hat{\alpha}_{|e|}^{(d')} = \alpha_{|e|}^{(d', D)} &= \frac{n^+_{|e|, d', D}}{n_{|e|, d', D}} \nonumber \\&= \frac{n^+_{|e|, d, D} + ... + n^+_{|d|, d, D}}{n_{|e|, d, D} + ... + n_{|d|, d, D}} \label{eq:emp_pos_prop_after_e_gt_thresh}\\&= \frac{\alpha_{|e|}^{(d, D)}n_{|e|, d, D} + ... + \alpha_{|d|}^{(d, D)}n_{|d|, d, D}}{n_{|e|, d, D} + ... + n_{|d|, d, D}} > \frac{1}{1+w}. \label{eq:gt_threshold}
\end{align}
Note that the right-hand side of Equality (\ref{eq:emp_pos_prop_after_e_gt_thresh}) is equal to $\tilde{n}^+_{e, D}/\tilde{n}_{e, D} = \tilde{\alpha}_{e, D}$, by Equations (\ref{eq:tilde_n+_obs_10.1}) and (\ref{eq:tilde_n_obs_10.1}) in Observation 10.1. Therefore, we also have $\hat{\alpha}_{|e|}^{(d')} = \tilde{\alpha}_{e, D}$. 

Inequality (\ref{eq:gt_threshold}) implies that the contribution $R_{|e|}(d', D, 1/(1+w), w)$ by the $|e|$-th ``rule'' (i.e. the final else clause) to $R(d', D, 1/(1+w), w)$ is given by
\begin{equation*}
R_{|e|}(d', D, 1/(1+w), w) = \frac{1}{n}n^-_{|e|, d', D} = \frac{1}{n}(n^-_{|e|, d, D} + ... + n^-_{|d|, d, D}),
\end{equation*}
where we have used (\ref{eq:n-_e_case2_lemma_4.4}).

It then follows that the empirical risk of misclassification by the rule list $d'$ is the same as that by the rule list $d$:
\begin{align*}
&\quad R(d', D, 1/(1+w), w) \\
&= R(e, D, 1/(1+w), w) + R_{|e|}(d', D, 1/(1+w), w) \\
&= R(e, D, 1/(1+w), w) + \frac{1}{n}(n^-_{|e|, d, D} + ... + n^-_{|d|, d, D}) \\
&= R(e, D, 1/(1+w), w) + \sum_{j=|e|}^{|d|} R_j(d, D, 1/(1+w), w) \\
&= R(d, D, 1/(1+w), w).
\end{align*}

Since we clearly have $|d'| = |e| \leq |d|$, we must also have
\begin{align*}
L(d', D, 1/(1+w), w, C) &= R(d', D, 1/(1+w), w) + C|d'| \\
&\leq R(d, D, 1/(1+w), w) + C|d| = L(d, D, 1/(1+w), w, C),
\end{align*}
as desired.

{\bf Case 3.} $\alpha_j^{(d, D)} \leq 1/(1+w)$ holds for all $j \in \{|e|, |e|+1, ..., |d|\}$. The proof is similar to Case 2, with $R_j(d, D, 1/(1+w), w)$ for all $j \in \{|e|, |e|+1, ..., |d|\}$ given by Equation (\ref{eq:jth_contrib_leq_thresh}), the ``greater than'' in Inequality \ref{eq:gt_threshold} replaced by ``less than or equal to'', and $R_{|e|}(d', D, 1/(1+w), w)$ given by
\begin{equation*}
R_{|e|}(d', D, 1/(1+w), w) = \frac{w}{n}n^+_{|e|, d', D} = \frac{w}{n}(n^+_{|e|, d, D} + ... + n^+_{|d|, d, D}).
\end{equation*}
\end{proof}

{\bf Corollary 4.5.} If $d^*$ is an optimal solution for a given instance $(D, A, w, C)$ of Program 2.9, then we must have $\alpha_j^{(d^*, D)} > 1/(1+w)$ for all $j \in \{0, 1, ..., |d^*| - 1\}$.

\begin{proof}
Suppose that $d^*$ were an optimal solution for a given instance $(D, A, w, C)$ of Program 2.9, such that $\alpha_k^{(d^*, D)} \leq 1/(1+w)$ form some $k \in \{0, 1, ..., |d^*| - 1\}$. Let
\begin{equation*}
e = \{(a_0^{(d^*)}, \alpha_0^{(d^*, D)}), ..., (a_{k-1}^{(d^*)}, \alpha_{k-1}^{(d^*, D)})\}
\end{equation*}
be a prefix consisting of the top $k$ rules in $d^*$. By Lemma 4.4, the falling rule list $\bar{e} = \{e, \tilde{\alpha}_{e, D}\}$ satisfies $L(\bar{e}, D, 1/(1+w), w, C) \leq L(d^*, D, 1/(1+w), w, C)$. In fact, the inequality is strict because the size of $\bar{e}$ is strictly less than that of $d^*$. This contradicts the optimality of $d^*$.
\end{proof}

Before we proceed with proving Theorem 4.6, we make two other observations.

{\bf Observation 10.2.}
For any rule list $d'$, we have
\begin{equation}\label{eq:n-_e_obs_10.2}
n^-_{|e|, d', D} = \left(\frac{1}{\alpha_{|e|}^{(d', D)}} - 1\right)n^+_{|e|, d', D},
\end{equation}

\begin{proof}
By Definition 2.5, we have
\begin{equation*}
\alpha_{|e|}^{(d', D)} = n^+_{|e|, d', D}/n_{|e|, d', D}.
\end{equation*}
Since $n_{|e|, d', D}$ denotes the total number of training inputs captured by the $|e|$-th antecedent in $d'$, which is exactly the sum of the number of positive training inputs captured by that antecedent (denoted $n^+_{|e|, d', D}$), and the number of negative training inputs captured by the same antecedent (denoted $n^-_{|e|, d', D}$), we have
\begin{equation*}
\alpha_{|e|}^{(d', D)} = \frac{n^+_{|e|, d', D}}{n^+_{|e|, d', D} + n^-_{|e|, d', D}}.
\end{equation*}
The desired equation follows from rearranging the terms.
\end{proof}

{\bf Observation 10.3.}
For any rule list
\begin{equation*}
d' = \{e, (a_{|e|}^{(d')}, \hat{\alpha}_{|e|}^{(d')}), \hat{\alpha}_{|e|+1}^{(d')}\} 
\end{equation*}
that has exactly one rule (excluding the final else clause) following a given prefix $e$, we have
\begin{equation}\label{eq:n+_e+1_obs_10.3}
n^+_{|e|+1, d', D} = \tilde{n}^+_{e, D} - n^+_{|e|, d', D},
\end{equation}
\begin{equation}\label{eq:n-_e+1_obs_10.3}
n^-_{|e|+1, d', D} = \tilde{n}^-_{e, D} - n^-_{|e|, d', D},
\end{equation}
and
\begin{equation}\label{eq:n_e+1_obs_10.3}
n_{|e|+1, d', D} = \tilde{n}_{e, D} - n_{|e|, d', D}.
\end{equation}
Note that since $n^+_{|e|+1, d', D}$, $n^-_{|e|+1, d', D}$, and $n_{|e|+1, d', D}$ are non-negative, Equations (\ref{eq:n+_e+1}), (\ref{eq:n-_e+1}), and (\ref{eq:n_e+1}) imply $n^+_{|e|, d', D} \leq \tilde{n}^+_{e, D}$, $n^-_{|e|, d', D} \leq \tilde{n}^-_{e, D}$, and $n_{|e|, d', D} \leq \tilde{n}_{e, D}$.

\begin{proof}
Applying Observation 10.1 with $|d'| = |e| + 1$, we have
\begin{equation*}
\tilde{n}^+_{e, D} = n^+_{|e|, d', D} + n^+_{|e|+1, d', D},
\end{equation*}
\begin{equation*}
\tilde{n}^-_{e, D} = n^-_{|e|, d', D} + n^-_{|e|+1, d', D},
\end{equation*}
and
\begin{equation*}
\tilde{n}_{e, D} = n_{|e|, d', D} + n_{|e|+1, d', D}.
\end{equation*}
Equations (\ref{eq:n+_e+1_obs_10.3}), (\ref{eq:n-_e+1_obs_10.3}), and (\ref{eq:n_e+1_obs_10.3}) follow from rearranging the terms in the above equations.
\end{proof}

We now prove Theorem 4.6.

{\bf Theorem 4.6.} Suppose that we are given an instance $(D, A, w, C)$ of Program 2.9 and a prefix $e$ that is feasible for Program 2.9 under the training data $D$ and the set of antecedents $A$. Then any falling rule list $d$ that begins with $e$ and is compatible with $D$ satisfies
\begin{equation*}
L(d, D, 1/(1+w), w, C) \geq L^*(e, D, w, C),
\end{equation*}
where
\begin{equation*}
L^*(e, D, w, C) = L(e, D, 1/(1+w), w, C) + \min\left(\frac{1}{n}\left(\frac{1}{\alpha_{|e|-1}^{(e, D)}} - 1\right)\tilde{n}^+_{e, D} + C, \frac{w}{n}\tilde{n}^+_{e, D}, \frac{1}{n}\tilde{n}^-_{e, D}\right)
\end{equation*}
is a lower bound on the objective value of any compatible falling rule list that begins with $e$, which we call a prefix bound for $e$, under the instance $(D, A, w, C)$ of Program 2.9. Furthermore, if
\begin{equation}\label{eq:terminating_condition_supp}
C \geq \min\left(\frac{w}{n}\tilde{n}^+_{e, D}, \frac{1}{n}\tilde{n}^-_{e, D}\right) - \frac{1}{n}\left(\frac{1}{\alpha_{|e|-1}^{(e, D)}} - 1\right)\tilde{n}^+_{e, D}
\end{equation}
holds, then the falling rule list $\bar{e} = \{e, \tilde{\alpha}_{e, D}\}$ satisfies $L(\bar{e}, D, 1/(1+w), w, C) = L^*(e, D, w, C)$.

\begin{proof}
Let $\mathcal{F}(\mathcal{X}, D, e)$ be the set of (hypothetical and non-hypothetical) falling rule lists that begin with $e$ and are compatible with $D$, and let $\mathcal{F}(\mathcal{X}, D, e, k)$ be the subset of $\mathcal{F}(\mathcal{X}, D, e)$, consisting of those falling rule lists in $\mathcal{F}(\mathcal{X}, D, e)$ that have exactly $k$ rules (excluding the final else clause) following the prefix $e$.

Let $d \in \mathcal{F}(\mathcal{X}, D, e)$.

{\bf Case 1.} $\alpha_{|e|-1}^{(e, D)} > 1/(1+w)$.

In this case, Lemma 4.4 implies
\begin{equation}\label{eq:L_ineq}
L(d, D, 1/(1+w), w, C) \geq \inf_{d' \in \mathcal{F}(\mathcal{X}, D, e, 1) \cup \mathcal{F}(\mathcal{X}, D, e, 0)} L(d', D, 1/(1+w), w, C).
\end{equation}

Note that we have $\mathcal{F}(\mathcal{X}, D, e, 0) = \{\bar{e}\}$, where $\bar{e} = \{e, \tilde{\alpha}_{e, D}\}$ is the falling rule list in which the final else clause immediately follows the prefix $e$, and the probability estimate of the final else clause is $\tilde{\alpha}_{e, D}$. To see this, we first observe $\bar{e} \in \mathcal{F}(\mathcal{X}, D, e, 0)$. This is because:\\
(i) $\bar{e}$ clearly begins with $e$, and has no additional rules (excluding the final else clause) following the prefix $e$;\\
(ii) the feasibility of $e$ implies $\alpha_{k-1}^{(e, D)} \geq \alpha_k^{(e, D)}$ for all $k \in \{1, 2, ..., |e|-1\}$ (otherwise we could not possibly have a falling rule list that begins with $e$, and we would violate Definition 4.1), and $\tilde{\alpha}_{e, D} \leq \alpha_{|e|-1}^{(e, D)}$ (by Proposition 4.2), which together imply that $\bar{e}$ is indeed a falling rule list; and\\
(iii) we have 
\begin{align*}
\tilde{\alpha}_{e, D} &= \frac{\tilde{n}^+_{e, D}}{\tilde{n}_{e, D}} \quad\text{(by the definition of } \tilde{\alpha}_{e, D} \text{ in Definition 2.5)} \\
&= \frac{n^+_{|\bar{e}|, \bar{e}, D}}{n_{|\bar{e}|, \bar{e}, D}} \quad\text{(by Equations (\ref{eq:tilde_n+_obs_10.1}) and (\ref{eq:tilde_n_obs_10.1}) in Observation 10.1, applied to } \bar{e} \text{)} \\
&= \alpha_{|\bar{e}|}^{(\bar{e}, D)}, \quad\text{(by the definition of the empirical positive proportion in Definition 2.5)} 
\end{align*}
which implies that $\bar{e}$ is indeed compatible with $D$.

Conversely, for any $d_0 = \{e, \hat{\alpha}_{|e|}^{(d_0)}\} \in \mathcal{F}(\mathcal{X}, D, e, 0)$, we must have
\begin{align*}
\hat{\alpha}_{|e|}^{(d_0)} &= \alpha_{|e|}^{(d_0, D)} \quad\text{(because } d_0 \text{ must be compatible with } D \text{)} \\
&= \frac{n^+_{|e|, d_0, D}}{n_{|e|, d_0, D}} \quad\text{(by the definition of the empirical positive proportion in Definition 2.5)} \\
&= \frac{\tilde{n}^+_{e, D}}{\tilde{n}_{e, D}} \quad\text{(by Equations (\ref{eq:tilde_n+_obs_10.1}) and (\ref{eq:tilde_n_obs_10.1}) in Observation 10.1, applied to } d_0 \text{ here)} \\
&= \tilde{\alpha}_{e, D},
\end{align*}
which implies $d_0 = \bar{e}$. This establishes $\mathcal{F}(\mathcal{X}, D, e, 0) = \{\bar{e}\}$.

Let $\mathcal{F}'(\mathcal{X}, D, e, 1)$ be the subset of $\mathcal{F}(\mathcal{X}, D, e, 1)$, consisting of those falling rule lists
\begin{equation*}
d' = \{e, (a_{|e|}^{(d')}, \alpha_{|e|}^{(d', D)}), \alpha_{|e|+1}^{(d', D)}\} \in \mathcal{F}(\mathcal{X}, D, e, 1)
\end{equation*}
with $\alpha_{|e|}^{(d', D)} > 1/(1+w)$ and $\alpha_{|e|+1}^{(d', D)} \leq 1/(1+w)$. Note that for any $d_1 = \{e, (a_{|e|}^{(d_1)}, \alpha_{|e|}^{(d_1, D)}), \alpha_{|e|+1}^{(d_1, D)}\} \in \mathcal{F}(\mathcal{X}, D, e, 1) - \mathcal{F}'(\mathcal{X}, D, e, 1)$, we have either $\alpha_{|e|}^{(d_1, D)} \geq \alpha_{|e|+1}^{(d_1, D)} > 1/(1+w)$ or $\alpha_{|e|+1}^{(d_1, D)} \leq \alpha_{|e|}^{(d_1, D)} \leq 1/(1+w)$, and Lemma 4.4 implies $L(d_1, D, 1/(1+w), w, C) \geq L(\bar{e}, D, 1/(1+w), w, C)$. This means 
\begin{equation}\label{eq:inf_F-F'}
\inf_{d' \in \mathcal{F}(\mathcal{X}, D, e, 1) - \mathcal{F}'(\mathcal{X}, D, e, 1)} L(d', D, 1/(1+w), w, C) \geq L(\bar{e}, D, 1/(1+w), w, C).
\end{equation}

Using $\mathcal{F}(\mathcal{X}, D, e, 0) = \{\bar{e}\}$ and (\ref{eq:inf_F-F'}), we can write the right-hand side of (\ref{eq:L_ineq}) as
\begin{align}
&\quad \inf_{d' \in \mathcal{F}(\mathcal{X}, D, e, 1) \cup \mathcal{F}(\mathcal{X}, D, e, 0)} L(d', D, 1/(1+w), w, C) \nonumber\\
&= \inf_{d' \in \mathcal{F}'(\mathcal{X}, D, e, 1) \cup (\mathcal{F}(\mathcal{X}, D, e, 1) - \mathcal{F}'(\mathcal{X}, D, e, 1)) \cup \{\bar{e}\}} L(d', D, 1/(1+w), w, C) \nonumber\\
&= \min\left(\inf_{d' \in \mathcal{F}'(\mathcal{X}, D, e, 1)} L(d', D, 1/(1+w), w, C), \right. \nonumber\\
                   & \left. \vphantom{\inf_{d' \in \mathcal{F}'(\mathcal{X}, D, e, 1)} L(d', D, 1/(1+w), w, C)} \quad\quad\inf_{d' \in \mathcal{F}(\mathcal{X}, D, e, 1) - \mathcal{F}'(\mathcal{X}, D, e, 1)} L(d', D, 1/(1+w), w, C), L(\bar{e}, D, 1/(1+w), w, C)\right) \nonumber\\
&= \min\left(\inf_{d' \in \mathcal{F}'(\mathcal{X}, D, e, 1)} L(d', D, 1/(1+w), w, C), L(\bar{e}, D, 1/(1+w), w, C)\right). \label{eq:bound_L}
\end{align}

The rest of the proof for this case proceeds in three steps.

{\bf Step 1.} Compute $L(\bar{e}, D, 1/(1+w), w, C)$.

Since the contribution by the final else clause to $L(\bar{e}, D, 1/(1+w), w, C)$ is given by
\[
R_{|e|}(\bar{e}, D, 1/(1+w), w) = \begin{cases}
\frac{1}{n}n^-_{|e|, \bar{e}, D} &\text{ if } \tilde{\alpha}_{e, D} > 1/(1+w) \\
\frac{w}{n}n^+_{|e|, \bar{e}, D} &\text{ otherwise,}
\end{cases}
\]
where we have used Equation (\ref{eq:jth_contrib}), and since Observation 10.1 implies $\tilde{n}^+_{e, D} = n^+_{|e|, \bar{e}, D}$ and $\tilde{n}^-_{e, D} = n^-_{|e|, \bar{e}, D}$, it is not difficult to see
\begin{equation*}
L(\bar{e}, D, 1/(1+w), w, C) = \begin{cases}
L(e, D, 1/(1+w), w, C) + \frac{1}{n}\tilde{n}^-_{e, D} &\text{ if } \tilde{\alpha}_{e, D} > 1/(1+w) \\
L(e, D, 1/(1+w), w, C) + \frac{w}{n}\tilde{n}^+_{e, D} &\text{ otherwise.}
\end{cases}
\end{equation*}

Since $\tilde{\alpha}_{e, D} > 1/(1+w)$ is equivalent to $\tilde{n}^+_{e, D}/(\tilde{n}^+_{e, D} + \tilde{n}^-_{e, D}) > 1/(1+w)$, or $w\tilde{n}^+_{e, D} > \tilde{n}^-_{e, D}$, and similarly $\tilde{\alpha}_{e, D} \leq 1/(1+w)$ is equivalent to $w\tilde{n}^+_{e, D} \leq \tilde{n}^-_{e, D}$, we can write
\begin{equation}\label{eq:L(e_bar)}
\begin{split}
L(\bar{e}, D, 1/(1+w), w, C) = \min&\left(L(e, D, 1/(1+w), w, C) + \frac{1}{n}\tilde{n}^-_{e, D}, \right. \\
  &\quad \left. \vphantom{\frac{1}{n}} L(e, D, 1/(1+w), w, C) + \frac{w}{n}\tilde{n}^+_{e, D}\right).
\end{split}
\end{equation}

{\bf Step 2.} Determine a lower bound of $L(d', D, 1/(1+w), w, C)$ for all $d' \in \mathcal{F}'(\mathcal{X}, D, e, 1)$.

Let $d' = \{e, (a_{|e|}^{(d')}, \alpha_{|e|}^{(d', D)}), \alpha_{|e|+1}^{(d', D)}\} \in \mathcal{F}'(\mathcal{X}, D, e, 1)$. Since the contribution by both the $|e|$-th rule and the final else clause to $L(d', D, 1/(1+w), w, C)$ is given by $R_{|e|}(d', D, 1/(1+w), w) + R_{|e|+1}(d', D, 1/(1+w), w) + C$, where $R_{|e|}(d', D, 1/(1+w), w)$ and $R_{|e|+1}(d', D, 1/(1+w), w)$ are defined by Equation (\ref{eq:jth_contrib}) and are given by
\[
R_{|e|}(d', D, 1/(1+w), w) = \frac{1}{n}n^-_{|e|, d', D} \quad\text{ and }\quad R_{|e|+1}(d', D, 1/(1+w), w) = \frac{w}{n}n^+_{|e|+1, d', D}
\]
(because we have $\alpha_{|e|}^{(d', D)} > 1/(1+w)$ and $\alpha_{|e|+1}^{(d', D)} \leq 1/(1+w)$ for $d' \in \mathcal{F}'(\mathcal{X}, D, e, 1)$), it is not difficult to see
\begin{equation}\label{eq:L(d')}
L(d', D, 1/(1+w), w, C) = L(e, D, 1/(1+w), w, C) + \frac{1}{n}n^-_{|e|, d', D} + \frac{w}{n}n^+_{|e|+1, d', D} + C.
\end{equation}

Substituting (\ref{eq:n-_e_obs_10.2}) in Observation 10.2 and (\ref{eq:n+_e+1_obs_10.3}) in Observation 10.3 into Equation (\ref{eq:L(d')}), we have
\begin{align}
&\quad L(d', D, 1/(1+w), w, C) \nonumber\\&= L(e, D, 1/(1+w), w, C) + \frac{1}{n}\left(\frac{1}{\alpha_{|e|}^{(d', D)}} - 1\right)n^+_{|e|, d', D} + \frac{w}{n}(\tilde{n}^+_{e, D} - n^+_{|e|, d', D}) + C \nonumber \\
&= L(e, D, 1/(1+w), w, C) + \frac{1}{n}\left(\left(\frac{1}{\alpha_{|e|}^{(d', D)}} - 1 - w\right)n^+_{|e|, d', D} + w\tilde{n}^+_{e, D}\right) + C. \label{eq:L(d')_func}
\end{align}

Note that Equation (\ref{eq:L(d')_func}) shows that given the prefix $e$, $L(d', D, 1/(1+w), w, C)$ is a function of $\alpha_{|e|}^{(d', D)}$ and of $n^+_{|e|, d', D}$. Since we have
\begin{equation*}
\frac{\partial L(d', D, 1/(1+w), w, C)}{\partial n^+_{|e|, d', D}} = \frac{1}{n}\left(\frac{1}{\alpha_{|e|}^{(d', D)}} - 1 - w\right) < 0
\end{equation*}
because $\alpha_{|e|}^{(d', D)} > 1/(1+w)$ holds for any $d' \in \mathcal{F}'(\mathcal{X}, D, e, 1)$, and
\begin{equation*}
\frac{\partial L(d', D, 1/(1+w), w, C)}{\partial \alpha_{|e|}^{(d', D)}} = -\frac{n^+_{|e|, d', D}}{n}\frac{1}{(\alpha_{|e|}^{(d', D)})^2} \leq 0,
\end{equation*}
we see that $L(d', D, 1/(1+w), w, C)$ is indeed a monotonically decreasing function of both $n^+_{|e|, d', D}$ and $\alpha_{|e|}^{(d', D)}$. Thus, we can obtain a lower bound of $L(d', D, 1/(1+w), w, C)$ by substituting $n^+_{|e|, d', D}$ and $\alpha_{|e|}^{(d', D)}$ with their respective upper bound. The inequality $n^+_{|e|, d', D} \leq \tilde{n}^+_{e, D}$ in Observation 10.3 gives an upper bound for $n^+_{|e|, d', D}$, and the inequality $\alpha_{|e|}^{(d', D)} \leq \alpha_{|e|-1}^{(d', D)} = \alpha_{|e|-1}^{(e, D)}$ from $d'$ being a falling rule list gives an upper bound for $\alpha_{|e|}^{(d', D)}$. Substituting these upper bounds into (\ref{eq:L(d')_func}), we obtain the following inequality, which gives a lower bound of $L(d', D, 1/(1+w), w, C)$:
\begin{align*}
&\quad L(d', D, 1/(1+w), w, C) \\&\geq L(e, D, 1/(1+w), w, C) + \frac{1}{n}\left(\left(\frac{1}{\alpha_{|e|-1}^{(e, D)}} - 1 - w\right)\tilde{n}^+_{e, D} + w\tilde{n}^+_{e, D}\right) + C \\
&= L(e, D, 1/(1+w), w, C) + \frac{1}{n}\left(\left(\frac{1}{\alpha_{|e|-1}^{(e, D)}} - 1\right)\tilde{n}^+_{e, D}\right) + C.
\end{align*}
This means
\begin{equation}\label{eq:inf_F'}
\inf_{d' \in \mathcal{F}'(\mathcal{X}, D, e, 1)} L(d', D, 1/(1+w), w, C) \geq L(e, D, 1/(1+w), w, C) + \frac{1}{n}\left(\left(\frac{1}{\alpha_{|e|-1}^{(e, D)}} - 1\right)\tilde{n}^+_{e, D}\right) + C.
\end{equation}

{\bf Step 3.} Put everything together.

Using (\ref{eq:L_ineq}), (\ref{eq:bound_L}), (\ref{eq:L(e_bar)}), and (\ref{eq:inf_F'}), we have
\begin{align*}
&\quad L(d, D, 1/(1+w), w, C) \\&\geq \min\left(\inf_{d' \in \mathcal{F}'(\mathcal{X}, D, e, 1)} L(d', D, 1/(1+w), w, C), L(\bar{e}, D, 1/(1+w), w, C)\right) \\
&\geq \min\left(L(e, D, 1/(1+w), w, C) + \frac{1}{n}\left(\left(\frac{1}{\alpha_{|e|-1}^{(e, D)}} - 1\right)\tilde{n}^+_{e, D}\right) + C, \right. \\
  &\quad\quad \left. \vphantom{\frac{1}{\alpha_{|e|-1}^{(d', D)}}} \min\left(L(e, D, 1/(1+w), w, C) + \frac{1}{n}\tilde{n}^-_{e, D}, L(e, D, 1/(1+w), w, C) + \frac{w}{n}\tilde{n}^+_{e, D}\right)\right) \\
&= L(e, D, 1/(1+w), w, C) + \min\left(\frac{1}{n}\left(\left(\frac{1}{\alpha_{|e|-1}^{(e, D)}} - 1\right)\tilde{n}^+_{e, D}\right) + C, \frac{w}{n}\tilde{n}^+_{e, D}, \frac{1}{n}\tilde{n}^-_{e, D}\right),
\end{align*}
as desired.

{\bf Case 2.} $\alpha_{|e|-1}^{(e, D)} \leq 1/(1+w)$.

This implies $\alpha_j^{(d, D)} \leq 1/(1+w)$ for all $j \in \{|e|, ..., |d|\}$. By Lemma 4.4, we have
\begin{equation*}
L(d, D, 1/(1+w), w, C) \geq L(\bar{e}, D, 1/(1+w), w, C).
\end{equation*}
Since $L(\bar{e}, D, 1/(1+w), w, C)$ is given by Equation (\ref{eq:L(e_bar)}), we have
\begin{equation}\label{eq:bound_L_case2}
L(d, D, 1/(1+w), w, C) \geq L(e, D, 1/(1+w), w, C) + \min\left(\frac{w}{n}\tilde{n}^+_{e, D}, \frac{1}{n}\tilde{n}^-_{e, D}\right).
\end{equation}
Given $\alpha_{|e|-1}^{(e, D)} \leq 1/(1+w)$, we must also have
\begin{equation*}
\frac{1}{n}\left(\left(\frac{1}{\alpha_{|e|-1}^{(d', D)}} - 1\right)\tilde{n}^+_{e, D}\right) + C \geq \frac{w}{n}\tilde{n}^+_{e, D} + C \geq \frac{w}{n} \tilde{n}^+_{e, D},
\end{equation*}
which means
\begin{equation}\label{eq:min_case2}
\min\left(\frac{w}{n}\tilde{n}^+_{e, D}, \frac{1}{n}\tilde{n}^-_{e, D}\right) =\min\left(\frac{1}{n}\left(\left(\frac{1}{\alpha_{|e|-1}^{(d', D)}} - 1\right)\tilde{n}^+_{e, D}\right) + C, \frac{w}{n}\tilde{n}^+_{e, D}, \frac{1}{n}\tilde{n}^-_{e, D}\right).
\end{equation}
Substituting (\ref{eq:min_case2}) into (\ref{eq:bound_L_case2}) completes the proof for Case 2.

Finally, if Inequality (\ref{eq:terminating_condition_supp}) holds, then we have
\begin{equation*}
\frac{1}{n}\left(\left(\frac{1}{\alpha_{|e|-1}^{(d', D)}} - 1\right)\tilde{n}^+_{e, D}\right) + C \geq \min\left(\frac{w}{n}\tilde{n}^+_{e, D}, \frac{1}{n}\tilde{n}^-_{e, D}\right),
\end{equation*}
which implies
\begin{equation*}
L^*(e, D, w, C) = L(e, D, 1/(1+w), w, C) + \min\left(\frac{w}{n}\tilde{n}^+_{e, D}, \frac{1}{n}\tilde{n}^-_{e, D}\right) = L(\bar{e}, D, 1/(1+w), w, C).
\end{equation*}
\end{proof}

\section{Proof of Theorem 5.2}


{\bf Theorem 5.2.}
Suppose that we are given an instance $(D, A, w, C, C_1)$ of Program 5.1 and a prefix $e$ that is compatible with $D$. Then any rule list $d$ that begins with $e$ and is compatible with $D$ satisfies
\begin{equation*}
\tilde{L}(d, D, 1/(1+w), w, C, C_1) \geq \tilde{L}^*(e, D, w, C, C_1),
\end{equation*}
where
\begin{equation}\label{eq:prefix_bound_soft_frl_supp}
\begin{split}
\tilde{L}^*(e, D, &w, C, C_1) = \tilde{L}(e, D, 1/(1+w), w, C, C_1) \\
+ \min&\left(\frac{1}{n}\left(\frac{1}{\alpha_{\min}^{(e, D)}} - 1\right)\tilde{n}^+_{e, D} + C + C_1\lfloor\tilde{\alpha}_{e, D} - \alpha_{\min}^{(e, D)}\rfloor_+ + \frac{w}{n}\tilde{n}^+_{e, D}\mathds{1}[\tilde{\alpha}_{e, D} \geq \alpha_{\min}^{(e, D)}], \right. \\
                   & \left. \vphantom{\frac{1}{\alpha_{|e|-1}^{(e, D)}} - 1} \inf_{\beta: \zeta < \beta \leq 1} g(\beta), \frac{w}{n}\tilde{n}^+_{e, D} + C_1\lfloor\tilde{\alpha}_{e, D} - \alpha_{\min}^{(e, D)}\rfloor_+, \frac{1}{n}\tilde{n}^-_{e, D} + C_1\lfloor\tilde{\alpha}_{e, D} - \alpha_{\min}^{(e, D)}\rfloor_+\right)
\end{split}
\end{equation}
is a lower bound on the objective value of any compatible rule list that begins with $e$, under the instance $(D, A, w, C, C_1)$ of Program 5.1. In Equation (\ref{eq:prefix_bound_soft_frl_supp}), $\alpha_{\min}^{(e, D)}$, $\zeta$, and $g$ are defined by
\begin{equation*}
\alpha_{\min}^{(e, D)} = \min_{k < |e|} \alpha_k^{(e, D)}, \quad \zeta = \max(\alpha_{\min}^{(e, D)}, \tilde{\alpha}_{e, D}, 1/(1+w)),
\end{equation*}
\begin{equation*}
g(\beta) = \frac{1}{n}\left(\frac{1}{\beta} - 1\right)\tilde{n}^+_{e, D} + C + C_1(\beta - \alpha_{\min}^{(e, D)}).
\end{equation*}
Note that $\inf_{\beta: \zeta < \beta \leq 1} g(\beta)$ can be computed analytically: $\inf_{\beta: \zeta < \beta \leq 1} g(\beta) = g(\beta^*)$ if $\beta^* = \sqrt{\tilde{n}^+_{e, D}/(C_1 n)}$ satisfies $\zeta < \beta^* \leq 1$, and $\inf_{\beta: \zeta < \beta \leq 1} g(\beta) = \min(g(\zeta), g(1))$ otherwise.

To prove Theorem 5.2, we need the following lemma:

{\bf Lemma.} Suppose that we are given an instance $(D, A, w, C, C_1)$ of Program 5.1, a prefix $e$ that is compatible with $D$, and a (possibly hypothetical) rule list $d$ that begins with $e$ and is compatible with $D$. Then there exists a rule list $d'$, possibly hypothetical with respect to $A$, such that $d'$ begins with $e$, has at most one more rule (excluding the final else clause) following $e$, is compatible with $D$, and satisfies
\begin{equation}\label{eq:lower_bound_from_merging_rules_lemma_for_5.2}
\tilde{L}(d', D, 1/(1+w), w, C, C_1) \leq \tilde{L}(d, D, 1/(1+w), w, C, C_1).
\end{equation}
Moreover, if either $\alpha_j^{(d, D)} > 1/(1+w)$ holds for all $j \in \{|e|, |e|+1, ..., |d|\}$, or $\alpha_j^{(d, D)} \leq 1/(1+w)$ holds for all $j \in \{|e|, |e|+1, ..., |d|\}$, then the rule list $\bar{e} = \{e, \tilde{\alpha}_{e, D}\}$ (i.e. the rule list in which the final else clause follows immediately the prefix $e$, and the probability estimate of the final else clause is $\tilde{\alpha}_{e, D}$) is compatible with $D$ and satisfies $\tilde{L}(\bar{e}, D, 1/(1+w), w, C, C_1) \leq \tilde{L}(d, D, 1/(1+w), w, C)$.

\begin{proof}
{\bf Case 1.} There exists some $k \in \{|e|, ..., |d|\}$ that satisfies $\alpha_k^{(d, D)} > 1/(1+w)$ and some $k' \in \{|e|, ..., |d|\}$ that satisfies $\alpha_{k'}^{(d, D)} \leq 1/(1+w)$. For any $j \in \{|e|, ..., |d|\}$ with $\alpha_j^{(d, D)} > 1/(1+w)$, the contribution $R_j(d, D, 1/(1+w), w)$ by the $j$-th rule to $R(d, D, 1/(1+w), w)$, defined by the right-hand side of Equation (\ref{eq:jth_contrib}) with $\tau = 1/(1+w)$, is given by
\begin{equation*}
R_j(d, D, 1/(1+w), w) = \frac{1}{n}n^-_{j, d, D}.
\end{equation*}
For any $j \in \{|e|, ..., |d|\}$ with $\alpha_j^{(d, D)} \leq 1/(1+w)$, the contribution $R_j(d, D, 1/(1+w), w)$ by the $j$-th rule to $R(d, D, 1/(1+w), w)$ is given by
\begin{equation*}
R_j(d, D, 1/(1+w), w) = \frac{w}{n}n^+_{j, d, D}.
\end{equation*}

The rest of the proof for this case proceeds in four steps.

{\bf Step 1.} Construct a hypothetical rule list $d'$ that begins with $e$, has exactly one more rule (excluding the final else clause) following $e$, and is compatible with $D$. In later steps, we shall show that the rule list $d'$ constructed in this step satisfies (\ref{eq:lower_bound_from_merging_rules_lemma_for_5.2}).

Let $d' = \{e, (a_{|e|}^{(d')}, \hat{\alpha}_{|e|}^{(d')}), \hat{\alpha}_{|e|+1}^{(d')}\}$ be the hypothetical rule list of size $|d'| = |e|+1$ that is compatible with $D$, and whose $|e|$-th antecedent $a_{|e|}^{(d')}$ is defined by
\begin{equation*}
a_{|e|}^{(d')}(\mathbf{x}) = \mathds{1}[\alpha_{\text{capt}(\mathbf{x}, d)}^{(d, D)} > 1/(1+w)] \cdot \mathds{1}[|e| \leq \text{capt}(\mathbf{x}, d) \leq |d|].
\end{equation*}

{\bf Step 2.} Show that the empirical risk of misclassification by the rule list $d'$ is the same as that by the rule list $d$.

To see this, we observe that the training instances in $D$ captured by $a_{|e|}^{(d')}$ in $d'$ are exactly those captured by the antecedents $a_j^{(d)}$, $|e| \leq j \leq |d|$, in $d$ whose empirical positive proportion satisfies $\alpha_j^{(d, D)} > 1/(1+w)$, and the training instances in $D$ captured by $a_{|e|+1}^{(d')}$ (i.e. the final else clause) in $d'$ are exactly those captured by the antecedents $a_j^{(d)}$, $|e| \leq j \leq |d|$, in $d$ whose empirical positive proportion satisfies $\alpha_j^{(d, D)} \leq 1/(1+w)$. This observation implies
\begin{equation}\label{eq:n+_e_lemma_for_5.2}
n^+_{|e|, d', D} = \sum_{j: |e| \leq j \leq |d| \wedge \alpha_j^{(d, D)} > 1/(1+w)} n^+_{j, d, D},
\end{equation}
\begin{equation}\label{eq:n-_e_lemma_for_5.2}
n^-_{|e|, d', D} = \sum_{j: |e| \leq j \leq |d| \wedge \alpha_j^{(d, D)} > 1/(1+w)} n^-_{j, d, D},
\end{equation}
\begin{equation}\label{eq:n_e_lemma_for_5.2}
n_{|e|, d', D} = \sum_{j: |e| \leq j \leq |d| \wedge \alpha_j^{(d, D)} > 1/(1+w)} n_{j, d, D},
\end{equation}
\begin{equation}\label{eq:n+_e+1_lemma_for_5.2}
n^+_{|e|+1, d', D} = \sum_{j: |e| \leq j \leq |d| \wedge \alpha_j^{(d, D)} \leq 1/(1+w)} n^+_{j, d, D}
\end{equation}
and
\begin{equation}\label{eq:n_e+1_lemma_for_5.2}
n_{|e|+1, d', D} = \sum_{j: |e| \leq j \leq |d| \wedge \alpha_j^{(d, D)} \leq 1/(1+w)} n_{j, d, D}.
\end{equation}

Since $d'$ is compatible with $D$, using the definition of a compatible rule list in Definition 2.6 and the definition of the empirical positive proportion in Definition 2.5, together with (\ref{eq:n+_e_lemma_for_5.2}), (\ref{eq:n_e_lemma_for_5.2}), (\ref{eq:n+_e+1_lemma_for_5.2}), and (\ref{eq:n_e+1_lemma_for_5.2}), we must have
\begin{align*}
\hat{\alpha}_{|e|}^{(d')} = \alpha_{|e|}^{(d', D)} = \frac{n^+_{|e|, d', D}}{n_{|e|, d', D}} &= \frac{\sum_{j: |e| \leq j \leq |d| \wedge \alpha_j^{(d, D)} > 1/(1+w)} n^+_{j, d, D}}{\sum_{j: |e| \leq j \leq |d| \wedge \alpha_j^{(d, D)} > 1/(1+w)} n_{j, d, D}} \\&= \frac{\sum_{j: |e| \leq j \leq |d| \wedge \alpha_j^{(d, D)} > 1/(1+w)} \alpha_j^{(d, D)}n_{j, d, D}}{\sum_{j: |e| \leq j \leq |d| \wedge \alpha_j^{(d, D)} > 1/(1+w)} n_{j, d, D}} > \frac{1}{1+w},
\end{align*}
and
\begin{align*}
\hat{\alpha}_{|e|+1}^{(d')} = \alpha_{|e|+1}^{(d', D)} = \frac{n^+_{|e|+1, d', D}}{n_{|e|+1, d', D}} &= \frac{\sum_{j: |e| \leq j \leq |d| \wedge \alpha_j^{(d, D)} \leq 1/(1+w)} n^+_{j, d, D}}{\sum_{j: |e| \leq j \leq |d| \wedge \alpha_j^{(d, D)} \leq 1/(1+w)} n_{j, d, D}} \\&= \frac{\sum_{j: |e| \leq j \leq |d| \wedge \alpha_j^{(d, D)} \leq 1/(1+w)} \alpha_j^{(d, D)}n_{j, d, D}}{\sum_{j: |e| \leq j \leq |d| \wedge \alpha_j^{(d, D)} \leq 1/(1+w)} n_{j, d, D}} \leq \frac{1}{1+w}.
\end{align*}
This means that the contribution $R_{|e|}(d', D, 1/(1+w), w)$ by the $|e|$-th rule to $R(d', D, 1/(1+w), w)$ is given by
\begin{equation*}
R_{|e|}(d', D, 1/(1+w), w) = \frac{1}{n}n^-_{|e|, d', D} = \frac{1}{n}\sum_{j: |e| \leq j \leq |d| \wedge \alpha_j^{(d, D)} > 1/(1+w)} n^-_{j, d, D},
\end{equation*}
where we have used (\ref{eq:n-_e_lemma_for_5.2}), and the contribution $R_{|e|+1}(d', D, 1/(1+w), w)$ by the $(|e|+1)$-st ``rule'' (i.e. the final else clause) to $R(d', D, 1/(1+w), w)$ is given by
\begin{equation*}
R_{|e|+1}(d', D, 1/(1+w), w) = \frac{w}{n}n^+_{|e|+1, d', D} = \frac{w}{n}\sum_{j: |e| \leq j \leq |d| \wedge \alpha_j^{(d, D)} \leq 1/(1+w)} n^+_{j, d, D},
\end{equation*}
where we have used (\ref{eq:n+_e+1_lemma_for_5.2}).

It then follows that the empirical risk of misclassification by the rule list $d'$ is the same as that by the rule list $d$:
\begin{align}
&\quad R(d', D, 1/(1+w), w) \nonumber\\
&= R(e, D, 1/(1+w), w) + R_{|e|}(d', D, 1/(1+w), w) + R_{|e|+1}(d', D, 1/(1+w), w) \nonumber\\
&= R(e, D, 1/(1+w), w) \nonumber\\&\quad + \frac{1}{n}\sum_{j: |e| \leq j \leq |d| \wedge \alpha_j^{(d, D)} > 1/(1+w)} n^-_{j, d, D} + \frac{w}{n}\sum_{j: |e| \leq j \leq |d| \wedge \alpha_j^{(d, D)} \leq 1/(1+w)} n^+_{j, d, D} \nonumber\\
&= R(e, D, 1/(1+w), w) + \sum_{j=|e|}^{|d|} R_j(d, D, 1/(1+w), w) \nonumber\\
&= R(d, D, 1/(1+w), w). \label{eq:same_emp_risk_lemma_for_5.2}
\end{align}

{\bf Step 3.} Show that the monotonicity penalty of the rule list $d'$ is at most that of $d$.

Let $S(d, D) = \sum_{j=0}^{|d|}\lfloor\alpha_j^{(d, D)} - \min_{k<j} \alpha_k^{(d, D)}\rfloor_+$ be the monotonicity penalty of the rule list $d$. We now show $S(d', D) \leq S(d, D)$. Let $S_j(d, D) = \lfloor\alpha_j^{(d, D)} - \min_{k<j} \alpha_k^{(d, D)}\rfloor_+$ be the monotonicity penalty for the $j$-th rule in $d$.

Let $l \in \{|e|, ..., |d|\}$ be any integer with
\begin{equation}\label{eq:def_l_lemma_for_5.2}
\alpha_l^{(d, D)} = \max_{j: |e| \leq j \leq |d| \wedge \alpha_j^{(d, D)} > 1/(1+w)} \alpha_j^{(d, D)}.
\end{equation}
Then the total monotonicity penalty for all the rules $(a_j^{(d)}, \alpha_j^{(d, D)})$ in $d$ with $|e| \leq j \leq |d|$ and $\alpha_j^{(d, D)} > 1/(1+w)$ satisfies
\begin{align}
\sum_{j: |e| \leq j \leq |d| \wedge \alpha_j^{(d, D)} > 1/(1+w)} S_j(d, D) &\geq S_l(d, D) \quad\text{(because } S_l(d, D) \text{ is included in the sum on the left)} \nonumber\\
&= \lfloor\alpha_l^{(d, D)} - \min_{k<l} \alpha_k^{(d, D)}\rfloor_+ \nonumber\\
&\geq \lfloor\alpha_l^{(d, D)} - \min_{k<|e|} \alpha_k^{(d, D)}\rfloor_+. \label{eq:monotonicity_penalty_d_lemma_for_5.2}
\end{align}
On the other hand, the monotonicity penalty for the $|e|$-th rule in $d'$ satisfies
\begin{equation}\label{eq:monotonicity_penalty_d'_lemma_for_5.2}
S_{|e|}(d', D) = \lfloor\alpha_{|e|}^{(d', D)} - \min_{k<|e|} \alpha_k^{(d', D)}\rfloor_+ \leq \lfloor\alpha_l^{(d, D)} - \min_{k<|e|} \alpha_k^{(d, D)}\rfloor_+,
\end{equation}
because we have $\min_{k<|e|} \alpha_k^{(d', D)} = \min_{k<|e|} \alpha_k^{(d, D)}$ ($d$ and $d'$ begin with the same prefix $e$), and
\begin{align*}
\alpha_{|e|}^{(d', D)} &= \frac{n^+_{|e|, d', D}}{n_{|e|, d', D}} \quad\text{(by the definition of the empirical positive proportion in Definition 2.5)}\\
&= \frac{\sum_{j: |e| \leq j \leq |d| \wedge \alpha_j^{(d, D)} > 1/(1+w)} n^+_{j, d, D}}{\sum_{j: |e| \leq j \leq |d| \wedge \alpha_j^{(d, D)} > 1/(1+w)} n_{j, d, D}} \quad\text{(by Equations (\ref{eq:n+_e_lemma_for_5.2}) and (\ref{eq:n_e_lemma_for_5.2}))}\\
&= \frac{\sum_{j: |e| \leq j \leq |d| \wedge \alpha_j^{(d, D)} > 1/(1+w)} \alpha_j^{(d, D)}n_{j, d, D}}{\sum_{j: |e| \leq j \leq |d| \wedge \alpha_j^{(d, D)} > 1/(1+w)} n_{j, d, D}} \quad\text{(by the definition of } \alpha_j^{(d, D)} \text{ in Definition 2.5)}\\
&\leq \frac{\sum_{j: |e| \leq j \leq |d| \wedge \alpha_j^{(d, D)} > 1/(1+w)} \alpha_l^{(d, D)}n_{j, d, D}}{\sum_{j: |e| \leq j \leq |d| \wedge \alpha_j^{(d, D)} > 1/(1+w)} n_{j, d, D}} \quad\text{(by the definition of } l \text{ in (\ref{eq:def_l_lemma_for_5.2}))}\\
&= \alpha_l^{(d, D)}.
\end{align*}

Combining (\ref{eq:monotonicity_penalty_d_lemma_for_5.2}) and (\ref{eq:monotonicity_penalty_d'_lemma_for_5.2}), we have
\begin{equation}\label{eq:monotonicity_penalty_d'_|e|_lemma_for_5.2}
S_{|e|}(d', D) \leq \sum_{j: |e| \leq j \leq |d| \wedge \alpha_j^{(d, D)} > 1/(1+w)} S_j(d, D).
\end{equation}
A similar argument will show
\begin{equation}\label{eq:monotonicity_penalty_d'_|e|+1_lemma_for_5.2}
S_{|e|+1}(d', D) \leq \sum_{j: |e| \leq j \leq |d| \wedge \alpha_j^{(d, D)} \leq 1/(1+w)} S_j(d, D).
\end{equation}

It then follows from (\ref{eq:monotonicity_penalty_d'_|e|_lemma_for_5.2}) and (\ref{eq:monotonicity_penalty_d'_|e|+1_lemma_for_5.2}) that the monotonicity penalty of $d'$ is at most that of $d$:
\begin{align}
S(d', D) &= \left(\sum_{j=0}^{|e|-1} S_j(d', D)\right) + S_{|e|}(d', D) + S_{|e|+1}(d', D) \nonumber\\
&\leq \left(\sum_{j=0}^{|e|-1} S_j(d, D)\right) + \sum_{j: |e| \leq j \leq |d| \wedge \alpha_j^{(d, D)} > 1/(1+w)} S_j(d, D) \\&\quad + \sum_{j: |e| \leq j \leq |d| \wedge \alpha_j^{(d, D)} \leq 1/(1+w)} S_j(d, D) \nonumber\\
&= S(d, D). \label{eq:bounded_monotonicity_penalty_lemma_for_5.2}
\end{align}

{\bf Step 4.} Put everything together.

Using (\ref{eq:same_emp_risk_lemma_for_5.2}) and (\ref{eq:bounded_monotonicity_penalty_lemma_for_5.2}), together with the observation $|d'| = |e| + 1 \leq |d|$, we must also have
\begin{align*}
\tilde{L}(d', D, 1/(1+w), w, C, C_1) &= R(d', D, 1/(1+w), w) + C|d'| + C_1 S(d', D) \\
&\leq R(d, D, 1/(1+w), w) + C|d| + C_1 S(d, D) \\
&= \tilde{L}(d, D, 1/(1+w), w, C, C_1).
\end{align*}

{\bf Case 2.} Either $\alpha_j^{(d, D)} > 1/(1+w)$ holds for all $j \in \{|e|, ..., |d|\}$, or $\alpha_j^{(d, D)} \leq 1/(1+w)$ holds for all $j \in \{|e|, ..., |d|\}$. The construction of $d' = \bar{e}$ and the proof for $R(d', D, 1/(1+w), w) = R(d, D, 1/(1+w), w)$ is similar to those given in the proof of Lemma 4.4. The proof for $S(d', D) \leq S(d, D)$ is similar to that in Case 1. The desired inequality then follows from $|d'| = |e| \leq |d|$.
\end{proof}

Before we proceed with proving Theorem 5.2, we make the following four observations. Observations 11.1, 11.2, and 11.3 are the same as Observations 10.1, 10.2 and 10.3. They are repeated here for convenience.

{\bf Observation 11.1}
For any rule list
\begin{equation*}
d' = \{e, (a_{|e|}^{(d')}, \hat{\alpha}_{|e|}^{(d')}), ..., (a_{|d'|-1}^{(d')}, \hat{\alpha}_{|d'|-1}^{(d')}), \hat{\alpha}_{|d'|}^{(d')}\} 
\end{equation*}
that begins with a given prefix $e$, we have
\begin{equation}\label{eq:tilde_n+_obs_11.1}
\tilde{n}^+_{e, D} = n^+_{|e|, d', D} + ... n^+_{|d'|, d', D},
\end{equation}
\begin{equation}\label{eq:tilde_n-_obs_11.1}
\tilde{n}^-_{e, D} = n^-_{|e|, d', D} + ... n^-_{|d'|, d', D},
\end{equation}
and
\begin{equation}\label{eq:tilde_n_obs_11.1}
\tilde{n}_{e, D} = n_{|e|, d', D} + ... n_{|d'|, d', D}.
\end{equation}

\begin{proof}
Same as Observation 10.1.
\end{proof}

{\bf Observation 11.2.}
For any rule list $d'$, we have
\begin{equation}\label{eq:n-_e}
n^-_{|e|, d', D} = \left(\frac{1}{\alpha_{|e|}^{(d', D)}} - 1\right)n^+_{|e|, d', D},
\end{equation}

\begin{proof}
Same as Observation 10.2.
\end{proof}

{\bf Observation 11.3.}
For any rule list
\begin{equation*}
d' = \{e, (a_{|e|}^{(d')}, \hat{\alpha}_{|e|}^{(d')}), \hat{\alpha}_{|e|+1}^{(d')}\} 
\end{equation*}
that has exactly one rule (excluding the final else clause) following a given prefix $e$, we have
\begin{equation}\label{eq:n+_e+1}
n^+_{|e|+1, d', D} = \tilde{n}^+_{e, D} - n^+_{|e|, d', D},
\end{equation}
\begin{equation}\label{eq:n-_e+1}
n^-_{|e|+1, d', D} = \tilde{n}^-_{e, D} - n^-_{|e|, d', D},
\end{equation}
and
\begin{equation}\label{eq:n_e+1}
n_{|e|+1, d', D} = \tilde{n}_{e, D} - n_{|e|, d', D}.
\end{equation}
Note that since $n^+_{|e|+1, d', D}$, $n^-_{|e|+1, d', D}$, and $n_{|e|+1, d', D}$ are non-negative, Equations (\ref{eq:n+_e+1}), (\ref{eq:n-_e+1}), and (\ref{eq:n_e+1}) imply $n^+_{|e|, d', D} \leq \tilde{n}^+_{e, D}$, $n^-_{|e|, d', D} \leq \tilde{n}^-_{e, D}$, and $n_{|e|, d', D} \leq \tilde{n}_{e, D}$.

\begin{proof}
Same as Observation 10.3.
\end{proof}

{\bf Observation 11.4.}
For any rule list
\begin{equation*}
d' = \{e, (a_{|e|}^{(d')}, \hat{\alpha}_{|e|}^{(d')}), \hat{\alpha}_{|e|+1}^{(d')}\} 
\end{equation*}
that has exactly one rule (excluding the final else clause) following a given prefix $e$, we have
\begin{equation}
\alpha_{|e|+1}^{(d', D)} = \frac{\tilde{n}^+_{e, D} - n^+_{|e|, d', D}}{\tilde{n}^+_{e, D} + \tilde{n}^-_{e, D} - \frac{1}{\alpha_{|e|}^{(d', D)}}n^+_{|e|, d', D}}. \label{eq:alpha_e+1}
\end{equation}

\begin{proof}
By Definition 2.5, we have
\begin{equation*}
\alpha_{|e|+1}^{(d', D)} = \frac{n^+_{|e|+1, d', D}}{n_{|e|+1, d', D}} = \frac{n^+_{|e|+1, d', D}}{n^+_{|e|+1, d', D} + n^-_{|e|+1, d', D}}.
\end{equation*}
Applying Equations (\ref{eq:n+_e+1}) and (\ref{eq:n-_e+1}) in Observation 11.3, we have
\begin{align*}
\alpha_{|e|+1}^{(d', D)} &= \frac{\tilde{n}^+_{e, D} - n^+_{|e|, d', D}}{(\tilde{n}^+_{e, D} - n^+_{|e|, d', D}) + (\tilde{n}^-_{e, D} - n^-_{|e|, d', D})} \\
&= \frac{\tilde{n}^+_{e, D} - n^+_{|e|, d', D}}{\tilde{n}^+_{e, D} + \tilde{n}^-_{e, D} - n^+_{|e|, d', D} - n^-_{|e|, d', D}}.
\end{align*}
Applying Equation (\ref{eq:n-_e}) in Observation 11.2, we have
\begin{align*}
\alpha_{|e|+1}^{(d', D)} &= \frac{\tilde{n}^+_{e, D} - n^+_{|e|, d', D}}{\tilde{n}^+_{e, D} + \tilde{n}^-_{e, D} - n^+_{|e|, d', D} - \left(\frac{1}{\alpha_{|e|}^{(d', D)}} - 1\right)n^+_{|e|, d', D}} \\
&= \frac{\tilde{n}^+_{e, D} - n^+_{|e|, d', D}}{\tilde{n}^+_{e, D} + \tilde{n}^-_{e, D} - \frac{1}{\alpha_{|e|}^{(d', D)}}n^+_{|e|, d', D}}.
\end{align*}
\end{proof}

We are now ready to prove Theorem 5.2.

\begin{proof}[Proof of Theorem 5.2]
Let $\mathcal{D}(\mathcal{X}, D, e)$ be the set of (hypothetical and non-hypothetical) rule lists that begin with $e$ and are compatible with $D$, and let $\mathcal{D}(\mathcal{X}, D, e, k)$ be the subset of $\mathcal{D}(\mathcal{X}, D, e)$, consisting of those rule lists in $\mathcal{D}(\mathcal{X}, D, e)$ that have exactly $k$ rules (excluding the final else clause) following the prefix $e$. Let $\mathcal{S}(\mathcal{X}, D, e, 1)$ be the subset of $\mathcal{D}(\mathcal{X}, D, e, 1)$, consisting of those rule lists
\begin{equation*}
d' = \{e, (a_{|e|}^{(d')}, \alpha_{|e|}^{(d', D)}), \alpha_{|e|+1}^{(d', D)}\} \in \mathcal{D}(\mathcal{X}, D, e, 1)
\end{equation*}
with $\alpha_{|e|}^{(d', D)} > 1/(1+w)$ and $\alpha_{|e|+1}^{(d', D)} \leq 1/(1+w)$.

Note that we have $\mathcal{D}(\mathcal{X}, D, e, 0) = \{\bar{e}\}$, where $\bar{e} = \{e, \tilde{\alpha}_{e, D}\}$ is the rule list in which the final else clause immediately follows the prefix $e$, and the probability estimate of the final else clause is $\tilde{\alpha}_{e, D}$, by a similar argument as that given in the proof of Theorem 4.6 for $\mathcal{F}(\mathcal{X}, D, e, 0) = \{\bar{e}\}$.

Let $d \in \mathcal{D}(\mathcal{X}, D, e)$.

The lemma that we have proved in this section, along with its proof, implies
\begin{equation}\label{eq:softobj_ineq}
\tilde{L}(d, D, 1/(1+w), w, C, C_1) \geq \inf_{d' \in \mathcal{S}(\mathcal{X}, D, e, 1) \bigcup \mathcal{D}(\mathcal{X}, D, e, 0)} \tilde{L}(d', D, 1/(1+w), w, C, C_1).
\end{equation}
This is because if $d$ obeys Case 1 in the proof of the lemma, then using the same argument as in the proof of the lemma we can construct a rule list $d_1 = \{e, (a_{|e|}^{(d_1)}, \alpha_{|e|}^{(d_1, D)}), \alpha_{|e|+1}^{(d_1, D)}\} \in \mathcal{S}(\mathcal{X}, D, e, 1)$ that satisfies
\begin{equation}\label{eq:softobj_case1}
\tilde{L}(d, D, 1/(1+w), w, C, C_1) \geq \tilde{L}(d_1, D, 1/(1+w), w, C, C_1).
\end{equation}
Since $d_1$ must also obey
\begin{align}
\tilde{L}(d_1, D, 1/(1+w), w, C, C_1) &\geq \inf_{d' \in \mathcal{S}(\mathcal{X}, D, e, 1)} \tilde{L}(d', D, 1/(1+w), w, C, C_1) \nonumber\\
&\geq \inf_{d' \in \mathcal{S}(\mathcal{X}, D, e, 1) \bigcup \mathcal{D}(\mathcal{X}, D, e, 0)} \tilde{L}(d', D, 1/(1+w), w, C, C_1) \label{eq:softobj_d1},
\end{align}
combining the inequalities in (\ref{eq:softobj_case1}) and (\ref{eq:softobj_d1}) gives us (\ref{eq:softobj_ineq}). On the other hand, if $d$ obeys Case 2 in the proof of the lemma, then by the lemma itself we know
\begin{equation}\label{eq:softobj_case2}
\tilde{L}(d, D, 1/(1+w), w, C, C_1) \geq \tilde{L}(\bar{e}, D, 1/(1+w), w, C, C_1).
\end{equation}
Since we have $\mathcal{D}(\mathcal{X}, D, e, 0) = \{\bar{e}\}$, it is straightforward to see
\begin{align}
\tilde{L}(\bar{e}, D, 1/(1+w), w, C, C_1) &= \inf_{d' \in \mathcal{D}(\mathcal{X}, D, e, 0)} \tilde{L}(d', D, 1/(1+w), w, C, C_1) \nonumber\\
&\geq \inf_{d' \in \mathcal{S}(\mathcal{X}, D, e, 1) \bigcup \mathcal{D}(\mathcal{X}, D, e, 0)} \tilde{L}(d', D, 1/(1+w), w, C, C_1) \label{eq:softobj_ebar_ineq}.
\end{align}
Combining the inequalities in (\ref{eq:softobj_case2}) and (\ref{eq:softobj_ebar_ineq}) again gives us (\ref{eq:softobj_ineq}).

Note that if $\mathcal{S}(\mathcal{X}, D, e, 1)$ is not empty, then the right-hand side of (\ref{eq:softobj_ineq}) can be expressed as
\begin{align}
&\quad \inf_{d' \in \mathcal{S}(\mathcal{X}, D, e, 1) \bigcup \mathcal{D}(\mathcal{X}, D, e, 0)} \tilde{L}(d', D, 1/(1+w), w, C, C_1) \nonumber\\
&= \inf_{d' \in \mathcal{S}(\mathcal{X}, D, e, 1) \bigcup \{\bar{e}\}} \tilde{L}(d', D, 1/(1+w), w, C, C_1) \nonumber\\
&= \min\left(\inf_{d' \in \mathcal{S}(\mathcal{X}, D, e, 1)} \tilde{L}(d', D, 1/(1+w), w, C, C_1), \tilde{L}(\bar{e}, D, 1/(1+w), w, C, C_1)\right). \label{eq:bound_Ltilde}
\end{align}

The rest of the proof proceeds in six steps.

{\bf Step 1.} Compute $\tilde{L}(\bar{e}, D, 1/(1+w), w, C, C_1)$.

Since the contribution by the final else clause to $\tilde{L}(\bar{e}, D, 1/(1+w), w, C, C_1)$ is given by $R_{|e|}(\bar{e}, D, 1/(1+w), w) + \lfloor\tilde{\alpha}_{e, D} - \alpha_{\min}^{(e, D)}\rfloor_+$, where $R_{|e|}(\bar{e}, D, 1/(1+w), w)$ is defined by  Equation (\ref{eq:jth_contrib}) and is given by
\[
R_{|e|}(\bar{e}, D, 1/(1+w), w) = \begin{cases}
\frac{1}{n}n^-_{|e|, \bar{e}, D} &\text{ if } \tilde{\alpha}_{e, D} > 1/(1+w) \\
\frac{w}{n}n^+_{|e|, \bar{e}, D} &\text{ otherwise,}
\end{cases}
\]
and since Observation 11.1 implies $\tilde{n}^+_{e, D} = n^+_{|e|, \bar{e}, D}$ and $\tilde{n}^-_{e, D} = n^-_{|e|, \bar{e}, D}$, it is not difficult to see
\begin{align*}
&\quad \tilde{L}(\bar{e}, D, 1/(1+w), w, C, C_1) \\
&= \begin{cases}
\tilde{L}(e, D, 1/(1+w), w, C, C_1) + \frac{1}{n}\tilde{n}^-_{e, D} + C_1 \lfloor\tilde{\alpha}_{e, D} - \alpha_{\min}^{(e, D)}\rfloor_+ &\text{ if } \tilde{\alpha}_{e, D} > 1/(1+w) \\
\tilde{L}(e, D, 1/(1+w), w, C, C_1) + \frac{w}{n}\tilde{n}^+_{e, D} + C_1 \lfloor\tilde{\alpha}_{e, D} - \alpha_{\min}^{(e, D)}\rfloor_+ &\text{ otherwise.}
\end{cases}
\end{align*}

Since $\tilde{\alpha}_{e, D} > 1/(1+w)$ is equivalent to $\tilde{n}^+_{e, D}/(\tilde{n}^+_{e, D} + \tilde{n}^-_{e, D}) > 1/(1+w)$, or $w\tilde{n}^+_{e, D} > \tilde{n}^-_{e, D}$, and similarly $\tilde{\alpha}_{e, D} \leq 1/(1+w)$ is equivalent to $w\tilde{n}^+_{e, D} \leq \tilde{n}^-_{e, D}$, we can write
\begin{equation}\label{eq:softobj_ebar}
\begin{split}
&\quad \tilde{L}(\bar{e}, D, 1/(1+w), w, C, C_1) \\&= \min\left(\tilde{L}(e, D, 1/(1+w), w, C, C_1) + \frac{1}{n}\tilde{n}^-_{e, D} + C_1 \lfloor\tilde{\alpha}_{e, D} - \alpha_{\min}^{(e, D)}\rfloor_+, \right. \\
  &\quad\quad\quad \left. \vphantom{\frac{1}{n}} \tilde{L}(e, D, 1/(1+w), w, C, C_1) + \frac{w}{n}\tilde{n}^+_{e, D} + C_1 \lfloor\tilde{\alpha}_{e, D} - \alpha_{\min}^{(e, D)}\rfloor_+\right) \\
  &= \tilde{L}(e, D, 1/(1+w), w, C, C_1) + \min\left(\frac{w}{n}\tilde{n}^+_{e, D}, \frac{1}{n}\tilde{n}^-_{e, D}\right) + C_1 \lfloor\tilde{\alpha}_{e, D} - \alpha_{\min}^{(e, D)}\rfloor_+.
\end{split}
\end{equation}

{\bf Step 2.} Partition the set $\mathcal{S}(\mathcal{X}, D, e, 1)$ into three subsets based on how the softly falling objective is computed.

For any $d' = \{e, (a_{|e|}^{(d')}, \alpha_{|e|}^{(d', D)}), \alpha_{|e|+1}^{(d', D)}\} \in \mathcal{S}(\mathcal{X}, D, e, 1)$, the softly falling objective is given by
\begin{align}
&\quad \tilde{L}(d', D, 1/(1+w), w, C, C_1) \nonumber\\
&= \tilde{L}(e, D, 1/(1+w), w, C, C_1) + \frac{1}{n}n^-_{|e|, d', D} + \frac{w}{n}n^+_{|e|+1, d', D} + C \nonumber\\&\quad + C_1 \lfloor\alpha_{|e|}^{(d', D)} - \alpha_{\min}^{(e, D)}\rfloor_+ + C_1 \lfloor\alpha_{|e|+1}^{(d', D)} - \alpha_{\min}^{(e, D)}\rfloor_+. \label{eq:Ltilde}
\end{align}
This is because for any $d' \in \mathcal{S}(\mathcal{X}, D, e, 1)$, the contribution by both the $|e|$-th rule and the final else clause to $\tilde{L}(d', D, 1/(1+w), w, C, C_1)$ is given by 
\[
R_{|e|}(d', D, 1/(1+w), w) + R_{|e|+1}(d', D, 1/(1+w), w) + C +  C_1 \lfloor\alpha_{|e|}^{(d', D)} - \alpha_{\min}^{(e, D)}\rfloor_+ + C_1 \lfloor\alpha_{|e|+1}^{(d', D)} - \alpha_{\min}^{(e, D)}\rfloor_+,
\]
where $R_{|e|}(d', D, 1/(1+w), w)$ and $R_{|e|+1}(d', D, 1/(1+w), w)$ are defined by Equation (\ref{eq:jth_contrib}) and are given by
\[
R_{|e|}(d', D, 1/(1+w), w) = \frac{1}{n}n^-_{|e|, d', D} \quad\text{ and }\quad R_{|e|+1}(d', D, 1/(1+w), w) = \frac{w}{n}n^+_{|e|+1, d', D}
\]
(because we have $\alpha_{|e|}^{(d', D)} > 1/(1+w)$ and $\alpha_{|e|+1}^{(d', D)} \leq 1/(1+w)$ for $d' \in \mathcal{S}(\mathcal{X}, D, e, 1)$).


Let
\begin{equation*}
\mathcal{S}_1(\mathcal{X}, D, e, 1) = \{d' = \{e, (a_{|e|}^{(d')}, \alpha_{|e|}^{(d', D)}), \alpha_{|e|+1}^{(d', D)}\} \in \mathcal{S}(\mathcal{X}, D, e, 1): \alpha_{\min}^{(e, D)} \geq \alpha_{|e|}^{(d', D)} > \alpha_{|e|+1}^{(d', D)}\},
\end{equation*}
\begin{equation*}
\mathcal{S}_2(\mathcal{X}, D, e, 1) = \{d' = \{e, (a_{|e|}^{(d')}, \alpha_{|e|}^{(d', D)}), \alpha_{|e|+1}^{(d', D)}\} \in \mathcal{S}(\mathcal{X}, D, e, 1): \alpha_{|e|}^{(d', D)} > \alpha_{\min}^{(e, D)} \geq \alpha_{|e|+1}^{(d', D)}\},
\end{equation*}
and
\begin{equation*}
\mathcal{S}_3(\mathcal{X}, D, e, 1) = \{d' = \{e, (a_{|e|}^{(d')}, \alpha_{|e|}^{(d', D)}), \alpha_{|e|+1}^{(d', D)}\} \in \mathcal{S}(\mathcal{X}, D, e, 1): \alpha_{|e|}^{(d', D)} > \alpha_{|e|+1}^{(d', D)} > \alpha_{\min}^{(e, D)}\},
\end{equation*}

It is easy to see
\begin{equation*}
\mathcal{S}(\mathcal{X}, D, e, 1) = \mathcal{S}_3(\mathcal{X}, D, e, 1) \cup \mathcal{S}_1(\mathcal{X}, D, e, 1) \cup \mathcal{S}_2(\mathcal{X}, D, e, 1).
\end{equation*}

We observe here that given the prefix $e$, we can write $\tilde{L}(d', D, 1/(1+w), w, C, C_1)$ as a function of $n^+_{|e|, d', D}$ and $\alpha_{|e|}^{(d', D)}$, by substituting (\ref{eq:n-_e}), (\ref{eq:n+_e+1}), and (\ref{eq:alpha_e+1}) in Observations 11.1, 11.2, and 11.3 into (\ref{eq:Ltilde}).

{\bf Step 3.} Determine a lower bound of $\tilde{L}(d', D, 1/(1+w), w, C, C_1)$ for all $d' \in \mathcal{S}_1(\mathcal{X}, D, e, 1)$.

Let $d' = \{e, (a_{|e|}^{(d')}, \alpha_{|e|}^{(d', D)}), \alpha_{|e|+1}^{(d', D)}\} \in \mathcal{S}_1(\mathcal{X}, D, e, 1)$. By the definition of $\mathcal{S}_1(\mathcal{X}, D, e, 1)$, we have
\begin{equation}\label{eq:alpha_ineq_s1}
\alpha_{\min}^{(e, D)} \geq \alpha_{|e|}^{(d', D)} > \frac{1}{1+w} \geq \alpha_{|e|+1}^{(d', D)}
\end{equation}
We first prove the following inequality
\begin{equation}\label{eq:tilde_alpha_ineq_s1}
\alpha_{\min}^{(e, D)} \geq \alpha_{|e|}^{(d', D)} > \max(1/(1+w), \tilde{\alpha}_{e, D}),
\end{equation}
which will be useful later.

To prove (\ref{eq:tilde_alpha_ineq_s1}), we use Definition 2.5 as well as (\ref{eq:n+_e+1}) and (\ref{eq:n_e+1}) in Observation 11.3 to obtain
\begin{equation}\label{eq:tilde_alpha}
\tilde{\alpha}_{e, D} = \frac{\tilde{n}^+_{e, D}}{\tilde{n}_{e, D}} = \frac{n^+_{|e|, d', D} + n^+_{|e|+1, d', D}}{n_{|e|, d', D} + n_{|e|+1, d', D}} = \frac{\alpha_{|e|}^{(d', D)}n_{|e|, d', D} + \alpha_{|e|+1}^{(d', D)}n_{|e|+1, d', D}}{n_{|e|, d', D} + n_{|e|+1, d', D}}.
\end{equation}
Substituting $\alpha_{|e|+1}^{(d', D)} < \alpha_{|e|}^{(d', D)}$ from (\ref{eq:alpha_ineq_s1}) into (\ref{eq:tilde_alpha}), we obtain $\tilde{\alpha}_{e, D} < \alpha_{|e|}^{(d', D)}$. Combining this inequality with $\alpha_{\min}^{(e, D)} \geq \alpha_{|e|}^{(d', D)} > \frac{1}{1+w}$ from (\ref{eq:alpha_ineq_s1}), we obtain (\ref{eq:tilde_alpha_ineq_s1}), as desired.

Note that since (\ref{eq:tilde_alpha_ineq_s1}) has to hold for any $d' \in \mathcal{S}_1(\mathcal{X}, D, e, 1)$, if $\alpha_{\min}^{(e, D)} \leq \max(1/(1+w), \tilde{\alpha}_{e, D})$ is true for the given prefix $e$, then $\mathcal{S}_1(\mathcal{X}, D, e, 1)$ is empty.

We now show that given the prefix $e$, the softly falling objective $\tilde{L}(d', D, 1/(1+w), w, C, C_1)$ for $d'$ is a monotonically decreasing function of both $n^+_{|e|, d', D}$ and $\alpha_{|e|}^{(d', D)}$.

To do so, we substitute (\ref{eq:n-_e}) and (\ref{eq:n+_e+1}) in Observations 11.1 and 11.2 into (\ref{eq:Ltilde}) to obtain
\begin{align}
&\quad \tilde{L}(d', D, 1/(1+w), w, C, C_1) \nonumber\\&= \tilde{L}(e, D, 1/(1+w), w, C, C_1) + \frac{1}{n}\left(\left(\frac{1}{\alpha_{|e|}^{(d', D)}} - 1 - w\right)n^+_{|e|, d', D} + w\tilde{n}^+_{e, D}\right) + C. \label{eq:Ltilde_s1}
\end{align}
Note that Equation (\ref{eq:Ltilde_s1}) shows that given the prefix $e$, $\tilde{L}(d', D, 1/(1+w), w, C, C_1)$ is a function of $n^+_{|e|, d', D}$ and $\alpha_{|e|}^{(d', D)}$. Since we have
\begin{equation*}
\frac{\partial \tilde{L}(d', D, 1/(1+w), w, C, C_1)}{\partial n^+_{|e|, d', D}} = \frac{1}{n}\left(\frac{1}{\alpha_{|e|}^{(d', D)}} - 1 - w\right) < 0
\end{equation*}
because $\alpha_{|e|}^{(d', D)} > 1/(1+w)$ holds for any $d' \in \mathcal{S}_1(\mathcal{X}, D, e, 1)$, and
\begin{equation*}
\frac{\partial \tilde{L}(d', D, 1/(1+w), w, C, C_1)}{\partial \alpha_{|e|}^{(d', D)}} = -\frac{n^+_{|e|, d', D}}{n}\frac{1}{(\alpha_{|e|}^{(d', D)})^2} \leq 0,
\end{equation*}
we see that $\tilde{L}(d', D, 1/(1+w), w, C, C_1)$ is indeed a monotonically decreasing function of both $n^+_{|e|, d', D}$ and $\alpha_{|e|}^{(d', D)}$. Thus, we can obtain a lower bound of $\tilde{L}(d', D, 1/(1+w), w, C, C_1)$ by substituting $n^+_{|e|, d', D}$ and $\alpha_{|e|}^{(d', D)}$ with their respective upper bound. The inequality $n^+_{|e|, d', D} \leq \tilde{n}^+_{e, D}$ in Observation 11.3 gives an upper bound for $n^+_{|e|, d', D}$, and the inequality $\alpha_{|e|}^{(d', D)} \leq \alpha_{\min}^{(e, D)}$ from (\ref{eq:alpha_ineq_s1}) gives an upper bound for $\alpha_{|e|}^{(d', D)}$. Substituting these upper bounds into (\ref{eq:Ltilde_s1}), we obtain the following inequality, which gives a lower bound of $\tilde{L}(d', D, 1/(1+w), w, C, C_1)$:
\begin{align*}
&\quad \tilde{L}(d', D, 1/(1+w), w, C, C_1) \\&\geq \tilde{L}(e, D, 1/(1+w), w, C, C_1) + \frac{1}{n}\left(\left(\frac{1}{\alpha_{\min}^{(e, D)}} - 1 - w\right)\tilde{n}^+_{e, D} + w\tilde{n}^+_{e, D}\right) + C \\
&= \tilde{L}(e, D, 1/(1+w), w, C, C_1) + \frac{1}{n}\left(\frac{1}{\alpha_{\min}^{(e, D)}} - 1\right)\tilde{n}^+_{e, D} + C.
\end{align*}

{\bf Step 4.} Determine a lower bound of $\tilde{L}(d', D, 1/(1+w), w, C, C_1)$ for all $d' \in \mathcal{S}_2(\mathcal{X}, D, e, 1)$.

Let $d' = \{e, (a_{|e|}^{(d')}, \alpha_{|e|}^{(d', D)}), \alpha_{|e|+1}^{(d', D)}\} \in \mathcal{S}_2(\mathcal{X}, D, e, 1)$. By the definition of $\mathcal{S}_2(\mathcal{X}, D, e, 1)$, we have
\begin{equation}\label{eq:alpha_ineq_s}
\alpha_{|e|}^{(d', D)} > \frac{1}{1+w}
\end{equation}
and
\begin{equation}\label{eq:alpha_ineq_s2}
\alpha_{|e|}^{(d', D)} > \alpha_{\min}^{(e, D)} \geq \alpha_{|e|+1}^{(d', D)}
\end{equation}
We first prove the following inequality
\begin{equation}\label{eq:tilde_alpha_ineq_s2}
1 \geq \alpha_{|e|}^{(d', D)} > \max(\alpha_{\min}^{(e, D)}, \tilde{\alpha}_{e, D}, 1/(1+w)) = \zeta,
\end{equation}
which will be useful later.

To prove (\ref{eq:tilde_alpha_ineq_s2}), we use Definition 2.5 as well as (\ref{eq:n+_e+1}) and (\ref{eq:n_e+1}) in Observation 11.3 to obtain (\ref{eq:tilde_alpha}). Substituting $\alpha_{|e|+1}^{(d', D)} < \alpha_{|e|}^{(d', D)}$ from (\ref{eq:alpha_ineq_s1}) into (\ref{eq:tilde_alpha}), we obtain $\tilde{\alpha}_{e, D} < \alpha_{|e|}^{(d', D)}$. Combining this inequality with (\ref{eq:alpha_ineq_s}) and $\alpha_{|e|}^{(d', D)} > \alpha_{\min}^{(e, D)}$ from (\ref{eq:alpha_ineq_s2}), we obtain (\ref{eq:tilde_alpha_ineq_s2}), as desired.

We now show that given the prefix $e$ and a particular value of $\alpha_{|e|}^{(d', D)}$ that obeys (\ref{eq:tilde_alpha_ineq_s2}), the softly falling objective $\tilde{L}(d', D, 1/(1+w), w, C, C_1)$ for $d'$ is a decreasing function of $n^+_{|e|, d', D}$.

To do so, we substitute (\ref{eq:n-_e}) and (\ref{eq:n+_e+1}) in Observations 11.1 and 11.2 into (\ref{eq:Ltilde}) to obtain
\begin{align}
&\quad \tilde{L}(d', D, 1/(1+w), w, C, C_1) \nonumber\\&= \tilde{L}(e, D, 1/(1+w), w, C, C_1) + \frac{1}{n}\left(\left(\frac{1}{\alpha_{|e|}^{(d', D)}} - 1 - w\right)n^+_{|e|, d', D} + w\tilde{n}^+_{e, D}\right) + C \nonumber\\&\quad + C_1(\alpha_{|e|}^{(d', D)} - \alpha_{\min}^{(e, D)}). \label{eq:Ltilde_s2}
\end{align}
Note that Equation (\ref{eq:Ltilde_s2}) shows that given the prefix $e$, $\tilde{L}(d', D, 1/(1+w), w, C, C_1)$ is a function of $n^+_{|e|, d', D}$ and $\alpha_{|e|}^{(d', D)}$. Differentiating $\tilde{L}(d', D, 1/(1+w), w, C, C_1)$ given in (\ref{eq:Ltilde_s2}) with respect to $n^+_{|e|, d', D}$, we obtain
\begin{equation}\label{eq:der_tildeL_s2}
\frac{\partial \tilde{L}(d', D, 1/(1+w), w, C, C_1)}{\partial n^+_{|e|, d', D}} = \frac{1}{n}\left(\frac{1}{\alpha_{|e|}^{(d', D)}} - 1 - w\right).
\end{equation}
Since $\alpha_{|e|}^{(d', D)}$ obeys (\ref{eq:tilde_alpha_ineq_s2}), in particular, it obeys $\alpha_{|e|}^{(d', D)} > 1/(1+w)$, we have
\begin{equation*}
\frac{1}{\alpha_{|e|}^{(d', D)}} - 1 - w < 0,
\end{equation*}
which then gives $\partial \tilde{L}(d', D, 1/(1+w), w, C, C_1)/\partial n^+_{|e|, d', D} < 0$. This means that given the prefix $e$ and a particular value of $\alpha_{|e|}^{(d', D)}$ that obeys (\ref{eq:tilde_alpha_ineq_s2}), $\tilde{L}(d', D, 1/(1+w), w, C, C_1)$ is a decreasing function of $n^+_{|e|, d', D}$.

Thus, given the prefix $e$ and a particular value of $\alpha_{|e|}^{(d', D)}$ that obeys (\ref{eq:tilde_alpha_ineq_s2}), we can obtain a lower bound of $\tilde{L}(d', D, 1/(1+w), w, C, C_1)$ by substituting $n^+_{|e|, d', D}$ with its upper bound. The inequality $n^+_{|e|, d', D} \leq \tilde{n}^+_{e, D}$ in Observation 11.3 gives an upper bound for $n^+_{|e|, d', D}$. Substituting $n^+_{|e|, d', D}$ with its upper bound $\tilde{n}^+_{e, D}$ into (\ref{eq:Ltilde_s2}), we obtain a lower bound of $\tilde{L}(d', D, 1/(1+w), w, C, C_1)$, denoted by $\tilde{g}(\alpha_{|e|}^{(d', D)})$, when $\alpha_{|e|}^{(d', D)}$ is held constant:
\begin{align*}
\tilde{g}(\alpha_{|e|}^{(d', D)}) &= \tilde{L}(e, D, 1/(1+w), w, C, C_1) + \frac{1}{n}\left(\frac{1}{\alpha_{|e|}^{(d', D)}} - 1\right)\tilde{n}^+_{e, D} + C + C_1(\alpha_{|e|}^{(d', D)} - \alpha_{\min}^{(e, D)}) \\
&= \tilde{L}(e, D, 1/(1+w), w, C, C_1) + g(\alpha_{|e|}^{(d', D)})
\end{align*}
where $g$ is defined in the statement of the theorem. In other words, given the prefix $e$ and a particular value of $\alpha_{|e|}^{(d', D)}$ that obeys (\ref{eq:tilde_alpha_ineq_s2}), we have $\tilde{L}(d', D, 1/(1+w), w, C, C_1) \geq \tilde{g}(\alpha_{|e|}^{(d', D)})$. Since (\ref{eq:tilde_alpha_ineq_s2}) is true for any $d' \in \mathcal{S}_2(\mathcal{X}, D, e, 1)$, we always have $\tilde{L}(d', D, 1/(1+w), w, C, C_1) \geq \tilde{g}(\alpha_{|e|}^{(d', D)})$ for any $d' \in \mathcal{S}_2(\mathcal{X}, D, e, 1)$. This implies
\begin{align*}
\tilde{L}(d', D, 1/(1+w), w, C, C_1) &\geq \inf_{\alpha_{|e|}^{(d', D)}: \zeta < \alpha_{|e|}^{(d', D)} \leq 1} \tilde{g}(\alpha_{|e|}^{(d', D)}) \\
&= \tilde{L}(e, D, 1/(1+w), w, C, C_1) + \inf_{\alpha_{|e|}^{(d', D)}: \zeta < \alpha_{|e|}^{(d', D)} \leq 1} g(\alpha_{|e|}^{(d', D)}).
\end{align*}

{\bf Step 5.} Determine a lower bound of $\tilde{L}(d', D, 1/(1+w), w, C, C_1)$ for all $d' \in \mathcal{S}_3(\mathcal{X}, D, e, 1)$.

Let $d' = \{e, (a_{|e|}^{(d')}, \alpha_{|e|}^{(d', D)}), \alpha_{|e|+1}^{(d', D)}\} \in \mathcal{S}_3(\mathcal{X}, D, e, 1)$. By the definition of $\mathcal{S}_3(\mathcal{X}, D, e, 1)$, we have
\begin{equation}\label{eq:alpha_ineq_s3}
\alpha_{|e|}^{(d', D)} > \frac{1}{1+w} \geq \alpha_{|e|+1}^{(d', D)} > \alpha_{\min}^{(e, D)}.
\end{equation}
We first prove the following inequality
\begin{equation}\label{eq:tilde_alpha_ineq_s3}
1 \geq \alpha_{|e|}^{(d', D)} > \max(\alpha_{\min}^{(e, D)}, \tilde{\alpha}_{e, D}, 1/(1+w)) = \zeta,
\end{equation}
which will be useful later.

To prove (\ref{eq:tilde_alpha_ineq_s3}), we use Definition 2.5 as well as (\ref{eq:n+_e+1}) and (\ref{eq:n_e+1}) in Observation 11.3 to obtain (\ref{eq:tilde_alpha}). Substituting $\alpha_{|e|+1}^{(d', D)} < \alpha_{|e|}^{(d', D)}$ from (\ref{eq:alpha_ineq_s3}) into (\ref{eq:tilde_alpha}), we obtain $\tilde{\alpha}_{e, D} < \alpha_{|e|}^{(d', D)}$. Combining this inequality with $\alpha_{|e|}^{(d', D)} > \frac{1}{1+w} > \alpha_{\min}^{(e, D)}$ from (\ref{eq:alpha_ineq_s3}), we obtain (\ref{eq:tilde_alpha_ineq_s3}), as desired.


To determine a lower bound of $\tilde{L}(d', D, 1/(1+w), w, C, C_1)$, we observe
\begin{align}
&\quad \tilde{L}(d', D, 1/(1+w), w, C, C_1) \nonumber\\
&\geq \tilde{L}(e, D, 1/(1+w), w, C, C_1) + \frac{1}{n}n^-_{|e|, d', D} + \frac{w}{n}n^+_{|e|+1, d', D} + C + C_1 \lfloor\alpha_{|e|}^{(d', D)} - \alpha_{\min}^{(e, D)}\rfloor_+ \label{eq:Ltilde_s3_ineq}\\
&= \tilde{L}(e, D, 1/(1+w), w, C, C_1) + \frac{1}{n}\left(\left(\frac{1}{\alpha_{|e|}^{(d', D)}} - 1 - w\right)n^+_{|e|, d', D} + w\tilde{n}^+_{e, D}\right) + C \nonumber\\&\quad + C_1 (\alpha_{|e|}^{(d', D)} - \alpha_{\min}^{(e, D)}) \label{eq:Ltilde_s3_eq}
\end{align}
where the last equality follows by substituting (\ref{eq:n-_e}) and (\ref{eq:n+_e+1}) in Observations 11.1 and 11.2 into (\ref{eq:Ltilde_s3_ineq}). Using (\ref{eq:tilde_alpha_ineq_s3}) and applying the same argument as in Step 4, the quantity labeled (\ref{eq:Ltilde_s3_eq}) is also lower-bounded by
\begin{equation*}
\tilde{L}(e, D, 1/(1+w), w, C, C_1) + \inf_{\alpha_{|e|}^{(d', D)}: \zeta < \alpha_{|e|}^{(d', D)} \leq 1} g(\alpha_{|e|}^{(d', D)}),
\end{equation*}
so that we again have
\begin{equation*}
\tilde{L}(d', D, 1/(1+w), w, C, C_1) \geq \tilde{L}(e, D, 1/(1+w), w, C, C_1) + \inf_{\alpha_{|e|}^{(d', D)}: \zeta < \alpha_{|e|}^{(d', D)} \leq 1} g(\alpha_{|e|}^{(d', D)}).
\end{equation*}

{\bf Step 6.} Put everything together.

Suppose, first, that $\mathcal{S}(\mathcal{X}, D, e, 1)$ is not empty.

In the case where $\mathcal{S}_1(\mathcal{X}, D, e, 1)$ is not empty, we observe the following inequality
\begin{equation}\label{eq:inf_S1}
\inf_{d' \in \mathcal{S}_1(\mathcal{X}, D, e, 1)} \tilde{L}(d', D, 1/(1+w), w, C, C_1) \geq \tilde{L}(e, D, 1/(1+w), w, C, C_1) + \frac{1}{n}\left(\frac{1}{\alpha_{\min}^{(e, D)}} - 1\right)\tilde{n}^+_{e, D} + C,
\end{equation}
which follows from the definition of $\inf$ being the greatest lower bound, as well as the lower bound of $\tilde{L}(d', D, 1/(1+w), w, C, C_1)$ for $d' \in \mathcal{S}_1(\mathcal{X}, D, e, 1)$, which we have derived in Step 3.

In the case where $\mathcal{S}_2(\mathcal{X}, D, e, 1) \cup \mathcal{S}_3(\mathcal{X}, D, e, 1)$ is not empty, we observe the following inequality
\begin{equation}\label{eq:inf_S2_S3}
\inf_{d' \in \mathcal{S}_2(\mathcal{X}, D, e, 1) \cup \mathcal{S}_3(\mathcal{X}, D, e, 1)} \tilde{L}(d', D, 1/(1+w), w, C, C_1) \geq \tilde{L}(e, D, 1/(1+w), w, C, C_1) + \inf_{\beta: \zeta < \beta \leq 1} g(\beta),
\end{equation}
which follows from the definition of $\inf$ being the greatest lower bound, as well as the lower bound of $\tilde{L}(d', D, 1/(1+w), w, C, C_1)$ for $d' \in \mathcal{S}_2(\mathcal{X}, D, e, 1)$, which we have derived in Step 4, and the lower bound of $\tilde{L}(d', D, 1/(1+w), w, C, C_1)$ for $d' \in \mathcal{S}_3(\mathcal{X}, D, e, 1)$, which we have derived in Step 5.

To derive a lower bound of $\tilde{L}(d', D, 1/(1+w), w, C, C_1)$ for $d' \in \mathcal{S}(\mathcal{X}, D, e, 1)$, we further observe that if $\alpha_{\min}^{(e, D)} \leq \max(1/(1+w), \tilde{\alpha}_{e, D})$ holds, then by our remark in Step 3, $\mathcal{S}_1(\mathcal{X}, D, e, 1)$ is empty, and consequently, using (\ref{eq:inf_S2_S3}), we have
\begin{align}
\inf_{d' \in \mathcal{S}(\mathcal{X}, D, e, 1)} \tilde{L}(d', D, 1/(1+w), w, C, C_1) &= \inf_{d' \in \mathcal{S}_2(\mathcal{X}, D, e, 1) \cup \mathcal{S}_3(\mathcal{X}, D, e, 1)} \tilde{L}(d', D, 1/(1+w), w, C, C_1) \nonumber\\
&\geq \tilde{L}(e, D, 1/(1+w), w, C, C_1) + \inf_{\beta: \zeta < \beta \leq 1} g(\beta). \label{eq:inf_S1_empty}
\end{align}

On the other hand, if $\alpha_{\min}^{(e, D)} > \max(1/(1+w), \tilde{\alpha}_{e, D})$ holds, then $\mathcal{S}_1(\mathcal{X}, D, e, 1)$ may or may not be empty. If, in addition, both $\mathcal{S}_1(\mathcal{X}, D, e, 1)$ and $\mathcal{S}_2(\mathcal{X}, D, e, 1) \cup \mathcal{S}_3(\mathcal{X}, D, e, 1)$ are not empty, then using (\ref{eq:inf_S1}) and (\ref{eq:inf_S2_S3}), we have
\begin{align}
&\quad \inf_{d' \in \mathcal{S}(\mathcal{X}, D, e, 1)} \tilde{L}(d', D, 1/(1+w), w, C, C_1) \nonumber\\
&= \min\left(\inf_{d' \in \mathcal{S}_1(\mathcal{X}, D, e, 1)} \tilde{L}(d', D, 1/(1+w), w, C, C_1), \inf_{d' \in \mathcal{S}_2(\mathcal{X}, D, e, 1) \cup \mathcal{S}_3(\mathcal{X}, D, e, 1)} \tilde{L}(d', D, 1/(1+w), w, C, C_1)\right) \nonumber\\
&\geq \tilde{L}(e, D, 1/(1+w), w, C, C_1) + \min\left(\frac{1}{n}\left(\frac{1}{\alpha_{\min}^{(e, D)}} - 1\right)\tilde{n}^+_{e, D} + C, \inf_{\beta: \zeta < \beta \leq 1} g(\beta)\right). \label{eq:inf_S_all}
\end{align}
If either $\mathcal{S}_1(\mathcal{X}, D, e, 1)$ or $\mathcal{S}_2(\mathcal{X}, D, e, 1) \cup \mathcal{S}_3(\mathcal{X}, D, e, 1)$ is empty, then $\inf_{d' \in \mathcal{S}(\mathcal{X}, D, e, 1)} \tilde{L}(d', D, 1/(1+w), w, C, C_1)$ is given by either
\[
\inf_{d' \in \mathcal{S}_2(\mathcal{X}, D, e, 1) \cup \mathcal{S}_3(\mathcal{X}, D, e, 1)} \tilde{L}(d', D, 1/(1+w), w, C, C_1) \quad\text{ or }\quad \inf_{d' \in \mathcal{S}_1(\mathcal{X}, D, e, 1)} \tilde{L}(d', D, 1/(1+w), w, C, C_1),
\]
both of which are lower-bounded by the quantity labeled (\ref{eq:inf_S_all}) because of (\ref{eq:inf_S2_S3}) and (\ref{eq:inf_S1}).  

Putting these cases together, we have
\begin{align}
&\quad \inf_{d' \in \mathcal{S}(\mathcal{X}, D, e, 1)} \tilde{L}(d', D, 1/(1+w), w, C, C_1) \nonumber\\
&\geq \tilde{L}(e, D, 1/(1+w), w, C, C_1) \nonumber\\&\quad + \begin{cases}
\min\left(\frac{1}{n}\left(\frac{1}{\alpha_{\min}^{(e, D)}} - 1\right)\tilde{n}^+_{e, D} + C, \inf_{\beta: \zeta < \beta \leq 1} g(\beta)\right) &\text{ if } \alpha_{\min}^{(e, D)} > \max(1/(1+w), \tilde{\alpha}_{e, D}), \\
\inf_{\beta: \zeta < \beta \leq 1} g(\beta) &\text{ otherwise.}
\end{cases} \label{eq:inf_S}
\end{align}

Combining (\ref{eq:softobj_ineq}), (\ref{eq:bound_Ltilde}), (\ref{eq:softobj_ebar}), and (\ref{eq:inf_S}), we have
\begin{align}
&\quad \tilde{L}(d, D, 1/(1+w), w, C, C_1) \nonumber\\
&\geq \tilde{L}(e, D, 1/(1+w), w, C, C_1) \nonumber\\
&\quad+ \begin{cases}
\min\left(\frac{1}{n}\left(\frac{1}{\alpha_{\min}^{(e, D)}} - 1\right)\tilde{n}^+_{e, D} + C, \inf_{\beta: \zeta < \beta \leq 1} g(\beta), \frac{w}{n}\tilde{n}^+_{e, D} + C_1\lfloor\tilde{\alpha}_{e, D} - \alpha_{\min}^{(e, D)}\rfloor_+, \right. \\
                   \quad\quad\quad \left. \vphantom{\frac{1}{\alpha_{|e|-1}^{(e, D)}} - 1} \frac{1}{n}\tilde{n}^-_{e, D} + C_1\lfloor\tilde{\alpha}_{e, D} - \alpha_{\min}^{(e, D)}\rfloor_+\right) \quad\quad\quad\quad\quad\quad\text{ if } \alpha_{\min}^{(e, D)} > \max(1/(1+w), \tilde{\alpha}_{e, D}), \\
\min\left(\inf_{\beta: \zeta < \beta \leq 1} g(\beta), \frac{w}{n}\tilde{n}^+_{e, D} + C_1\lfloor\tilde{\alpha}_{e, D} - \alpha_{\min}^{(e, D)}\rfloor_+, \right. \\
                   \quad\quad\quad \left. \vphantom{\frac{w}{n}} \frac{1}{n}\tilde{n}^-_{e, D} + C_1\lfloor\tilde{\alpha}_{e, D} - \alpha_{\min}^{(e, D)}\rfloor_+\right) \quad\quad\quad\quad\quad\quad\text{ otherwise.}
\end{cases}\label{eq:softfrl_prefix_bound_by_case}
\end{align}

Note that the quantity labeled (\ref{eq:softfrl_prefix_bound_by_case}) is precisely equal to $\tilde{L}^*(e, D, w, C, C_1)$ given by Equation (\ref{eq:prefix_bound_soft_frl_supp}) in the statement of the theorem, because:\\
(i) if $\alpha_{\min}^{(e, D)} > \max(1/(1+w), \tilde{\alpha}_{e, D})$ holds, then the first term in the minimum on the right-hand side of Equation (\ref{eq:prefix_bound_soft_frl_supp}) is precisely $\frac{1}{n}\left(\frac{1}{\alpha_{\min}^{(e, D)}} - 1\right)\tilde{n}^+_{e, D} + C$;\\
(ii) if $\alpha_{\min}^{(e, D)} > \max(1/(1+w), \tilde{\alpha}_{e, D})$ does not hold, then we have $\alpha_{\min}^{(e, D)} \leq 1/(1+w)$ or $\alpha_{\min}^{(e, D)} \leq \tilde{\alpha}_{e, D})$ -- in the former case where $\alpha_{\min}^{(e, D)} \leq 1/(1+w)$ holds, we have
\[
\frac{1}{n}\left(\frac{1}{\alpha_{\min}^{(e, D)}} - 1\right)\tilde{n}^+_{e, D} \geq \frac{w}{n}\tilde{n}^+_{e, D},
\]
which implies that the first term in the minimum on the right-hand side of Equation (\ref{eq:prefix_bound_soft_frl_supp}) is bounded below by $\frac{w}{n}\tilde{n}^+_{e, D} + C_1\lfloor\tilde{\alpha}_{e, D} - \alpha_{\min}^{(e, D)}\rfloor_+$, and thus has no influence over the computation of the minimum; in the latter case where $\alpha_{\min}^{(e, D)} \leq \tilde{\alpha}_{e, D})$ holds, the first term in the minimum on the right-hand side of Equation (\ref{eq:prefix_bound_soft_frl_supp}) is clearly bounded below by $\frac{w}{n}\tilde{n}^+_{e, D} + C_1\lfloor\tilde{\alpha}_{e, D} - \alpha_{\min}^{(e, D)}\rfloor_+$, and again has no influence over the computation of the minimum.

This proves that $\tilde{L}^*(e, D, w, C, C_1)$ given by Equation (\ref{eq:prefix_bound_soft_frl_supp}) is indeed a lower bound of $\tilde{L}(d, D, 1/(1+w), w, C, C_1)$ for $d \in \mathcal{D}(\mathcal{X}, D, e)$, in the case where $\mathcal{S}(\mathcal{X}, D, e, 1)$ is not empty. In the case where $\mathcal{S}(\mathcal{X}, D, e, 1)$ is empty, using (\ref{eq:softobj_ineq}) and (\ref{eq:softobj_ebar}), along with the fact $\mathcal{D}(\mathcal{X}, D, e, 0) = \{\bar{e}\}$, we have
\begin{align*}
\tilde{L}(d, D, 1/(1+w), w, C, C_1) &\geq \inf_{d' \in \mathcal{D}(\mathcal{X}, D, e, 0)} \tilde{L}(d', D, 1/(1+w), w, C, C_1) \\
&= \tilde{L}(\bar{e}, D, 1/(1+w), w, C, C_1) \\
&= \tilde{L}(e, D, 1/(1+w), w, C, C_1) + \min\left(\frac{w}{n}\tilde{n}^+_{e, D}, \frac{1}{n}\tilde{n}^-_{e, D}\right) + C_1 \lfloor\tilde{\alpha}_{e, D} - \alpha_{\min}^{(e, D)}\rfloor_+,
\end{align*}
where the last quantity is clearly lower-bounded by $\tilde{L}^*(e, D, w, C, C_1)$ defined in Equation (\ref{eq:prefix_bound_soft_frl_supp}). We have now proven that $\tilde{L}^*(e, D, w, C, C_1)$ given by Equation (\ref{eq:prefix_bound_soft_frl_supp}) is a lower bound of $\tilde{L}(d, D, 1/(1+w), w, C, C_1)$ for $d \in \mathcal{D}(\mathcal{X}, D, e)$.

Finally, we compute $\inf_{\beta: \zeta < \beta \leq 1} g(\beta)$ analytically. Since the derivative of $g$ is given by
\begin{equation*}
g'(\beta) = -\frac{\tilde{n}^+_{e, D}}{n\beta^2} + C_1,
\end{equation*}
and $\beta$ must be positive, the only stationary point $\beta^*$ of $g$ that could satisfy the constraint $\zeta < \beta^* \leq 1$ is given by
\begin{equation*}
\beta^* = \sqrt{\frac{\tilde{n}^+_{e, D}}{C_1 n}},
\end{equation*}
and the second derivative test confirms that $\beta^*$ is a local minimum of $g$. It then follows that $\inf_{\beta: \zeta < \beta \leq 1} g(\beta)$ is given by
\begin{equation*}
\inf_{\beta: \zeta < \beta \leq 1} g(\beta) = \begin{cases}
g(\beta^*) &\text{ if } \zeta < \beta^* \leq 1 \\
\min(g(\zeta), g(1)) &\text{ otherwise}.
\end{cases}
\end{equation*}
\end{proof}

\section{Additional Rule Lists Demonstrating the Effect of Varying Parameter Values}


In this section, we include some additional rule lists created using Algorithm FRL and Algorithm softFRL with varying parameter values. The default parameter values we used in creating these rule lists are $w = 7$, $C = 0.000001$, and $C_1 = 0.5$. In each of the following subsections, the rule lists were created with default parameter values, other than the parameter that was being varied.

\subsection{Effect of Varying $w$ on Algorithm FRL}

Running Algorithm FRL with $w = 1$ on the bank-full dataset produces the following falling rule list:

\begin{table}[H]
\centering
\begin{tabular}{llllll}
 & antecedent & & probability  & positive & negative \\
 &            & &              & support  & support  \\\hline
IF      & poutcome=success              & THEN success prob. is & 0.65 & 934  & 495  \\
        & AND loan=no                   &                       &      &      &      \\
ELSE IF & poutcome=success              & THEN success prob. is & 0.62 & 31   & 19   \\
        & AND marital=married           &                       &      &      &      \\
ELSE IF & poutcome=success              & THEN success prob. is & 0.56 & 9    & 7    \\
        & AND campaign=1                &                       &      &      &      \\
ELSE    &                               & success prob. is      & 0.10 & 4315 & 39401
\end{tabular}
\centering
\caption{Falling rule list for bank-full dataset, created using Algorithm FRL with $w = 1$}
\end{table}

Running Algorithm FRL with $w = 3$ on the bank-full dataset produces the following falling rule list:

\begin{table}[H]
\centering
\begin{tabular}{llllll}
 & antecedent & & probability  & positive & negative \\
 &            & &              & support  & support  \\\hline
IF      & poutcome=success              & THEN success prob. is & 0.65 & 677  & 361  \\
        & AND previous $\geq$ 2         &                       &      &      &      \\
ELSE IF & poutcome=success              & THEN success prob. is & 0.65 & 185  & 99   \\
        & AND campaign=1                &                       &      &      &      \\
ELSE IF & poutcome=success              & THEN success prob. is & 0.63 & 111  & 65   \\
        & AND loan=no                   &                       &      &      &      \\
ELSE IF & poutcome=success              & THEN success prob. is & 0.56 & 5    & 4    \\
        & AND marital=married           &                       &      &      &      \\
ELSE IF & 60 $\leq$ age $<$ 100         & THEN success prob. is & 0.30 & 390  & 919  \\
        & AND housing=no                &                       &      &      &      \\
ELSE    &                               & success prob. is      & 0.09 & 3921 & 38474
\end{tabular}
\centering
\caption{Falling rule list for bank-full dataset, created using Algorithm FRL with $w = 3$}
\end{table}

Running Algorithm FRL with $w = 5$ on the bank-full dataset produces the following falling rule list:

\begin{table}[H]
\centering
\begin{tabular}{llllll}
 & antecedent & & probability  & positive & negative \\
 &            & &              & support  & support  \\\hline
IF      & poutcome=success              & THEN success prob. is & 0.65 & 978  & 531  \\
        & AND default=no                &                       &      &      &      \\
ELSE IF & 60 $\leq$ age $<$ 100         & THEN success prob. is & 0.29 & 426  & 1030 \\
        & AND loan=no                   &                       &      &      &      \\
ELSE IF & 17 $\leq$ age $<$ 30          & THEN success prob. is & 0.25 & 504  & 1539 \\
        & AND housing=no                &                       &      &      &      \\
ELSE IF & previous $\geq$ 2             & THEN success prob. is & 0.23 & 242  & 796  \\
        & AND housing=no                &                       &      &      &      \\
ELSE    &                               & success prob. is      & 0.08 & 3139 & 36026
\end{tabular}
\centering
\caption{Falling rule list for bank-full dataset, created using Algorithm FRL with $w = 5$}
\end{table}

Running Algorithm FRL with $w = 7$ on the bank-full dataset produces the following falling rule list:

\begin{table}[H]
\centering
\begin{tabular}{llllll}
 & antecedent & & probability  & positive & negative \\
 &            & &              & support  & support  \\\hline
IF      & poutcome=success              & THEN success prob. is & 0.65 & 978  & 531  \\
        & AND default=no                &                       &      &      &      \\
ELSE IF & 60 $\leq$ age $<$ 100         & THEN success prob. is & 0.28 & 434  & 1113 \\
        & AND default=no                &                       &      &      &      \\
ELSE IF & 17 $\leq$ age $<$ 30          & THEN success prob. is & 0.25 & 504  & 1539 \\
        & AND housing=no                &                       &      &      &      \\
ELSE IF & previous $\geq$ 2             & THEN success prob. is & 0.23 & 242  & 794  \\
        & AND housing=no                &                       &      &      &      \\
ELSE IF & campaign=1                    & THEN success prob. is & 0.14 & 658  & 4092 \\
        & AND housing=no                &                       &      &      &      \\
ELSE IF & previous $\geq$ 2 AND         & THEN success prob. is & 0.13 & 108  & 707  \\
        & education=tertiary            &                       &      &      &      \\
ELSE    &                               & success prob. is      & 0.07 & 2365 & 31146
\end{tabular}
\centering
\caption{Falling rule list for bank-full dataset, created using Algorithm FRL with $w = 7$}
\end{table}

As the positive class weight $w$ increases, the falling rule list created using Algorithm FRL tends to have rules whose probability estimates are smaller. This is not surprising -- a larger value of $w$ means a smaller threshold $\tau = 1/(1+w)$, and by including rules whose probability estimates are not much larger than the threshold, the falling rule list produced by the algorithm will more likely predict positive, thereby reducing the (weighted) empirical risk of misclassification. Note that Algorithm FRL will never include rules whose probability estimates are less than the threshold (see Corollary 4.5).

\subsection{Effect of Varying $w$ on Algorithm softFRL}

Running Algorithm softFRL with $w = 1$ on the bank-full dataset produces the following softly falling rule list:

\begin{table}[H]
\centering
\begin{tabular}{lllllll}
 & antecedent & & probability  & positive   & positive & negative \\
 &            & &              & proportion & support  & support  \\\hline
IF      & poutcome=success              & THEN prob. is & 0.67 & 0.67 & 557  & 280  \\
        & AND campaign=1                &               &      &      &      &      \\
ELSE IF & poutcome=success              & THEN prob. is & 0.65 & 0.65 & 263  & 143  \\
        & AND marital=married           &               &      &      &      &      \\
ELSE IF & poutcome=success              & THEN prob. is & 0.61 & 0.61 & 154  & 98    \\
        & AND loan=no                   &               &      &      &      &      \\
ELSE    &                               & prob. is      & 0.10 & 0.10 & 4315 & 39401
\end{tabular}
\centering
\caption{Softly falling rule list for bank-full dataset, created using Algorithm softFRL with $w = 1$}
\end{table}

Note that there is an extra column ``positive proportion'' in a table showing a softly falling rule list. This column gives the empirical positive proportion of each antecedent in the softly falling rule list. When the probability estimate of a rule is less than the positive proportion of the antecedent in the same rule, we know that the softly falling rule list has been transformed from a non-falling compatible rule list, and that the monotonicity penalty has been incurred in the process of running Algorithm softFRL.

Running Algorithm softFRL with $w = 3$ on the bank-full dataset produces the following softly falling rule list:

\begin{table}[H]
\centering
\begin{tabular}{lllllll}
 & antecedent & & probability  & positive   & positive & negative \\
 &            & &              & proportion & support  & support  \\\hline
IF      & poutcome=success              & THEN prob. is & 0.65 & 0.65 & 547  & 289  \\
        & AND marital=married           &               &      &      &      &      \\
ELSE IF & poutcome=success              & THEN prob. is & 0.65 & 0.65 & 418  & 225  \\
        & AND loan=no                   &               &      &      &      &      \\
ELSE IF & poutcome=success              & THEN prob. is & 0.56 & 0.56 & 9    & 7    \\
        & AND campaign=1                &               &      &      &      &      \\
ELSE IF & poutcome=success              & THEN prob. is & 0.33 & 0.33 & 4    & 8    \\
        & AND previous $\geq$ 2         &               &      &      &      &      \\
ELSE IF & 60 $\leq$ age $<$ 100         & THEN prob. is & 0.30 & 0.30 & 390  & 919  \\
        & AND housing=no                &               &      &      &      &      \\
ELSE IF & previous $\geq$ 2             & THEN prob. is & 0.15 & 0.15 & 281  & 1559 \\
        & AND campaign=1                &               &      &      &      &      \\
ELSE    &                               & prob. is      & 0.09 & 0.09 & 3640 & 36915
\end{tabular}
\centering
\caption{Softly falling rule list for bank-full dataset, created using Algorithm softFRL with $w = 3$}
\end{table}

Running Algorithm softFRL with $w = 5$ on the bank-full dataset produces the following softly falling rule list:

\begin{table}[H]
\centering
\begin{tabular}{lllllll}
 & antecedent & & probability  & positive   & positive & negative \\
 &            & &              & proportion & support  & support  \\\hline
IF      & poutcome=success              & THEN prob. is & 0.65 & 0.65 & 978  & 533   \\
ELSE IF & 60 $\leq$ age $<$ 100         & THEN prob. is & 0.29 & 0.29 & 426  & 1030  \\
        & AND loan=no                   &               &      &      &      &       \\
ELSE IF & poutcome=unknown              & THEN prob. is & 0.11 & 0.11 & 2380 & 18659 \\
        & AND contact=cellular          &               &      &      &      &       \\
ELSE    &                               & prob. is      & 0.07 & 0.07 & 1505 & 19700
\end{tabular}
\centering
\caption{Softly falling rule list for bank-full dataset, created using Algorithm softFRL with $w = 5$}
\end{table}

Running Algorithm softFRL with $w = 7$ on the bank-full dataset produces the following softly falling rule list:

\begin{table}[H]
\centering
\begin{tabular}{lllllll}
 & antecedent & & probability  & positive   & positive & negative \\
 &            & &              & proportion & support  & support  \\\hline
IF      & poutcome=success              & THEN prob. is & 0.65 & 0.65 & 978  & 533   \\
ELSE IF & 60 $\leq$ age $<$ 100         & THEN prob. is & 0.28 & 0.28 & 435  & 1120  \\
ELSE IF & marital=single                & THEN prob. is & 0.18 & 0.18 & 970  & 4504  \\
        & AND housing=no                &               &      &      &      &       \\
ELSE IF & contact=cellular              & THEN prob. is & 0.10 & 0.10 & 2255 & 19970 \\
        & AND default=no                &               &      &      &      &       \\
ELSE    &                               & prob. is      & 0.05 & 0.05 & 651  & 13795
\end{tabular}
\centering
\caption{Softly falling rule list for bank-full dataset, created using Algorithm softFRL with $w = 7$}
\end{table}

As the positive class weight $w$ increases, the softly falling rule list created using Algorithm softFRL also tends to have rules whose probability estimates are smaller. This is again not surprising -- a larger value of $w$ means a smaller threshold $\tau = 1/(1+w)$, and by including rules whose probability estimates are not much larger than the threshold, the softly falling rule list produced by the algorithm will more likely predict positive, thereby reducing the (weighted) empirical risk of misclassification.

\subsection{Effect of Varying $C$ on Algorithm FRL}

Running Algorithm FRL with $C = 0.000001$ on the bank-full dataset produces the following falling rule list:

\begin{table}[H]
\centering
\begin{tabular}{llllll}
 & antecedent & & probability  & positive & negative \\
 &            & &              & support  & support  \\\hline
IF      & poutcome=success              & THEN success prob. is & 0.65 & 978  & 531  \\
        & AND default=no                &                       &      &      &      \\
ELSE IF & 60 $\leq$ age $<$ 100         & THEN success prob. is & 0.28 & 434  & 1113 \\
        & AND default=no                &                       &      &      &      \\
ELSE IF & 17 $\leq$ age $<$ 30          & THEN success prob. is & 0.25 & 504  & 1539 \\
        & AND housing=no                &                       &      &      &      \\
ELSE IF & previous $\geq$ 2             & THEN success prob. is & 0.23 & 242  & 794  \\
        & AND housing=no                &                       &      &      &      \\
ELSE IF & campaign=1                    & THEN success prob. is & 0.14 & 658  & 4092 \\
        & AND housing=no                &                       &      &      &      \\
ELSE IF & previous $\geq$ 2 AND         & THEN success prob. is & 0.13 & 108  & 707  \\
        & education=tertiary            &                       &      &      &      \\
ELSE    &                               & success prob. is      & 0.07 & 2365 & 31146
\end{tabular}
\centering
\caption{Falling rule list for bank-full dataset, created using Algorithm FRL with $C = 0.000001$}
\end{table}

Running Algorithm FRL with $C = 0.01$ on the bank-full dataset produces the following falling rule list:

\begin{table}[H]
\centering
\begin{tabular}{llllll}
 & antecedent & & probability  & positive & negative \\
 &            & &              & support  & support  \\\hline
IF      & poutcome=success              & THEN success prob. is & 0.65 & 978  & 531  \\
        & AND default=no                &                       &      &      &      \\
ELSE IF & 60 $\leq$ age $<$ 100         & THEN success prob. is & 0.29 & 426  & 1030 \\
        & AND loan=no                   &                       &      &      &      \\
ELSE IF & 17 $\leq$ age $<$ 30          & THEN success prob. is & 0.20 & 653  & 2621 \\
        & AND contact=cellular          &                       &      &      &      \\
ELSE IF & campaign=1                    & THEN success prob. is & 0.15 & 803  & 4634 \\
        & AND housing=no                &                       &      &      &      \\
ELSE    &                               & success prob. is      & 0.07 & 2429 & 31106
\end{tabular}
\centering
\caption{Falling rule list for bank-full dataset, created using Algorithm FRL with $C = 0.01$}
\end{table}

Running Algorithm FRL with $C = 0.1$ on the bank-full dataset produces the following falling rule list:

\begin{table}[H]
\centering
\begin{tabular}{llllll}
 & antecedent & & probability  & positive & negative \\
 &            & &              & support  & support  \\\hline
IF      & housing=no                    & THEN success prob. is & 0.20 & 2883 & 11799 \\
        & AND contact=cellular          &                       &      &      &      \\
ELSE    &                               & success prob. is      & 0.08 & 2406 & 28123
\end{tabular}
\centering
\caption{Falling rule list for bank-full dataset, created using Algorithm FRL with $C = 0.1$}
\end{table}

As the cost $C$ of adding a rule increases, the size of the falling rule list created by Algorithm FRL decreases, as expected.

\subsection{Effect of Varying $C$ on Algorithm softFRL}

Running Algorithm softFRL with $C = 0.000001$ on the bank-full dataset produces the following softly falling rule list:

\begin{table}[H]
\centering
\begin{tabular}{lllllll}
 & antecedent & & probability  & positive   & positive & negative \\
 &            & &              & proportion & support  & support  \\\hline
IF      & poutcome=success              & THEN prob. is & 0.65 & 0.65 & 978  & 533   \\
ELSE IF & 60 $\leq$ age $<$ 100         & THEN prob. is & 0.28 & 0.28 & 435  & 1120  \\
ELSE IF & marital=single                & THEN prob. is & 0.18 & 0.18 & 970  & 4504  \\
        & AND housing=no                &               &      &      &      &       \\
ELSE IF & contact=cellular              & THEN prob. is & 0.10 & 0.10 & 2255 & 19970 \\
        & AND default=no                &               &      &      &      &       \\
ELSE    &                               & prob. is      & 0.05 & 0.05 & 651  & 13795
\end{tabular}
\centering
\caption{Softly falling rule list for bank-full dataset, created using Algorithm softFRL with $C = 0.000001$}
\end{table}

Running Algorithm softFRL with $C = 0.01$ on the bank-full dataset produces the following softly falling rule list:

\begin{table}[H]
\centering
\begin{tabular}{lllllll}
 & antecedent & & probability  & positive   & positive & negative \\
 &            & &              & proportion & support  & support  \\\hline
IF      & poutcome=success              & THEN prob. is & 0.65 & 0.65 & 934  & 495   \\
        & AND loan=no                   &               &      &      &      &       \\
ELSE IF & housing=no                    & THEN prob. is & 0.16 & 0.16 & 2245 & 11535 \\
        & AND contact=cellular          &               &      &      &      &       \\
ELSE IF & housing=yes                   & THEN prob. is & 0.07 & 0.07 & 1677 & 22591 \\
        & AND default=no                &               &      &      &      &       \\
ELSE    &                               & prob. is      & 0.07 & 0.08 & 433  & 5301
\end{tabular}
\centering
\caption{Softly falling rule list for bank-full dataset, created using Algorithm softFRL with $C = 0.01$}
\end{table}

Running Algorithm softFRL with $C = 0.1$ on the bank-full dataset produces the following softly falling rule list:

\begin{table}[H]
\centering
\begin{tabular}{lllllll}
 & antecedent & & probability  & positive   & positive & negative \\
 &            & &              & proportion & support  & support  \\\hline
IF      & housing=no                    & THEN prob. is & 0.20 & 0.20 & 2883 & 11799 \\
        & AND contact=cellular          &                       &      &      &      &       \\
ELSE    &                               & prob. is      & 0.08 & 0.08 & 2406 & 28123
\end{tabular}
\centering
\caption{Softly falling rule list for bank-full dataset, created using Algorithm softFRL with $C = 0.1$}
\end{table}

As the cost $C$ of adding a rule increases, the size of the softly falling rule list created by Algorithm softFRL decreases, as expected.

\subsection{Effect of Varying $C_1$ on Algorithm softFRL}


Running Algorithm softFRL with $C_1 \in \{0.005, 0.05, 0.5\}$ on the bank-full dataset produces the softly falling rule lists shown in Tables \ref{soft_frl_c1_0.005}, \ref{soft_frl_c1_0.05}, and \ref{soft_frl_c1_0.5}.

\begin{table}[htbp]
\centering
\begin{tabular}{lllllll}
 & antecedent & & probability  & positive   & positive & negative \\
 &            & &              & proportion & support  & support  \\\hline
IF      & poutcome=success              & THEN prob. is & 0.65 & 0.65 & 978  & 533   \\
ELSE IF & 60 $\leq$ age $<$ 100         & THEN prob. is & 0.30 & 0.30 & 599  & 1177  \\
        & AND housing=no                &               &      &      &      &       \\
ELSE IF & marital=single                & THEN prob. is & 0.18 & 0.18 & 970  & 4504  \\
        & AND housing=no                &               &      &      &      &       \\
ELSE IF & marital=single                & THEN prob. is & 0.08 & 0.08 & 456  & 4936  \\
        & AND previous=0                &               &      &      &      &       \\
ELSE IF & campaign $\geq$ 3             & THEN prob. is & 0.06 & 0.06 & 323  & 5294  \\
        & AND education=secondary       &               &      &      &      &       \\
ELSE IF & 30 $\leq$ age < 40            & THEN prob. is & 0.06 & 0.08 & 568  & 6849  \\
        & AND previous=0                &               &      &      &      &       \\
ELSE IF & education=tertiary            & THEN prob. is & 0.06 & 0.14 & 361  & 2237  \\
        & AND housing=no                &               &      &      &      &       \\
ELSE IF & loan=yes                      & THEN prob. is & 0.05 & 0.05 & 106  & 1972  \\
        & AND previous=0                &               &      &      &      &       \\
ELSE IF & education=secondary           & THEN prob. is & 0.05 & 0.09 & 595  & 5779  \\
        & AND default=no                &               &      &      &      &       \\
ELSE IF & campaign=1                    & THEN prob. is & 0.05 & 0.08 & 233  & 2564  \\
ELSE IF & housing=no                    & THEN prob. is & 0.05 & 0.05 & 68   & 1176  \\
        & AND previous=0                &               &      &      &      &       \\
ELSE IF & job=management                & THEN prob. is & 0.05 & 0.10 & 75   & 693   \\
        & AND contact=cellular          &               &      &      &      &       \\
ELSE IF & job=technician                & THEN prob. is & 0.05 & 0.07 & 10   & 143   \\
        & AND poutcome=unknown          &               &      &      &      &       \\
ELSE IF & marital=married               & THEN prob. is & 0.05 & 0.06 & 110  & 1841  \\
ELSE IF & campaign $\geq$ 3             & THEN prob. is & 0.05 & 0.06 & 16   & 238   \\
        & AND housing=yes               &               &      &      &      &       \\
ELSE IF & marital=single                & THEN prob. is & 0.05 & 0.13 & 13   & 91    \\
        & AND housing=yes               &               &      &      &      &       \\
ELSE IF & housing=yes                   & THEN prob. is & 0.05 & 0.10 & 8    & 69    \\
        & AND contact=cellular          &               &      &      &      &       \\
ELSE IF & job=blue-collar               & THEN prob. is & 0.05 & 0.16 & 4    & 21    \\
        & AND loan=no                   &               &      &      &      &       \\
ELSE    &                               & prob. is      & 0.05 & 0.07 & 5    & 63
\end{tabular}
\centering
\caption{Softly falling rule list for bank-full dataset, created using Algorithm softFRL with $C_1 = 0.005$}\label{soft_frl_c1_0.005}
\end{table}

\begin{table}[htbp]
\centering
\begin{tabular}{lllllll}
 & antecedent & & probability  & positive   & positive & negative \\
 &            & &              & proportion & support  & support  \\\hline
IF      & poutcome=success              & THEN prob. is & 0.65 & 0.65 & 978  & 531   \\
        & AND default=no                &               &      &      &      &       \\
ELSE IF & housing=yes                   & THEN prob. is & 0.07 & 0.07 & 1686 & 22974 \\
ELSE IF & 50 $\leq$ age $<$ 60          & THEN prob. is & 0.07 & 0.09 & 367  & 3806  \\
        & AND poutcome=unknown          &               &      &      &      &       \\
ELSE IF & contact=cellular              & THEN prob. is & 0.07 & 0.18 & 1927 & 8961  \\
        & AND default=no                &               &      &      &      &       \\
ELSE IF & campaign=1                    & THEN prob. is & 0.07 & 0.08 & 126  & 1374  \\
        & AND poutcome=unknown          &               &      &      &      &       \\
ELSE IF & campaign $\geq$ 3             & THEN prob. is & 0.07 & 0.08 & 93   & 1110  \\
        & AND loan=no                   &               &      &      &      &       \\
ELSE IF & campaign=2                    & THEN prob. is & 0.07 & 0.09 & 18   & 192   \\
        & AND education=tertiary        &               &      &      &      &       \\        
ELSE IF & loan=no                       & THEN prob. is & 0.07 & 0.10 & 72   & 648   \\
ELSE    &                               & prob. is      & 0.06 & 0.06 & 22   & 326
\end{tabular}
\centering
\caption{Softly falling rule list for bank-full dataset, created using Algorithm softFRL with $C_1 = 0.05$}\label{soft_frl_c1_0.05}
\end{table}

\begin{table}[h]
\centering
\begin{tabular}{lllllll}
 & antecedent & & probability  & positive   & positive & negative \\
 &            & &              & proportion & support  & support  \\\hline
IF      & poutcome=success              & THEN prob. is & 0.65 & 0.65 & 978  & 533   \\
ELSE IF & 60 $\leq$ age $<$ 100         & THEN prob. is & 0.28 & 0.28 & 435  & 1120  \\
ELSE IF & marital=single                & THEN prob. is & 0.18 & 0.18 & 970  & 4504  \\
        & AND housing=no                &               &      &      &      &       \\
ELSE IF & contact=cellular              & THEN prob. is & 0.10 & 0.10 & 2255 & 19970 \\
        & AND default=no                &               &      &      &      &       \\
ELSE    &                               & prob. is      & 0.05 & 0.05 & 651  & 13795
\end{tabular}
\centering
\caption{Softly falling rule list for bank-full dataset, created using Algorithm softFRL with $C_1 = 0.5$}\label{soft_frl_c1_0.5}
\end{table}

When the monotonicity penalty $C_1$ is small, the softly falling rule list created by Algorithm softFRL exhibits the ``pulling down'' of the empirical positive proportion for a substantial number of rules, because with little monotonicity penalty the algorithm will more likely choose a rule list that frequently violates monotonicity but that has a small empirical risk on the training set, in the hope of getting more of the training instances ``right''. This is also why the softly falling rule list tends to be longer when $C_1$ is small: in minimizing the empirical risk on the training set with little regularization (the default $C = 0.000001$ is very small), the algorithm tends to overfit the training data.

When $C_1$ becomes larger, the softly falling rule list created by Algorithm softFRL exhibits less ``pulling down'' of the empirical positive proportion. This is consistent with our expectation that when $C_1$ is larger, the penalty for violating monotonicity is higher and the algorithm will less likely choose a rule list that frequently violates monotonicity.

\section{Additional Experiments Comparing Algorithm FRL and Algorithm softFRL to Other Classification Algorithms}

\begin{figure}[h!]
    \centering
    \begin{subfigure}[b]{0.45\textwidth}
        \includegraphics[clip, width=\textwidth]{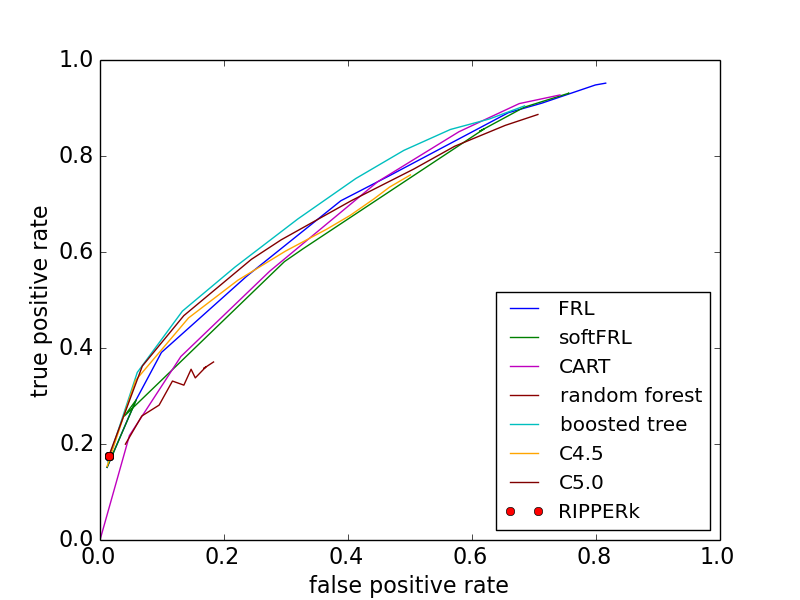}
        \caption{ROC curves on the test set using different $w$ values for the first additional training-test split}
        \label{fig:roc_split1}
    \end{subfigure}
    ~
    \centering
    \begin{subfigure}[b]{0.45\textwidth}
        \includegraphics[clip, width=\textwidth]{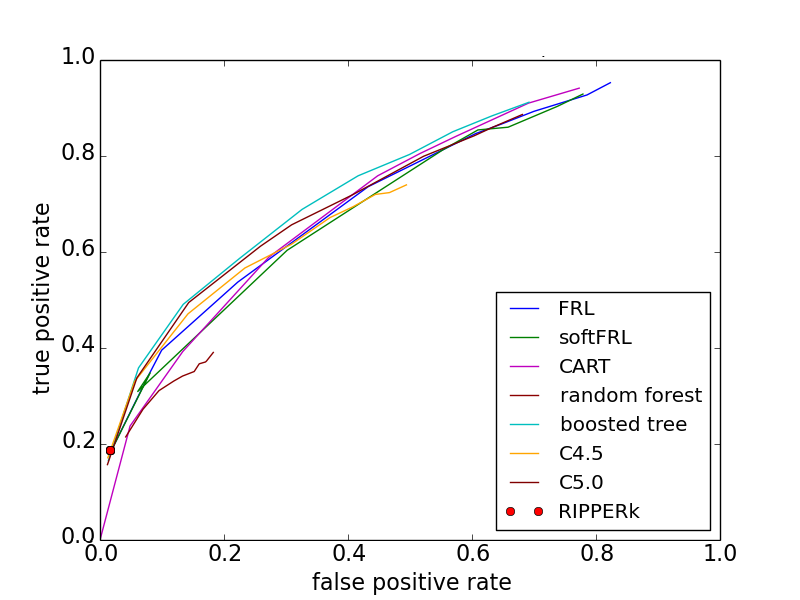}
        \caption{ROC curves on the test set using different $w$ values for the second additional training-test split}
        \label{fig:roc_split2}
    \end{subfigure}
    ~
    \centering
    \begin{subfigure}[b]{0.45\textwidth}
        \includegraphics[clip, width=\textwidth]{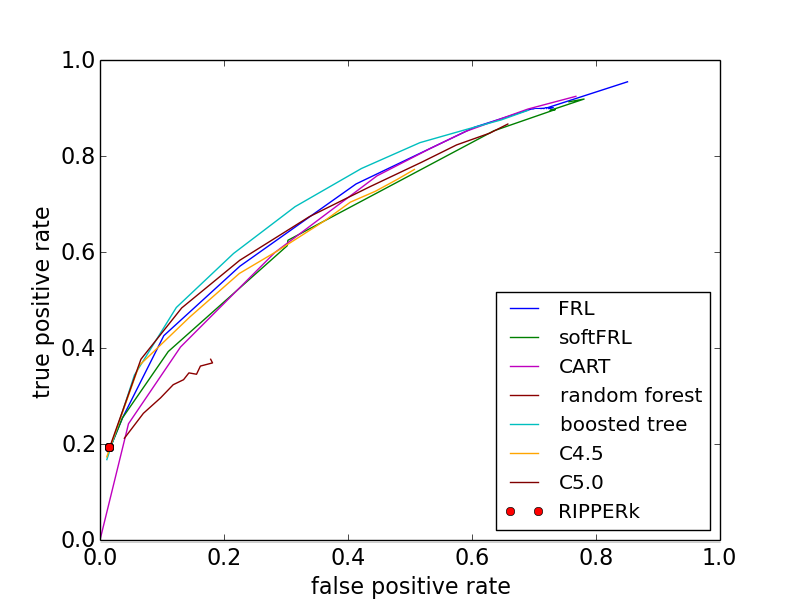}
        \caption{ROC curves on the test set using different $w$ values for the third additional training-test split}
        \label{roc_split3}
    \end{subfigure}
    ~
    \centering
    \begin{subfigure}[b]{0.45\textwidth}
        \includegraphics[clip, width=\textwidth]{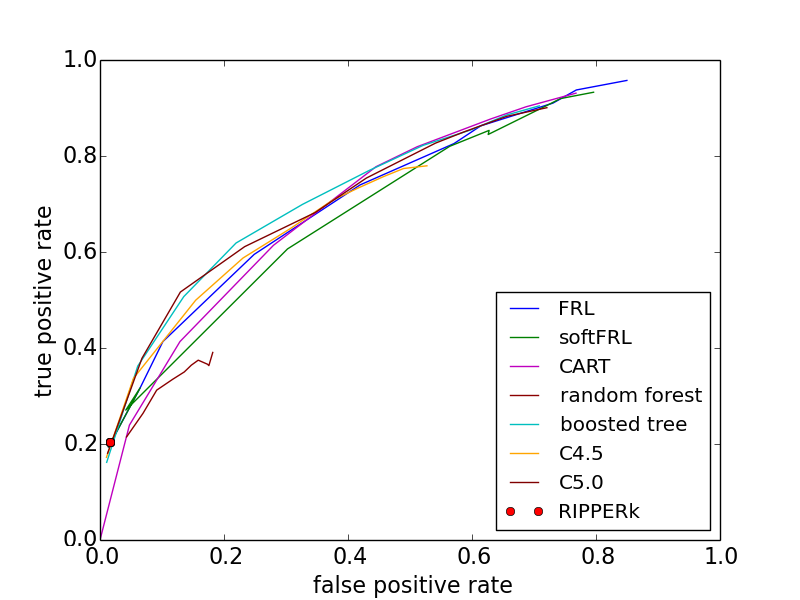}
        \caption{ROC curves on the test set using different $w$ values for the fourth additional training-test split}
        \label{roc_split4}
    \end{subfigure}
    \caption{ROC curves on the test set using different $w$ values for four additional training-test splits}\label{fig:roc_additional_splits}
\end{figure}

Figure \ref{fig:roc_additional_splits} shows the ROC curves on the test set using different values of $w$, for four additional training-test splits. As we can see, the curves in Figure \ref{fig:roc_additional_splits} lie close to each other, again demonstrating the effectiveness of our algorithms in producing falling rule lists that, when used as classifiers, are comparable with classifiers produced by other widely used classification algorithms, in a cost-sensitive setting.

\section{Additional Experiments Comparing Bayesian Approach to Our Optimization Approach}


We conducted a set of experiments comparing the Bayesian approach to our optimization approach. We trained falling rule lists on the entire bank-full dataset using both the Bayesian approach and our optimization approach (Algorithm FRL), and plotted the weighted training loss over real runtime. In particular, for each positive class weight $w \in \{1, 3, 5, 7\}$, we set the threshold to $1/(1+w)$ (By Theorem 2.8, this is the threshold with the least weighted training loss for any given rule list), and computed the weighted training loss using this threshold. For the Bayesian approach, we recorded the runtime and computed the weighted training loss for every $100$ iterations of Markov chain Monte-Carlo sampling with simulated annealing, up to $6000$ iterations. For our optimization approach, we ran Algorithm FRL for $3000$ iterations and recorded the runtime and the weighted training loss whenever the algorithm finds a falling rule list with a smaller (regularized) weighted training loss. Since we want to focus our experiments on the efficiency of searching the model space, the runtimes recorded do not include the time for mining the antecedents. Due to the random nature of both approaches, the experiments were repeated several times.

\begin{figure}[h!]
    \centering
    \begin{subfigure}[b]{0.45\textwidth}
        \includegraphics[clip, width=\textwidth]{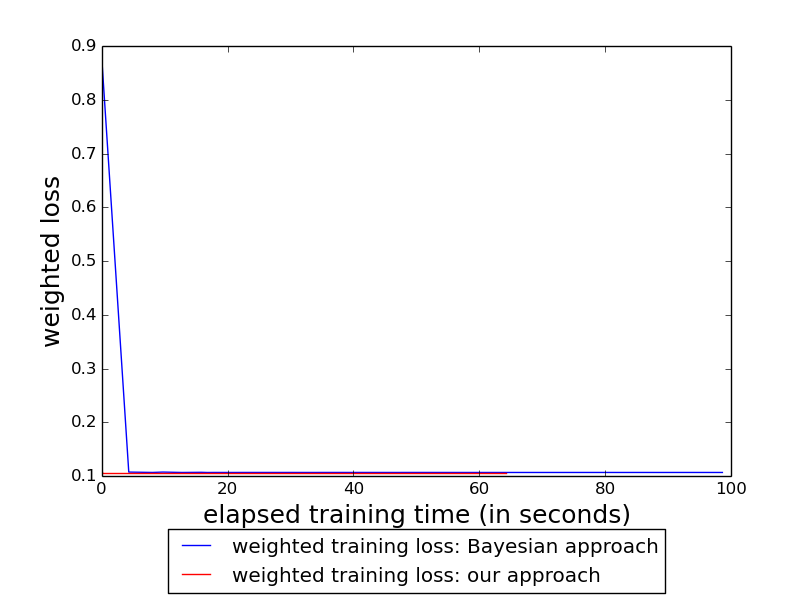}
        \caption{positive class weight $w = 1$}
        \label{fig:comp_w=1_run1}
    \end{subfigure}
    ~
    \centering
    \begin{subfigure}[b]{0.45\textwidth}
        \includegraphics[clip, width=\textwidth]{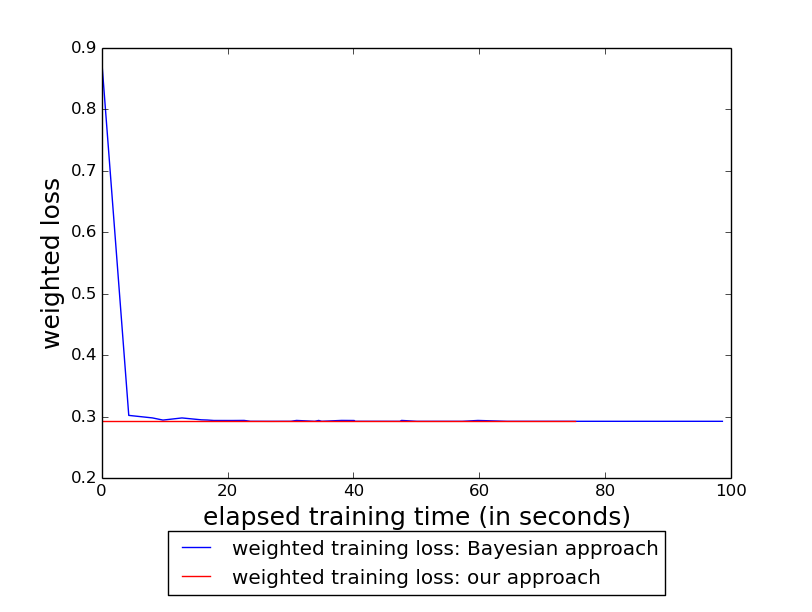}
        \caption{positive class weight $w = 3$}
        \label{fig:comp_w=3_run1}
    \end{subfigure}
    ~
    \centering
    \begin{subfigure}[b]{0.45\textwidth}
        \includegraphics[clip, width=\textwidth]{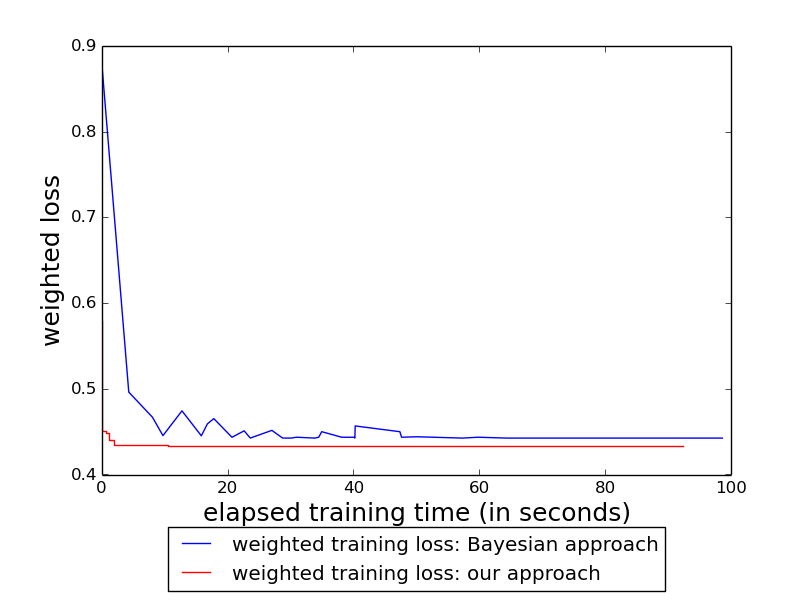}
        \caption{positive class weight $w = 5$}
        \label{fig:comp_w=5_run1}
    \end{subfigure}
    ~
    \centering
    \begin{subfigure}[b]{0.45\textwidth}
        \includegraphics[clip, width=\textwidth]{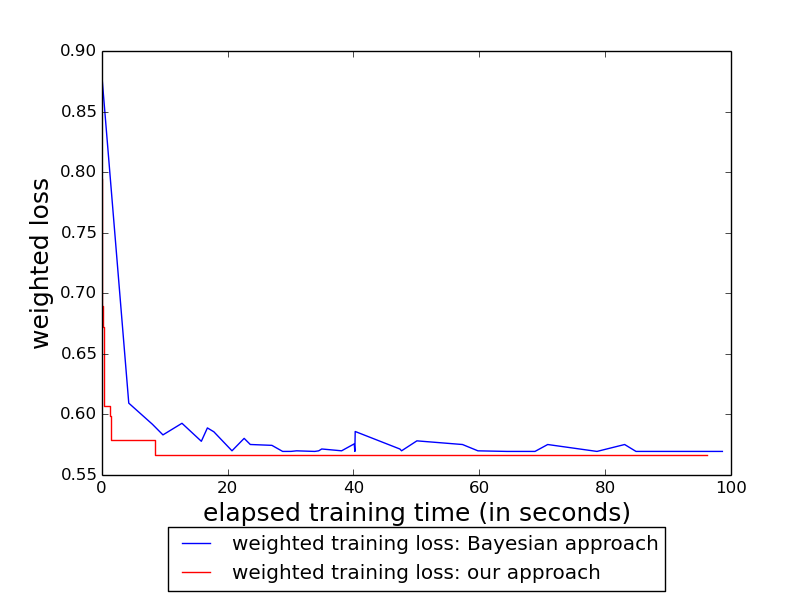}
        \caption{positive class weight $w = 7$}
        \label{fig:comp_w=7_run1}
    \end{subfigure}
    \caption{Plots of the weighted training loss over real runtime for the Bayesian approach and our optimization approach (Algorithm FRL): first additional run}\label{fig:compare_with_BayesianFRL_run1}
\end{figure}

\begin{figure}[h!]
    \centering
    \begin{subfigure}[b]{0.45\textwidth}
        \includegraphics[clip, width=\textwidth]{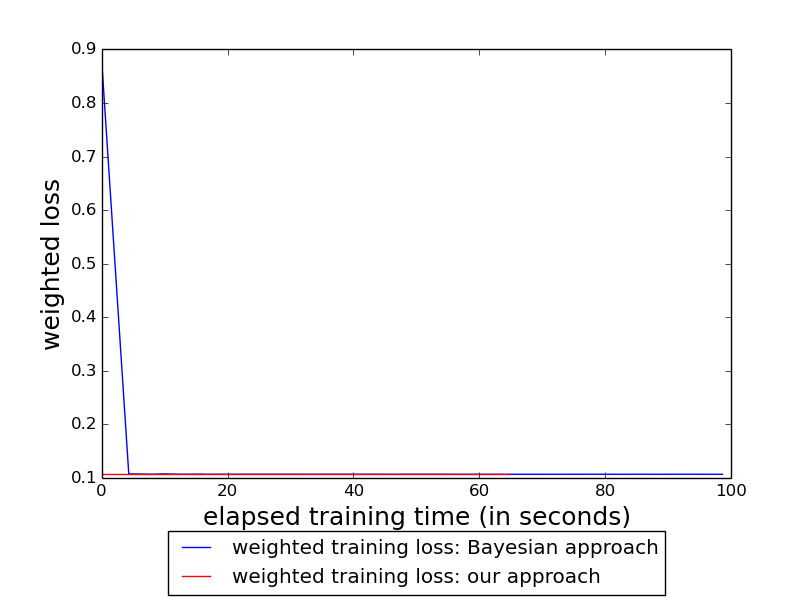}
        \caption{positive class weight $w = 1$}
        \label{fig:comp_w=1_run2}
    \end{subfigure}
    ~
    \centering
    \begin{subfigure}[b]{0.45\textwidth}
        \includegraphics[clip, width=\textwidth]{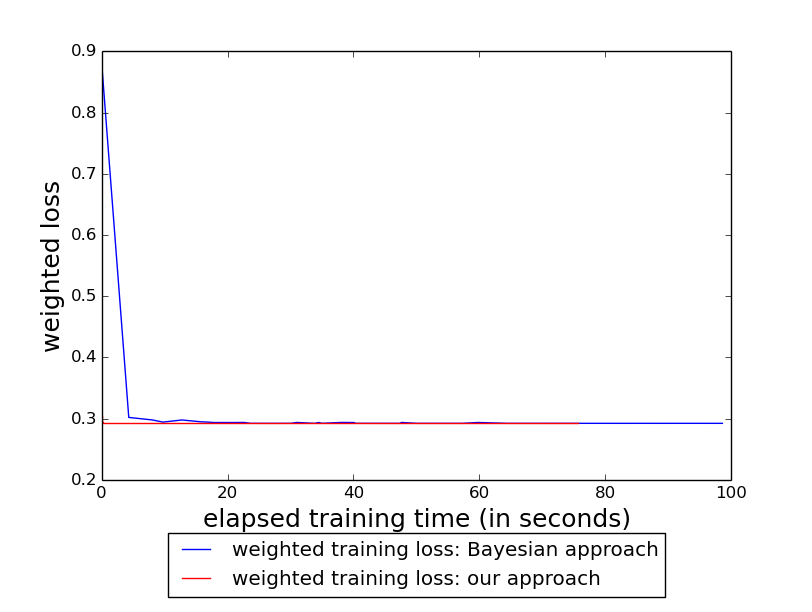}
        \caption{positive class weight $w = 3$}
        \label{fig:comp_w=3_run2}
    \end{subfigure}
    ~
    \centering
    \begin{subfigure}[b]{0.45\textwidth}
        \includegraphics[clip, width=\textwidth]{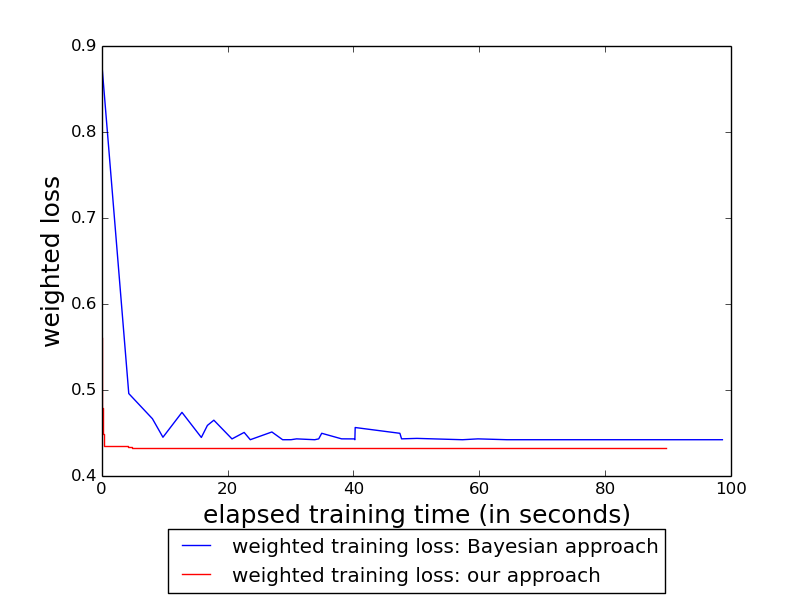}
        \caption{positive class weight $w = 5$}
        \label{fig:comp_w=5_run2}
    \end{subfigure}
    ~
    \centering
    \begin{subfigure}[b]{0.45\textwidth}
        \includegraphics[clip, width=\textwidth]{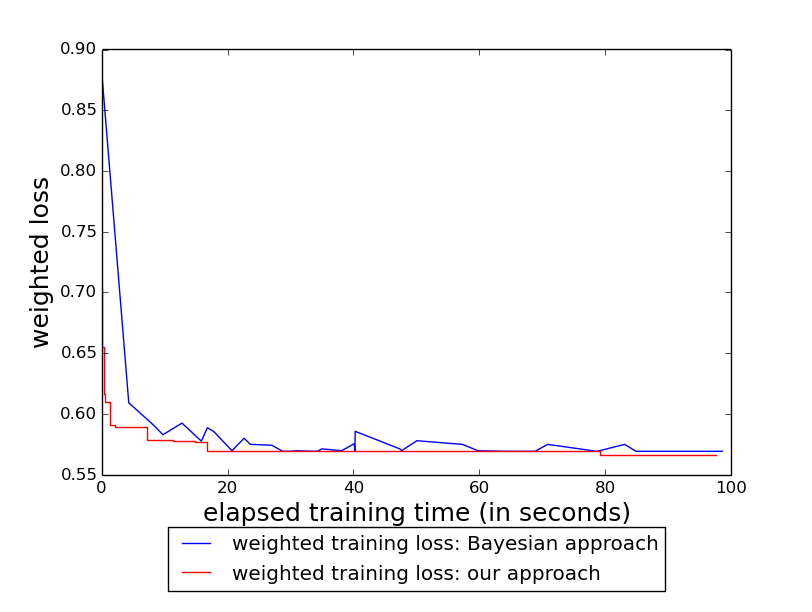}
        \caption{positive class weight $w = 7$}
        \label{fig:comp_w=7_run2}
    \end{subfigure}
    \caption{Plots of the weighted training loss over real runtime for the Bayesian approach and our optimization approach (Algorithm FRL): second additional run}\label{fig:compare_with_BayesianFRL_run2}
\end{figure}

\begin{figure}[h!]
    \centering
    \begin{subfigure}[b]{0.45\textwidth}
        \includegraphics[clip, width=\textwidth]{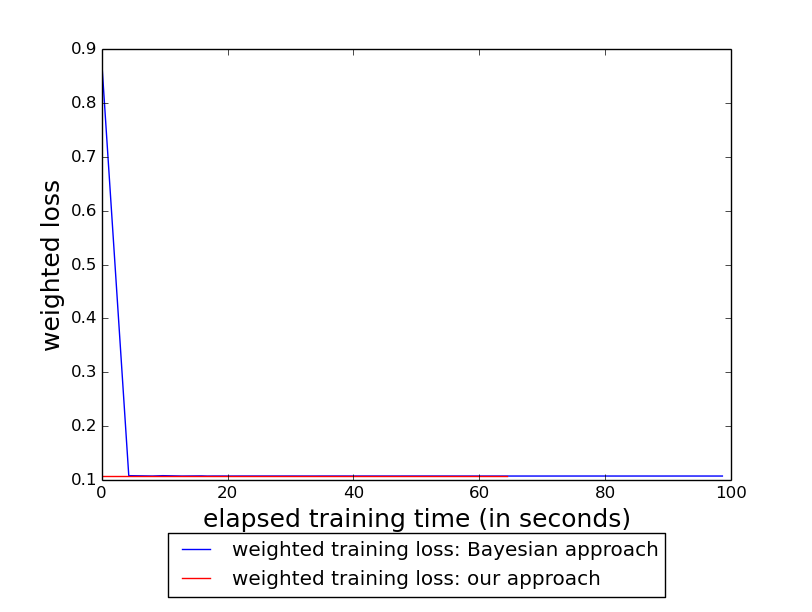}
        \caption{positive class weight $w = 1$}
        \label{fig:comp_w=1_run3}
    \end{subfigure}
    ~
    \centering
    \begin{subfigure}[b]{0.45\textwidth}
        \includegraphics[clip, width=\textwidth]{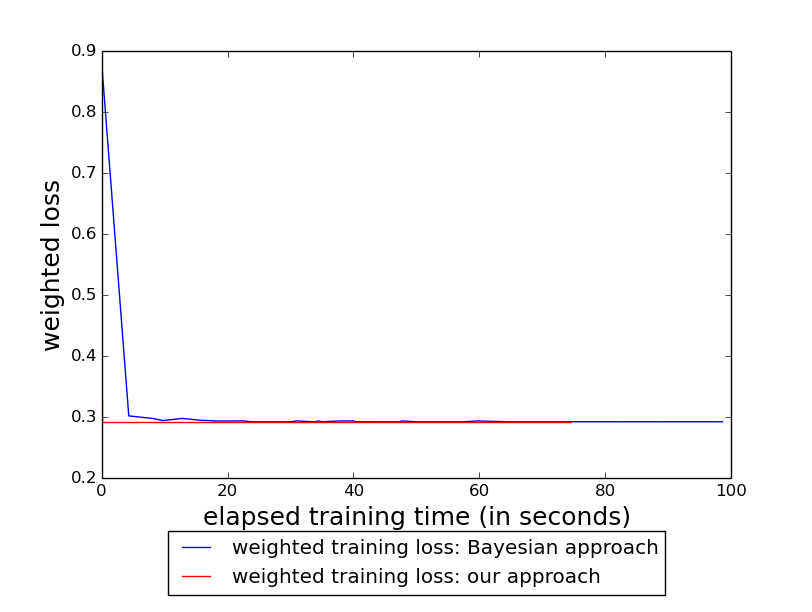}
        \caption{positive class weight $w = 3$}
        \label{fig:comp_w=3_run3}
    \end{subfigure}
    ~
    \centering
    \begin{subfigure}[b]{0.45\textwidth}
        \includegraphics[clip, width=\textwidth]{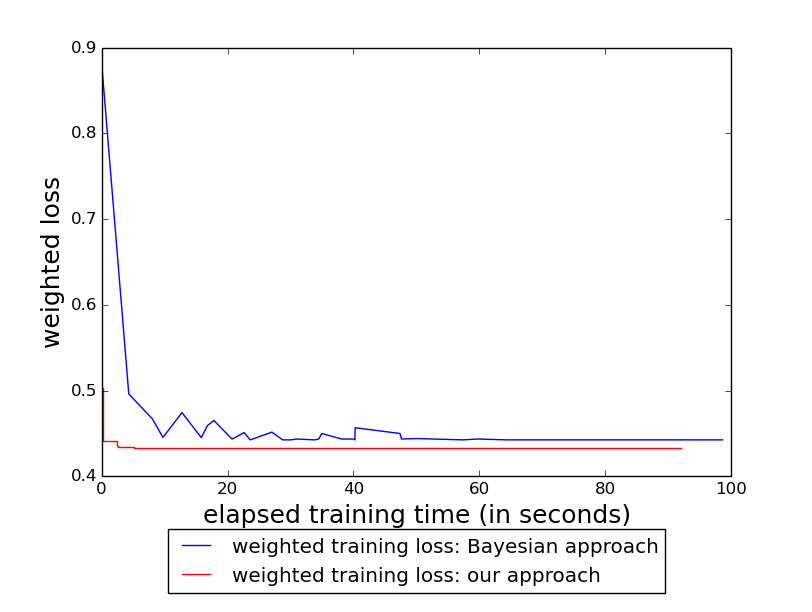}
        \caption{positive class weight $w = 5$}
        \label{fig:comp_w=5_run3}
    \end{subfigure}
    ~
    \centering
    \begin{subfigure}[b]{0.45\textwidth}
        \includegraphics[clip, width=\textwidth]{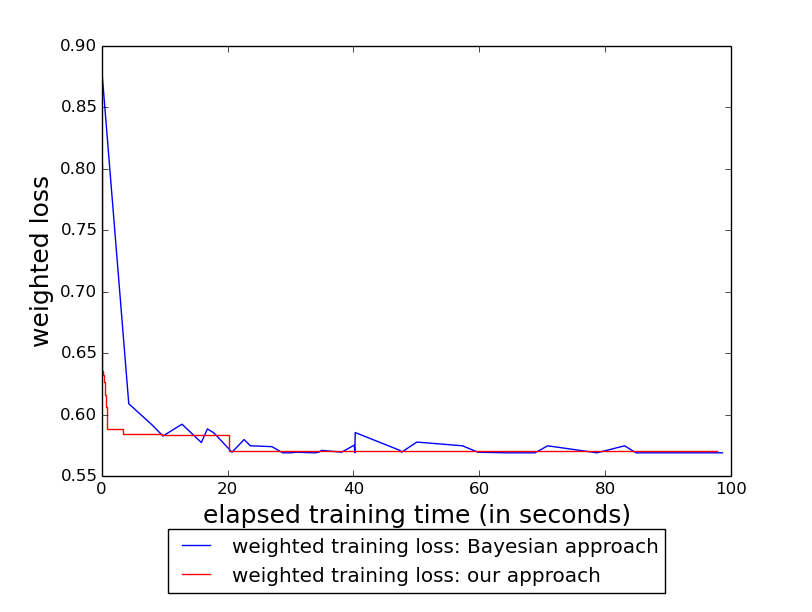}
        \caption{positive class weight $w = 7$}
        \label{fig:comp_w=7_run3}
    \end{subfigure}
    \caption{Plots of the weighted training loss over real runtime for the Bayesian approach and our optimization approach (Algorithm FRL): third additional run}\label{fig:compare_with_BayesianFRL_run3}
\end{figure}

\begin{figure}[h!]
    \centering
    \begin{subfigure}[b]{0.45\textwidth}
        \includegraphics[clip, width=\textwidth]{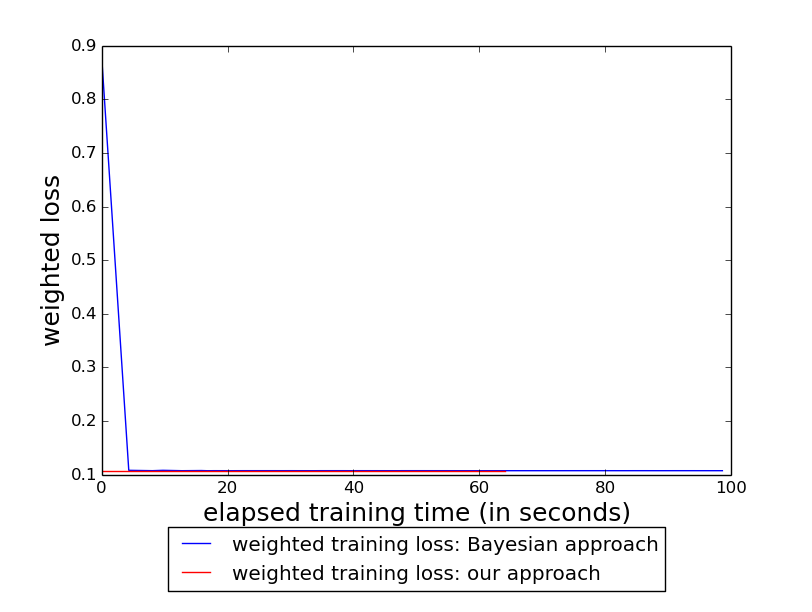}
        \caption{positive class weight $w = 1$}
        \label{fig:comp_w=1_run4}
    \end{subfigure}
    ~
    \centering
    \begin{subfigure}[b]{0.45\textwidth}
        \includegraphics[clip, width=\textwidth]{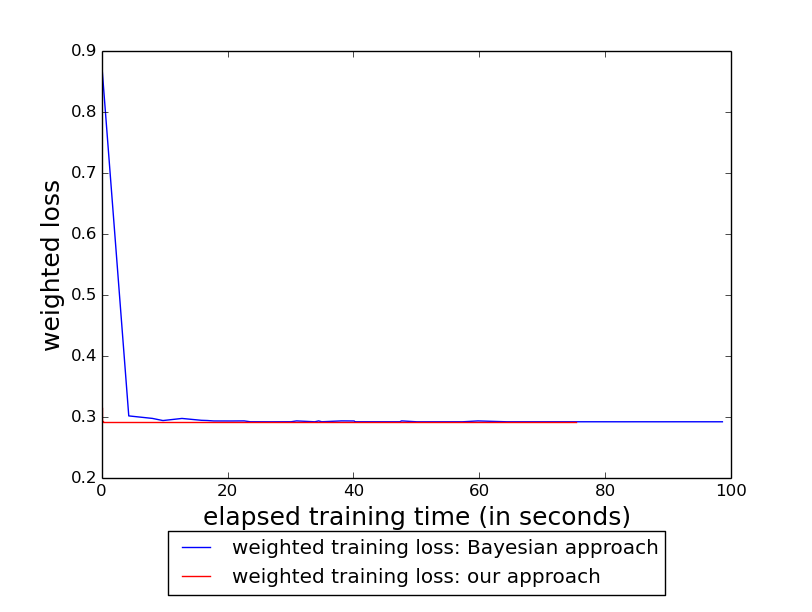}
        \caption{positive class weight $w = 3$}
        \label{fig:comp_w=3_run4}
    \end{subfigure}
    ~
    \centering
    \begin{subfigure}[b]{0.45\textwidth}
        \includegraphics[clip, width=\textwidth]{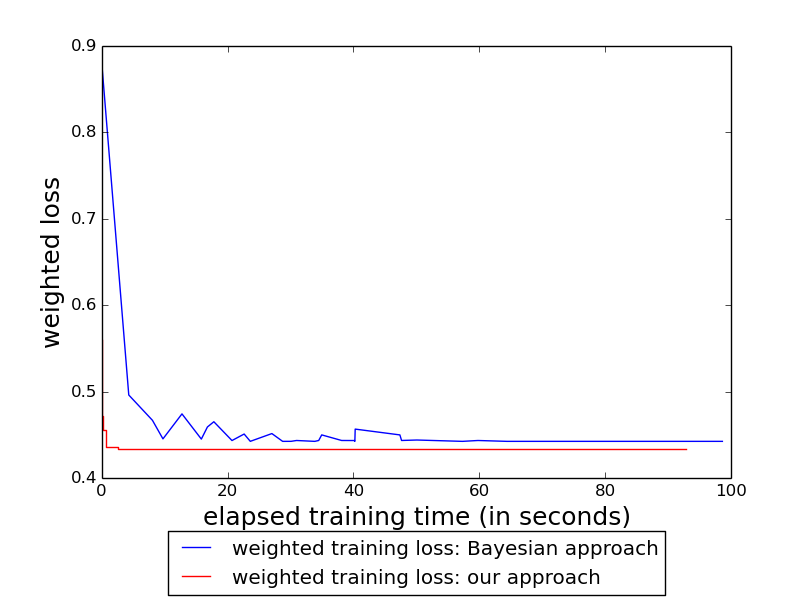}
        \caption{positive class weight $w = 5$}
        \label{fig:comp_w=5_run4}
    \end{subfigure}
    ~
    \centering
    \begin{subfigure}[b]{0.45\textwidth}
        \includegraphics[clip, width=\textwidth]{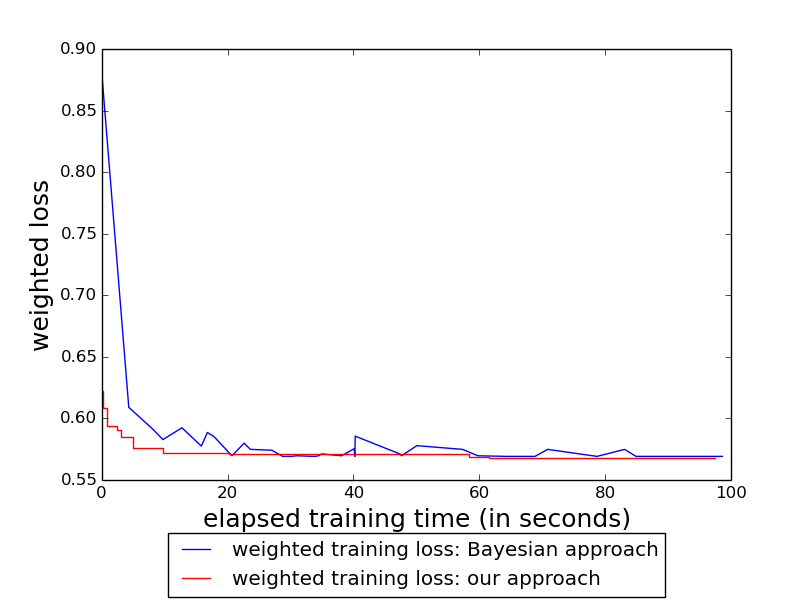}
        \caption{positive class weight $w = 7$}
        \label{fig:comp_w=7_run4}
    \end{subfigure}
    \caption{Plots of the weighted training loss over real runtime for the Bayesian approach and our optimization approach (Algorithm FRL): fourth additional run}\label{fig:compare_with_BayesianFRL_run4}
\end{figure}

Figures \ref{fig:compare_with_BayesianFRL_run1} to \ref{fig:compare_with_BayesianFRL_run4} show the plots of the weighted training loss over real runtime for the Bayesian approach and our optimization approach (Algorithm FRL), for four additional runs of the same algorithms. Due to the random nature of both approaches, it is sometimes possible that our approach (Algorithm FRL) may find in $3000$ iterations a falling rule list with a slightly larger weighted training loss, compared to the Bayesian approach with $6000$ iterations (see Figure \ref{fig:comp_w=7_run3}). However, in general, our approach tends to find a falling rule list with a smaller weighted training loss faster, due to aggressive pruning of the search space.

It is worth pointing out that both the Bayesian approach and our optimization approach produce similar falling rule lists. Table \ref{table:bfrl_1} shows a falling rule list for the bank-full dataset, obtained in a particular run of the Bayesian approach with $6000$ iterations. Table \ref{table:frl_1} shows a falling rule list for the same dataset, obtained in a particular run of Algorithm FRL with $3000$ iterations and the positive class weight $w = 7$. As we can see, the top four rules in both falling rule lists are identical. Tables \ref{table:bfrl_2} and \ref{table:frl_2} show another pair of falling rule lists obtained using both approaches in different runs, and in this case, both approaches have identified some common rules for a high chance of marketing success. This means that both the Bayesian approach and our optimization approach tend to identify similar conditions that are significant, but our approach has the added advantage of faster training convergence over the Bayesian approach in general.

\begin{table}[h!]
\centering
\begin{tabular}{llllll}
 & antecedent & & probability  & positive & negative \\
 &            & &              & support  & support  \\\hline
IF      & poutcome=success              & THEN success prob. is & 0.65 & 978  & 531  \\
        & AND default=no                &                       &      &      &      \\
ELSE IF & 60 $\leq$ age $<$ 100         & THEN success prob. is & 0.29 & 426  & 1030 \\
        & AND loan=no                   &                       &      &      &      \\
ELSE IF & 17 $\leq$ age $<$ 30          & THEN success prob. is & 0.25 & 504  & 1539 \\
        & AND housing=no                &                       &      &      &      \\
ELSE IF & campaign=1                    & THEN success prob. is & 0.15 & 787  & 4471 \\
        & AND housing=no                &                       &      &      &      \\
ELSE IF & education=tertiary            & THEN success prob. is & 0.12 & 460  & 3313 \\
        & AND housing=no                &                       &      &      &      \\
ELSE IF & marital=single                & THEN success prob. is & 0.11 & 550  & 4331 \\
        & AND contact=cellular          &                       &      &      &      \\
ELSE IF & contact=cellular              & THEN success prob. is & 0.08 & 1080 & 12709\\
ELSE    &                               & success prob. is      & 0.04 & 504  & 11998
\end{tabular}
\centering
\caption{Falling rule list for bank-full dataset, trained using the Bayesian approach with $6000$ iterations.} \label{table:bfrl_1}
\end{table}

\begin{table}[h!]
\begin{tabular}{llllll}
 & antecedent & & probability  & positive & negative \\
 &            & &              & support  & support  \\\hline
IF      & poutcome=success              & THEN success prob. is & 0.65 & 978  & 531  \\
        & AND default=no                &                       &      &      &      \\
ELSE IF & 60 $\leq$ age $<$ 100         & THEN success prob. is & 0.29 & 426  & 1030 \\
        & AND loan=no                   &                       &      &      &      \\
ELSE IF & 17 $\leq$ age $<$ 30          & THEN success prob. is & 0.25 & 504  & 1539 \\
        & AND housing=no                &                       &      &      &      \\
ELSE IF & campaign=1                    & THEN success prob. is & 0.15 & 787  & 4471 \\
        & AND housing=no                &                       &      &      &      \\
ELSE    &                               & success prob. is      & 0.07 & 2594 & 32351
\end{tabular}
\centering
\caption{Falling rule list for bank-full dataset, trained using the optimization approach (Algorithm FRL) with $3000$ iterations and the positive class weight $w = 7$.} \label{table:frl_1}
\end{table}

\begin{table}[h!]
\centering
\begin{tabular}{llllll}
 & antecedent & & probability  & positive & negative \\
 &            & &              & support  & support  \\\hline
IF      & poutcome=success              & THEN success prob. is & 0.70 & 729  & 311  \\
        & AND housing=no                &                       &      &      &      \\
ELSE IF & poutcome=success              & THEN success prob. is & 0.53 & 249  & 222  \\
ELSE IF & 60 $\leq$ age $<$ 100         & THEN success prob. is & 0.29 & 426  & 1030 \\
        & AND loan=no                   &                       &      &      &      \\
ELSE IF & 17 $\leq$ age $<$ 30          & THEN success prob. is & 0.25 & 504  & 1538 \\
        & AND housing=no                &                       &      &      &      \\
ELSE IF & education=tertiary            & THEN success prob. is & 0.14 & 790  & 4750 \\
        & AND housing=no                &                       &      &      &      \\
ELSE IF & marital=single                & THEN success prob. is & 0.12 & 648  & 4754 \\
        & AND contact=cellular          &                       &      &      &      \\
ELSE IF & 1000 $\leq$ balance $<$ 2000  & THEN success prob. is & 0.11 & 135  & 1061 \\
        & AND housing=no                &                       &      &      &      \\
ELSE IF & campaign=1                    & THEN success prob. is & 0.10 & 571  & 4904 \\
        & AND contact=cellular          &                       &      &      &      \\
ELSE IF & contact=cellular              & THEN success prob. is & 0.08 & 587  & 6800 \\
        & AND loan=no                   &                       &      &      &      \\
ELSE    &                               & success prob. is      & 0.04 & 650  & 14552
\end{tabular}
\centering
\caption{Falling rule list for bank-full dataset, trained using the Bayesian approach with $6000$ iterations.} \label{table:bfrl_2}
\end{table}

\begin{table}[h!]
\begin{tabular}{llllll}
 & antecedent & & probability  & positive & negative \\
 &            & &              & support  & support  \\\hline
IF      & poutcome=success              & THEN success prob. is & 0.70 & 729  & 311  \\
        & AND housing=no                &                       &      &      &      \\
ELSE IF & poutcome=success              & THEN success prob. is & 0.55 & 185  & 154  \\
        & AND previous $\geq$ 2         &                       &      &      &      \\
ELSE IF & poutcome=success              & THEN success prob. is & 0.48 & 64   & 68   \\
        & AND default=no                &                       &      &      &      \\
ELSE IF & 60 $\leq$ age $<$ 100         & THEN success prob. is & 0.29 & 426  & 1030 \\
        & AND loan=no                   &                       &      &      &      \\
ELSE IF & previous $\geq$ 2             & THEN success prob. is & 0.25 & 302  & 921  \\
        & AND housing=no                &                       &      &      &      \\
ELSE IF & 17 $\leq$ age $<$ 30          & THEN success prob. is & 0.24 & 444  & 1413 \\
        & AND housing=no                &                       &      &      &      \\
ELSE IF & education=tertiary            & THEN success prob. is & 0.13 & 671  & 4435 \\
        & AND housing=no                &                       &      &      &      \\
ELSE    &                               & success prob. is      & 0.07 & 2468 & 31590
\end{tabular}
\centering
\caption{Falling rule list for bank-full dataset, trained using the optimization approach (Algorithm FRL) with $3000$ iterations and the positive class weight $w = 7$.} \label{table:frl_2}
\end{table}

\end{document}